\pdfoutput=1
\newcommand{\JASABlind}{0}
\newcommand{\JASAAuthorBlock}{%
Mehrdad Mohammadi\textsuperscript{1},
Qi Zheng\textsuperscript{2}, and
Ruoqing Zhu\textsuperscript{1,*}\\[-0.2em]
{\small \textsuperscript{1}Department of Statistics, University of Illinois Urbana-Champaign, Champaign, IL 61820, USA}\\[-0.2em]
{\small \textsuperscript{2}Department of Bioinformatics and Biostatistics, University of Louisville, Louisville, KY 40202, USA}\\[-0.2em]
{\small Emails: \texttt{mehrdad3@illinois.edu}; \texttt{qi.zheng@louisville.edu}; \texttt{rqzhu@illinois.edu}}\\[-0.2em]
{\small \textsuperscript{*}Corresponding author: \texttt{rqzhu@illinois.edu}}%
}
\newcommand{\JASAPdfAuthor}{Mehrdad Mohammadi; Qi Zheng; Ruoqing Zhu}
\newcommand{\JASARepositoryText}{Available at \url{https://github.com/mehrdadmhmdi/ke-drl}.}
\newcommand{\JASASimulationRepositoryText}{The simulation code is available in the project repository, \url{https://github.com/mehrdadmhmdi/ke-drl}.}
\newcommand{\JASASupplementAuthorLine}{\par\medskip{\normalsize\JASAAuthorBlock\par}}

\newcommand{\JASAFundingStatement}{This work used Delta and DeltaAI computing resource at University of Illinois Urbana-Champaign from the Advanced Cyberinfrastructure Coordination Ecosystem: Services \& Support (ACCESS) program, supported by U.S. National Science Foundation grants \#2138259, \#2138286, \#2138307, \#2137603, and \#2138296.}

\newcommand{\JASAAIStatement}{During the preparation of this work, the authors used ChatGPT-5.6/Claude Fable to assist with language refinement, manuscript organization, and copyediting to improve readability. Following the use of this tool, the authors thoroughly reviewed, verified, and edited the generated content. The authors assume full responsibility for the data, interpretations, conclusions, and overall academic integrity of the final manuscript.}

\documentclass[12pt]{article}
\usepackage{geometry}
\usepackage{amsmath,amsthm,amssymb,amsfonts,mathrsfs,bm}
\usepackage{titlefoot}
\usepackage{mathabx}
\usepackage{graphicx}
\usepackage{enumerate}
\usepackage{natbib}
\usepackage{url}
\usepackage{pifont}
\usepackage[dvipsnames]{xcolor}
\usepackage[colorlinks=true,linkcolor=purple,citecolor=purple,hyperfootnotes=false]{hyperref}
\usepackage{upgreek}
\usepackage{scalerel}
\usepackage{booktabs}
\usepackage{array}
\usepackage{longtable}
\usepackage{lscape}
\usepackage{relsize,exscale}
\usepackage{algorithm}
\usepackage{algorithmicx}
\usepackage{algpseudocode}
\usepackage{wrapfig}
\usepackage[shortlabels]{enumitem}
\usepackage{soul}
\usepackage{bbold}
\usepackage{setspace}
\usepackage{multirow}
\usepackage{caption}
\usepackage{float}
\usepackage{subcaption}
\usepackage{thmtools}
\AtBeginDocument{\setlength{\bibsep}{0pt}}

\newif\ifsupplementonly
\ifdefined\SUPPLEMENTONLY
  \supplementonlytrue
\else
  \supplementonlyfalse
\fi
\newif\ifmainonly
\ifdefined\MAINONLY
  \mainonlytrue
\else
  \mainonlyfalse
\fi
\ifsupplementonly
  \IfFileExists{main-xrefs.tex}{\input{main-xrefs.tex}}{\typeout{JASA_draft: main-xrefs.tex not found; standalone-supplement cross-references undefined}}
  \makeatletter
  \let\savedHyWarningNoLine\Hy@WarningNoLine
  \let\Hy@WarningNoLine\@gobble
  \hypersetup{draft}
  \AtBeginDocument{\let\Hy@WarningNoLine\savedHyWarningNoLine}
  \makeatother
\fi
\ifmainonly
  \IfFileExists{supplement-xrefs.tex}{\input{supplement-xrefs.tex}}{\typeout{JASA_draft: supplement-xrefs.tex not found; main-to-supplement cross-references undefined}}
  \makeatletter
  \let\savedHyWarningNoLine\Hy@WarningNoLine
  \let\Hy@WarningNoLine\@gobble
  \hypersetup{draft}
  \AtBeginDocument{\let\Hy@WarningNoLine\savedHyWarningNoLine}
  \makeatother
\fi
\providecommand{\JASAAuthorBlock}{Anonymous Author(s)}
\providecommand{\JASAPdfAuthor}{}
\providecommand{\JASARepositoryText}{Available in the anonymized repository accompanying the submission.}
\providecommand{\JASASimulationRepositoryText}{The simulation code is available in the anonymized repository accompanying the submission.}
\providecommand{\JASABlind}{1}
\providecommand{\JASASupplementAuthorLine}{}
\providecommand{\JASAFundingStatement}{}
\newcommand{\JASASupplementSpacing}{1.45}
\newcommand{\JASAReferenceSpacing}{1.45}
\newcommand{\JASADisclosureStatement}{The authors report there are no competing interests to declare.}
\providecommand{\JASAAIStatement}{During the preparation of this work, the authors used ChatGPT-5.6/Claude Fable to assist with language refinement, manuscript organization, and copyediting to improve readability. Following the use of this tool, the authors thoroughly reviewed, verified, and edited the generated content. The authors assume full responsibility for the data, interpretations, conclusions, and overall academic integrity of the final manuscript.}
\newcommand{\JASADeclarations}{%
  \if1\JASABlind\else
    \section*{Funding}\noindent\JASAFundingStatement\par
  \fi
  \section*{Disclosure Statement}\noindent\JASADisclosureStatement\par
  \section*{Declaration of Generative AI Use}\noindent\JASAAIStatement\par
}
\hypersetup{
  pdftitle={Vector-Valued Distributional Reinforcement Learning Policy Evaluation: A Hilbert Space Embedding Approach},
  pdfauthor={\JASAPdfAuthor}
}
\newcommand{\JASATitleText}{Vector-Valued Distributional Reinforcement Learning Policy Evaluation: A Hilbert Space Embedding Approach}
\ifdefined
  \ifsupplementonly
    \AtBeginDocument{\begingroup\edef\x{/Title (Supplementary Material: \JASATitleText) /Author (\JASAPdfAuthor)}\expandafter\pdfinfo\expandafter{\x}\endgroup}
  \fi
  \ifmainonly
    \AtBeginDocument{\begingroup\edef\x{/Title (\JASATitleText) /Author (\JASAPdfAuthor)}\expandafter\pdfinfo\expandafter{\x}\endgroup}
  \fi
\fi

\newcommand{\E}{{\mathbb{E}}}
\newcommand{\T}{\mathcal{T}}

\newcommand{\p}{\mathbb{P}}
\newcommand{\Q}{\mathbb{Q}}
\newcommand{\R}{\mathbb{R}}

\newcommand{\one}{\mathbb{1}}
\newcommand{\bso}{{\boldsymbol \omega}}

\newcommand{\bB}{\mathbb{B}}
\newcommand{\Bgamma}{\raisebox{.12\baselineskip}{\large\ensuremath{\boldsymbol{\gamma}}}}
\newcommand{\indep}{\mathrel{\rotatebox[origin=c]{90}{$\models$}}}

\DeclareMathOperator{\argmin}{argmin}

\makeatletter
\newcommand{\labelword}[2]{%
  \phantomsection
  #1\def\@currentlabel{\unexpanded{#1}}\label{#2}%
}
\makeatother

\newtheorem{theorem}{Theorem}[section]

\newtheorem{proposition}[theorem]{Proposition}
\newtheorem{corollary}[theorem]{Corollary}

\newtheorem{lemma}[theorem]{Lemma}
\newtheorem{definition}[theorem]{Definition}
\newtheorem{remark}[theorem]{Remark}
\newtheorem{assumption}[theorem]{Assumption}
\providecommand*{\theHtheorem}{}
\providecommand*{\theHclaim}{}
\providecommand*{\theHproposition}{}
\providecommand*{\theHcorollary}{}
\providecommand*{\theHconjecture}{}
\providecommand*{\theHlemma}{}
\providecommand*{\theHdefinition}{}
\providecommand*{\theHremark}{}
\providecommand*{\theHassumption}{}
\renewcommand*{\theHtheorem}{theorem.\thesection.\arabic{theorem}}
\renewcommand*{\theHclaim}{claim.\thesection.\arabic{theorem}}
\renewcommand*{\theHproposition}{proposition.\thesection.\arabic{theorem}}
\renewcommand*{\theHcorollary}{corollary.\thesection.\arabic{theorem}}
\renewcommand*{\theHconjecture}{conjecture.\thesection.\arabic{theorem}}
\renewcommand*{\theHlemma}{lemma.\thesection.\arabic{theorem}}
\renewcommand*{\theHdefinition}{definition.\thesection.\arabic{theorem}}
\renewcommand*{\theHremark}{remark.\thesection.\arabic{theorem}}
\renewcommand*{\theHassumption}{assumption.\thesection.\arabic{theorem}}

\newcommand{\mb}{\mathbb}
\newcommand{\mc}{\mathcal}
\newcommand{\tx}{\text}
\newcommand{\bx}{{\mathbf x}}
\newcommand{\bX}{{\mathbf X}}
\newcommand{\by}{{\mathbf y}}
\newcommand{\bz}{{\mathbf z}}
\newcommand{\bZ}{{\mathbf Z}}
\newcommand{\bR}{{\mathbf R}}
\newcommand{\br}{{\mathbf r}}
\newcommand{\bu}{{\mathbf u}}
\newcommand{\bw}{{\mathbf w}}
\newcommand{\bc}{{\mathbf c}}
\newcommand{\bk}{{\bf k}}
\newcommand{\bs}{{\bf s}}
\newcommand{\ba}{{\bf a}}
\newcommand{\bS}{{\bf S}}
\newcommand{\bA}{{\bf A}}

\newcommand{\mbC}{{\mb C}}

\newcommand{\balpha}{{\boldsymbol \alpha}}

\def\h2s{{\hspace{0.2in}}}

\begin{document}
\bibliographystyle{chicago}
\def\spacingset#1{\renewcommand{\baselinestretch}%
{#1}\small\normalsize}
\spacingset{1}
\ifsupplementonly
\else
\if1\JASABlind
\bigskip\bigskip\bigskip
\begin{center}
{\LARGE\bf Vector-Valued Distributional Reinforcement Learning Policy Evaluation: A Hilbert Space Embedding Approach}
\end{center}
\smallskip\bigskip
\else
\title{\bf Vector-Valued Distributional Reinforcement Learning Policy Evaluation: A Hilbert Space Embedding Approach}
\author{\JASAAuthorBlock}
\date{}
\maketitle
\fi
\medskip
\begin{abstract}
\noindent We propose Kernel Embedding Distributional Reinforcement Learning (KE-DRL) an offline method to estimate kernel mean embeddings of conditional multivariate return distributions from observed offline trajectories with continuous state-action inputs and vector-valued rewards. KE-DRL approximates the conditional embedding and estimates its a finite-dictionary representation coefficients through a maximum mean discrepancy Bellman criterion. For a regular class of return distributions, we show that a Mat\'ern embedding and a Bellman-invariant Sobolev-moment class identifies the unique distributional Bellman fixed point and and derive a H\"older-type bound that controls Wasserstein error by the population embedding Bellman residual. For directly observed responses, we provide finite-sample and uniform error bounds for a regularized conditional mean embedding estimator. Simulation experiments evaluate pointwise embedding recovery against Monte Carlo benchmarks across multiple behavior-target policy pairs. An application to Expedia hotel-search data illustrates conditional multivariate return evaluation under two policies.
\end{abstract}
\noindent%
{\it Keywords:} Distributional Reinforcement Learning, Off-policy Evaluation, Kernel Mean Embeddings, Maximum Mean Discrepancy, Mat\'ern Kernels
\vfill
\newpage
\spacingset{1.75}
\section{Introduction}\label{sec:intro}
Many policy-evaluation problems involve continuous actions and several long-term outcomes. Examples include pricing and promotion policies on digital platforms and medical treatment policies that balance efficacy, adverse events, and cost. Such policies often cannot be evaluated experimentally and inference relies on offline reinforcement learning (RL) trajectories generated under a behavior policy. The object of evaluation is then the target-policy conditional law of the discounted return vector, and policies that agree in mean return can differ in the joint law of their reward coordinates. Distributional RL (DRL) generalizes classical RL by modeling the return distribution rather than only its expectation \citep{bellemare2017distributional}. Algorithms such as C51, QR-DQN, and MMDQN \citep{dabney2018distributional,nguyen2021distributional} have reported empirical improvements, but the methodology and theory are most developed for scalar returns. Extending these methods to continuous multivariate returns requires suitable representations and tractable discrepancies.

\noindent A core challenge is defining a tractable discrepancy between multivariate return distributions. Wasserstein distances provide the Bellman contraction geometry \citep{bellemare2017distributional} but can be computationally costly in multiple dimensions \citep{peyre2019computational}. Recent kernel, Sinkhorn, and embedding-based method approximations reduce this cost while introducing additional approximation choices and 
provide alternative representations for continuous return distributions \citep{xi2022distributional,sun2024distributional,wenliang2024distributional_embedding}. This motivates a discrepancy that is computationally tractable and admits a formal comparison with the Wasserstein metric. For finite-dimensional projected operators, contraction under the feature-space norm is not automatic unless the representation is compatible with the Bellman update.

\noindent This paper develops \emph{Kernel Embedding Distributional Reinforcement Learning (KE-DRL)}, which combines conditional mean embeddings with an MMD Bellman criterion for offline multivariate distributional policy evaluation. The underlying distributional object is the target-policy conditional return law, and the primary statistical estimand is its kernel mean embedding. Accurate recovery of distributional features from that embedding is a separate, generally ill-posed inverse problem rather than part of the primary estimand (See Supplementary Material~\ref{app:emb_ill_posed}). The RKHS representation provides a computable discrepancy through the maximum mean discrepancy (MMD) \citep{song2009hilbert}. Its relation to Wasserstein geometry depends on the kernel and the class of return laws. Gaussian kernels do not satisfy the Fourier-decay condition used in our comparison \citep{sriperumbudur2010hilbert,vayer2023controlling}. We therefore use the $\mbox{Mat\'{e}rn}$ family, whose polynomial spectral decay supports a comparison between MMD and $W_1$ on the regular distribution class stated in Section~\ref{sec:theory}. The kernel representation yields a tractable MMD Bellman criterion, while the comparison connects that criterion to $W_1$ on the stated regular class.

\noindent When rewards are vector-valued, most multi-objective methods use scalarization or reward decomposition. They estimate several scalar value functions and combine them through a preference model, rather than estimating the joint discounted return law in $\R^d$; see the survey \citet{roijers2013survey} and representative reward-decomposition methods in \citet{van2017hybrid,lin2020rd2}. A smaller literature directly studies multivariate distributional policy evaluation and off-policy evaluation. This literature includes generative formulations based on the distributional Bellman equation \citep{freirich2019distributional}, multidimensional DRL objectives for joint returns \citep{zhang2021distributional}, and recent likelihood-based or Bellman-residual methods for estimating return distributions from offline data \citep{wu2023distributional,hong2025distributional}, including Wasserstein-based estimation of multivariate discounted return distributions \citep{qi2025distributional}. Against this background, KE-DRL combines conditional mean embeddings with a Mat\'ern-induced MMD Bellman criterion for continuous state and action inputs and vector-valued rewards. The fitted embedding supports plug-in evaluation of functions in $\mc H_{\mc Z}$. Sharper quantities, such as densities or nonsmooth tail probabilities, require regularized post-estimation recovery, as discussed in Section~\ref{sec:stat_recovery}. We evaluate the embedding estimator in simulation and illustrate both the estimated embedding and the regularized recovery procedure in the Expedia analysis.

\noindent Our contributions are threefold:
(\textbf{i}) We develop KE-DRL, an offline algorithm for estimating conditional mean embeddings of multivariate target-policy returns with continuous state and action inputs.
(\textbf{ii}) On the stated Bellman-invariant Sobolev-moment class, we prove that the Mat\'ern mean-embedding discrepancy uniquely characterizes the target-policy return distribution as the Wasserstein Bellman fixed point. We further derive finite-sample error bounds and uniform convergence rates for the regularized conditional mean embedding estimator constructed from observed transition-return pairs. (\textbf{iii}) We provide a data-adaptive return-grid construction, a held-out-input projected Bellman diagnostic, simulation and real-data illustrations, and an open-source implementation.\footnote{\JASARepositoryText}

\noindent The remainder of the paper is organized as follows. Section~\ref{sec:methodology} formalizes offline multivariate distributional policy evaluation and constructs the KE-DRL estimator. Section~\ref{sec:theory} establishes the population fixed-point identification result and error bounds for a regularized conditional mean embedding estimator. Section~\ref{sec:implementation} presents the estimation algorithm, a held-out Bellman-residual diagnostic, and post-estimation recovery of probability summaries. Sections~\ref{sec:sim} and~\ref{sec:realdata} report the simulation study and the Expedia hotel-search application, and Section~\ref{sec:conclusion} concludes. Proofs, derivations, and additional implementation and empirical details are collected in the Supplementary Material.
\section{Methodology}\label{sec:methodology}
\noindent This section formalizes the estimation problem and builds the KE-DRL estimator in stages. Section~\ref{sec:background} describes the offline data and the distributional Bellman equation that characterizes the target return law. Section~\ref{sec:kernel_ipm} represents conditional return laws by kernel mean embeddings and introduces a finite-dictionary approximation of the conditional embedding map. Section~\ref{sec:bellman_discrepancy} embeds the Bellman update itself, yielding a computable projected Bellman residual, and Section~\ref{sec:ope_weights} identifies the target-policy quantities in that residual from behavior-policy data. Section~\ref{sec:optimization} assembles these components into the regularized criterion that defines the estimator.
\subsection{Offline Distributional Policy Evaluation}\label{sec:background}


\noindent We observe an offline dataset generated by a stationary behavior policy $\beta$ in a Markov decision process with an unknown transition kernel. The dataset comprises $n$ trajectories, $\mathcal{D} = \{\mathbf{h}_i\}_{i=1}^n$, where each trajectory $\mathbf{h}_i = \{(\bs_{it}, \ba_{it}, \br_{it})\}_{t=0}^{T_i-1}$ contains $T_i$ observations with states $\bs_{it} \in \mathcal{S} \subset \mathbb{R}^p$, actions $\ba_{it} \in \mathcal{A} \subset \mathbb{R}^q$, and vector-valued rewards $\br_{it} \in \mathcal{R} \subset \mathbb{R}^d$. 

\noindent For estimation, we pool the one-step transitions across all trajectories for $t = 0, \ldots, T_i - 2$, yielding a sample $\{(\bx_j, \br_j, \bx_j')\}_{j=1}^N$ of size $N = \sum_{i=1}^n (T_i - 1)$. Here, $\bx_j = (\bs_j^\top, \ba_j^\top)^\top \in \mathcal{X} := \mathcal{S} \times \mathcal{A}$ denotes the current state-action pair, and $\bx_j' = ((\bs_j')^\top, (\ba_j')^\top)^\top$ denotes the observed successor state-action pair.

\paragraph{Notation.} 
Random variables are denoted by uppercase letters (e.g., $\bX = (\bS^\top, \bA^\top)^\top$), and their realizations by lowercase bold letters (e.g., $\bx = (\bs^\top, \ba^\top)^\top$). Scalars are set in standard weight, while vectors are bolded. Gram matrices, operator matrices, and the coefficient matrix $\mathbb{B}$ appear in blackboard bold ($\mathbb{X}, \mathbb{B}$). A prime (${}'$) denotes generic or sample successor variables (e.g., $\bX', \bx', \bx_j'$). We use the compact joint notation $\bx$ when state and action need not be distinguished, and the expanded tuple $(\bs, \ba)$ when explicitly referring to the underlying transition dynamics. A complete notation summary is provided in Supplementary Material~\ref{app:notation_reference}.

\noindent  For a target policy $\pi$ and discount factor $\gamma \in (0,1)$, the conditional discounted return vector given an initial state-action pair $\bx = (\bs^\top, \ba^\top)^\top$ is defined as
\begin{equation}\label{eq:return_def_new}
\bZ^\pi(\bx)
:=
\sum_{t=0}^{\infty}\gamma^t\,\bR(\bS_t,\bA_t),
\qquad
\bS_0=\bs,\quad \bA_0=\ba
\end{equation}
where $\bS_{t+1}\sim p(\cdot\mid \bS_t,\bA_t)$ and $\bA_{t+1}\sim \pi(\cdot\mid \bS_{t+1})$ for all $t\ge0$. We denote the conditional law of this multivariate return by $\mc V^\pi_{\bx} := \mc L\{\bZ^\pi(\bx)\}$, which serves as our core distributional object. The primary statistical estimand is its kernel mean embedding, introduced in Section~\ref{sec:kernel_ipm}. 

\noindent Let $\bX' = (\bS'^\top, \bA'^\top)^\top$ with $\bS'\sim p(\cdot\mid\bs,\ba)$ and $\bA'\sim\pi(\cdot\mid\bS')$. The return distribution satisfies the distributional Bellman equation \citep{rosler1992fixed,bellemare2017distributional}:
\begin{equation}\label{eq:DBO_new}
\bZ^\pi(\bx)
\overset{D}{=}
\bR(\bx) + \gamma\,\bZ^\pi(\bX')
\end{equation}
For any candidate return process $\bZ$, the distributional Bellman operator $\T^\pi$ is defined by
\begin{equation}\label{eq:DBO}
(\T^\pi\bZ)(\bx)
\overset{D}{=}
\bR(\bx) + \gamma\,\bZ(\bX')
\end{equation}
where $(\bR, \bX')$ is distributed according to the joint one-step target-policy transition law $\mathcal{M}^\pi(d\br, d\bx' \mid \bx)$. Comparing \eqref{eq:DBO} with \eqref{eq:DBO_new} shows that the law of $\bZ^\pi$ is a distributional fixed point of $\T^\pi$. 

\noindent We let $\mc{P}_X^\pi(d\bx'\mid\bx)$ and $F_{\bR}(d\br\mid\bx)$ denote the $\bX'$- and $\bR$-marginals of $\mathcal{M}^\pi(\cdot\mid\bx)$, respectively. Because the successor action $\bA'\sim\pi(\cdot\mid\bS')$ is selected after the reward $\bR$ is realized, $\bA'$ is conditionally independent of $\bR$ given $(\bX, \bS')$. Consequently, the transition kernel $\mc{P}_{\bX}^\pi$ depends directly on $\pi$, whereas the reward distribution $F_{\bR}(\cdot\mid\bx)$ remains policy-invariant for all $\bx$.

\noindent In the univariate scalar-reward setting, \citet{rosler1992fixed} and \citet{bellemare2017distributional} established that the distributional Bellman operator is a contraction mapping under the Wasserstein metric, ensuring a unique fixed point. Our development departs from this classical setting in two key respects:
\begin{enumerate}
    \item \textbf{Multivariate Returns and Kernel Embeddings:} Because the returns here are multivariate, direct computation of Wasserstein distances is computationally intractable. Sections~\ref{sec:kernel_ipm} and~\ref{sec:bellman_discrepancy} circumvent this by comparing return laws via maximum mean discrepancies in a reproducing kernel Hilbert space (RKHS), which can be efficiently estimated from offline data. Section~\ref{sec:theory} proves that this discrepancy uniquely identifies the true fixed point.
    \item \textbf{Off-Policy Distribution Shift:} The logged trajectories are generated under the behavior policy $\beta$ rather than $\pi$. Identifying target-policy distributions from these offline samples requires the Markov property, sequential conditional exchangeability, and overlap conditions formalized in Assumption~\ref{ass:off_policy_identification} (Section~\ref{sec:ope_weights}).
\end{enumerate}

\subsection{Kernel Representation of Conditional Return Laws} \label{sec:kernel_ipm}

Let $k_{\mc Z}$ and $k_{\mc X}$ be positive-definite kernels on the return space $\mc Z \subset \R^d$ and $\mc X = \mc S \times \mc A$, with RKHSs $\mc H_{\mc Z}$ and $\mc H_{\mc X}$, respectively. For any probability law $\Q$ on $\mc Z$ satisfying 
$\E_{\bZ, \bZ'\sim \Q}\{k_{\mc Z}(\bZ,\bZ')\}^{1/2}<\infty$
its kernel mean embedding is $\mu_{\Q}(\cdot):=\E_{\bZ\sim\Q}\{k_{\mc Z}(\bZ,\cdot)\} \in \mc H_{\mc Z}$. For every $g\in\mc H_{\mc Z}$, $\E_{\bZ\sim\Q}\{g(\bZ)\}=\langle g,\mu_{\Q}\rangle_{\mc H_{\mc Z}}$. For a given $\bx\in\mc X$, denote the conditional mean embedding of 
a return $\bZ|\bX=\bx$ by
\enlargethispage{2pt}
\[
\mu_{\bZ\mid\bX=\bx}(\cdot):= \E\{k_{\mc Z}(\bZ,\cdot)\mid\bX=\bx\}.
\]
Then  the conditional mean embedding of the return law under the target-policy is denoted by $\mu_{\bZ^{\pi}\mid\bX=\bx}(\cdot)$. Were target-policy returns observable at the sample inputs, the conditional embedding could be estimated directly via regularized kernel regression; Section~\ref{sec:tikhonov_reference} analyzes this ideal estimator as a theoretical benchmark. In offline settings, however, such returns are inaccessible. To circumvent this, the proposed KE-DRL estimator approximates the conditional embedding on a finite return grid, expands the coefficient functions over a state-action basis, and fits them by matching both sides of the distributional Bellman equation---as developed in the remainder of this section.

\noindent Let $\mathbf Z_m:=\{\bz_1,\ldots,\bz_m\}\subset\mc Z$ be the return grid and define $\bk_{\mc Z}(\cdot):=\big(k_{\mc Z}(\bz_1,\cdot),\ldots,k_{\mc Z}(\bz_m,\cdot)\big)^\top$. The integer $m$ controls only the number of return atoms used to represent the multivariate return embedding. Supplementary Material~\ref{app:Grid-Z} describes the adaptive construction of the return grid $\mathbf Z_m$ used in our implementation. Independently, we choose a state-action basis dictionary $\mc C_L:=\{\underline \bx_1,\ldots,\underline \bx_L\}\subseteq\mc X$, where $L$ is chosen by the analyst. The basis points may be the full set of pooled inputs, a subsample, inducing points, cluster centers, or any design covering the empirical support. Define $\bk_L(\bx):=\big( k_{\mc X}(\underline \bx_1,\bx),\ldots, k_{\mc X}(\underline \bx_L,\bx)\big)^\top\in\R^L$ and $\mathbb K_L:=\big[ k_{\mc X}(\underline \bx_\ell,\underline \bx_r)\big]_{\ell,r=1}^{L}\in\R^{L\times L}$.

\noindent For a target policy $\pi$, approximate the target conditional embedding $\mu_{\bZ^{\pi}\mid\bx}$ by the following finite-dictionary RKHS element: 
 \begin{equation}\label{eq:mu_Z}
\widehat\mu_{\bZ|\bx}(\cdot\ ;\mathbb B)
=\sum_{i=1}^{m}\omega_i(\bx;\mathbb B)k_{\mc Z}(\bz_i,\cdot)
= \bso(\bx;\mathbb B)^\top\bk_{\mc Z}(\cdot)
\end{equation}
with some matrix $\mathbb B\in\R^{L\times m}$

\begin{equation}\label{eq:mu_Z}
\widehat\mu_{\bZ\mid\bx}(\cdot\ ;\mathbb B)
=\sum_{i=1}^{m}\omega_i(\bx;\mathbb B)k_{\mc Z}(\bz_i,\cdot)
= \bso(\bx;\mathbb B)^\top\bk_{\mc Z}(\cdot)
\end{equation}
where $\bso(\bx;\mathbb B)=(\omega_1(\bx;\mathbb B),\ldots,\omega_m(\bx;\mathbb B))^\top\in\R^m$. The coefficient functions are parameterized by the matrix $\mathbb B\in\R^{L\times m}$, such that
\begin{align}\label{eq:hat_w}
\omega_i(\bx;\mathbb B)
&= \sum_{\ell=1}^{L}b_{\ell i} k_{\mc X}(\underline \bx_\ell,\bx) \text{ and }
\bso(\bx;\mathbb B) = \mathbb B^\top\bk_L(\bx) 
\end{align}
The map $\widehat\mu_{\bZ\mid\bx}(\cdot\ ;\mathbb B)$ is defined for an arbitrary matrix $\mathbb B$: the hat indicates a finite-dictionary approximation to the conditional embedding, not that $\mathbb B$ has been estimated from data. 

\noindent Because the entries of $\mathbb B$ are unconstrained, the candidate is an RKHS element corresponding to a finite signed measure and need not be the embedding of a probability law. The fitted map is obtained by substituting the optimized matrix $\mathbb B=\widehat{\mathbb B}^{\pi}$ from Section~\ref{sec:optimization}. The coefficient matrix has dimension $L\times m$, whereas the standard empirical conditional-embedding estimator attaches one coefficient to each of the $N$ pooled transitions; when $L\ll N$, the parameterization is correspondingly smaller. Moreover, a single policy-specific matrix defines the map at every evaluation point, so no separate optimization is carried out at individual inputs.

\noindent We therefore distinguish the target conditional law, its RKHS mean embedding, the possibly signed finite-dictionary candidate in \eqref{eq:mu_Z}, and probability summaries obtained through the recovery procedure in Section~\ref{sec:stat_recovery}.

\subsection{Pushforward Embeddings and the Projected Bellman Residual} \label{sec:bellman_discrepancy}
\noindent This subsection connects the conditional return embedding to the distributional Bellman recursion through the pushforward of the return law under the Bellman map. The Bellman target at input $\bx$ has the affine form $\bR+\gamma\bZ(\bX')$. Rather than estimating the target density directly, we embed the law of this Bellman target in $\mc H_{\mc Z}$. 

\noindent For any candidate return process $\bZ=\{\bZ(\bx):\bx\in\mc X \}$, let $\Q_\bZ(\cdot\mid\bx)$ denote its conditional return-law map given $\bx$. Then the conditional mean embedding of the candidate law at $\bx$ is $\mu_{\bZ\mid\bx}(\cdot):=\int_{\mc Z}
k_{\mc Z}(\bz,\cdot)\ d\Q_\bZ(\bz\mid\bx)$. The conditional law produced by the distributional Bellman operator is thus
the mixture
\begin{equation}\label{eq:pushforward_bellman_law}
\Q_{\T^\pi\bZ}(\cdot\mid\bx)
=
\int_{\mc R\times\mc X}
\Q_\bZ\left(\frac{\cdot-\br}{\gamma}\mid\bx'\right)\
\mathcal{M}^\pi(d\br,d\bx'\mid\bx)
\end{equation}
By the definition of mean embedding and  a change of variables, 
\begin{align}
\h2s \mu_{\T^\pi\bZ \mid\bX=\bx}(\cdot)&=\int_{\mc Z}k_{\mc Z}(\bz,\cdot)
\int_{\mc R\times\mc X}
\left\{\frac{1}{\gamma}\ d\Q_\bZ\left(\frac{\bz-\br}{\gamma}\mid\bx'\right)\right\}
\mathcal{M}^\pi(d\br,d\bx'\mid\bx) \nonumber\\
& = \int_{\mc R\times\mc X}
\left\{\int_{\mc Z}k_{\mc Z}(\br+\gamma\bz,\cdot)\ d\Q_\bZ(\bz\mid\bx')\right\}
\mathcal{M}^\pi(d\br,d\bx'\mid\bx)
\label{eq:pushforward_bellman_eb}
\end{align}
which embeds the Bellman mixture (see \eqref{eq:pushforward_bellman_law}) directly in $\mc H_{\mc Z}$.
Lemma~\ref{lem:pushforward_operator} in Supplementary Material~\ref{app:pushforward_operator} shows that, for the Mat\'ern return kernel, $\int_{\mc Z}k_{\mc Z}(\br+\gamma\bz,\cdot)\ d\Q_\bZ(\bz\mid\bx')$ in \eqref{eq:pushforward_bellman_eb} is well defined for every embeddable $\Q$ and satisfies $\|\int_{\mc Z}k_{\mc Z}(\br+\gamma\bz,\cdot)\ d\Q_\bZ(\bz\mid\bx')\|_{\mc H_{\mc Z}}\le c_\gamma\|\int_{\mc Z}k_{\mc Z}(\bz,\cdot)\ d\Q_\bZ(\bz\mid\bx')$ for a constant $c_\gamma$ that does not depend on $\br$.

\noindent The criterion compares two embeddings generated by the same conditional law map $\Q_\bZ$. The first directly embeds $\Q_\bZ(\cdot\mid\bx)$, while the second embeds the mixture of pushforward laws in \eqref{eq:pushforward_bellman_law}. Consequently, the discrepancy is not evaluated between two separately estimated return distributions. The candidate $\Q_\bZ$ is a distributional Bellman fixed point if and only if these two embeddings coincide for all $\bx \in \mc X$. The corresponding pointwise population Bellman residual, measured by the RKHS distance induced by the kernel \(k_{\mathcal Z}\), is
\begin{equation}\label{eq:exact_bellman_discrepancy}
\Bgamma_{k}(\bx; \Q_\bZ) := \left\| \mu_{\bZ\mid\bx} - \mu_{\T^\pi\bZ\mid\bx} \right\|_{\mc H_{\mc Z}}
\end{equation}

\noindent Given a weighting distribution $\mc G$ on $\mc X$, the population Bellman loss is
\begin{equation}\label{eq:exact_population_loss}
\mc L(\Q_\bZ)
:=
\int_{\mc X}
\Bgamma_{k}^2(\bx; \Q_\bZ) \, \mathrm{d}\mc G(\bx)
\end{equation}

\noindent The population criterion above is defined for conditional law maps. For computation, the pushforward and integration operations extend linearly to the finite signed combinations of kernel sections in \eqref{eq:mu_Z}. Applied to a candidate coefficient matrix $\bB$, this extension gives the finite-dictionary Bellman target embedding
\begin{equation}\label{eq:projected_pair_embedding}
\widehat{\mu}_{\T^\pi\bZ\mid\bx}(\cdot;\mb B)
=
\int_{\mc R\times\mc X}
\sum_{i=1}^m
\omega_i(\bx';\bB)\,
k_{\mc Z}(\br+\gamma\bz_i,\cdot)\
\mathcal{M}^\pi(d\br,d\bx'\mid\bx)
\end{equation}
See results in the upplementary Material~\ref{app:pushforward_operator}. This is the exact linear Bellman image of the finite-dictionary candidate, but it still involves a joint integral, under $\mathcal{M}^\pi(\cdot\mid\bx)$, over the immediate reward and the successor state-action input. The following assumption states when this joint integral factors into separately estimable continuation and reward-shift terms. It is not implied by the finite return grid or by the RKHS representation alone.

\begin{assumption}[Reward-continuation moment factorization]\label{ass:reward_continuation_separation}
Let $\mc X_{\mathrm{sep}}\subseteq\mc X$ be a fixed region containing the support of any weighting distribution used in the population criterion and all fitting or held-out inputs at which the separated construction is evaluated. For every $\bx\in\mc X_{\mathrm{sep}}$, every return-grid atom $\bz_i$, and every candidate matrix $\mathbb B\in\R^{L\times m}$, the relevant Bochner integrals exist in $\mc H_{\mc Z}$ and
\[
\begin{aligned}
&\int_{\mc R\times\mc X}
\omega_i(\bx';\bB)\,
k_{\mc Z}(\br+\gamma\bz_i,\cdot)\,
\mathcal{M}^\pi(d\br,d\bx'\mid\bx)
\\
&\quad=
\left\{
\int_{\mc X}\omega_i(\bx';\bB)\,\mc{P}_X^\pi(d\bx'\mid\bx)
\right\}
\times
\left\{
\int_{\mc R}k_{\mc Z}(\br+\gamma\bz_i,\cdot)\,F_{\bR}(d\br\mid\bx)
\right\}
\end{aligned}
\]
Here equality is in $\mc H_{\mc Z}$. The left-hand side integrates the pair $(\br,\bx')$ jointly under the one-step law $\mathcal{M}^\pi(\cdot\mid\bx)$; the right-hand side is the product of the integrals of the two factors under the marginals of that same law, $\mc{P}_X^\pi(\cdot\mid\bx)$ for the successor input and $F_{\bR}(\cdot\mid\bx)$ for the reward, which does not depend on the target policy.
\end{assumption}
\noindent Assumption~\ref{ass:reward_continuation_separation} is a conditional zero-covariance restriction between the reward-shift feature and the continuation coefficient. It does not make the reward deterministic or independent of the full future return. The condition $\bR\indep\bS'\mid\bX$ is sufficient, but is not imposed directly. The factorization is used only for the separated finite-dimensional construction below; it neither defines the target return law nor enters the population fixed-point result. Supplementary Material~\ref{app:paired_bellman_estimator} gives the full interpretation, conditions under which the restriction can fail, and a paired reward-transition construction that preserves $(\bR,\bX')$ when separation is inappropriate.

\begin{proposition}[Finite-dictionary Bellman target embedding]\label{prop:DBO_empirical}
Under Assumption~\ref{ass:reward_continuation_separation}, for a fixed candidate matrix $\mathbb B$ and any $\bx\in\mc X_{\mathrm{sep}}$, the distributional Bellman operator embedding \eqref{eq:projected_pair_embedding} has the linear form
\begin{align}\label{eq:mu_TZ}
\widehat{\mu}_{\T^\pi\bZ\mid\bx}(\cdot;\mb B)
&=
\sum_{i=1}^m\omega_i^\pi(\bx;\mathbb B)\ell_i(\bx,\cdot)
\end{align}
where
\[
\omega_i^\pi(\bx;\mathbb B)
:=\int_{\mc X}\omega_i(\bx';\mathbb B)\,\mc{P}_X^\pi(d\bx'\mid\bx)
\qquad
\ell_i(\bx,\cdot)
:=\int_{\mc R}k_{\mc Z}(\br+\gamma\bz_i,\cdot)\,F_{\bR}(d\br\mid\bx)
\]
integrate the continuation coefficient and the reward-shifted kernel section under the $\bX'$- and $\bR$-marginals of $\mathcal{M}^\pi(\cdot\mid\bx)$, respectively.
\end{proposition}
\noindent Thus the Bellman target embedding is a weighted sum of reward-shifted kernel sections. The shift is induced by $\gamma\bz_i$, and the weights average the successor-input coefficients under the target-policy transition law. In offline policy evaluation, $\omega_i^\pi(\bx;\mathbb B)$ is not available by direct sampling from $\pi$; Section~\ref{sec:ope_weights} estimates it from behavior-policy data using density-ratio weights.
\noindent For each $\bx\in\mc X$, define the squared projected Bellman residual by the RKHS norm
\begin{equation}\label{eq:gamma_def}
\widehat{\Bgamma}_{k}^{\,2}(\bx;\mathbb B)
:=
\big\|\widehat{\mu}_{\T^\pi\bZ\mid\bx}(\cdot;\mb B)-\widehat{\mu}_{\bZ\mid\bx}(\cdot;\mb B)\big\|_{\mc H_{\mc Z}}^2
\end{equation}
Using \eqref{eq:mu_Z} and \eqref{eq:mu_TZ}, the squared residual \eqref{eq:gamma_def} has the following pointwise matrix form.
\begin{align}
\widehat{\Bgamma}_{k}^{\,2}(\bx;\mathbb B)
&=
\big\|\widehat{\mu}_{\T^\pi\bZ\mid\bx}-\widehat{\mu}_{\bZ\mid\bx}\big\|_{\mc H_{\mc Z}}^2
\nonumber\\
&=
\bso(\bx;\mathbb B)^\top
\mathbb K_{\mc Z}
\bso(\bx;\mathbb B)
-2\bso(\bx;\mathbb B)^\top
\mathbb H(\bx)
\bso^\pi(\bx;\mathbb B)
+\bso^\pi(\bx;\mathbb B)^\top
\mathbb G(\bx)
\bso^\pi(\bx;\mathbb B)
\label{eq:gamma_with_W}
\end{align}
where $\mathbb K_{\mc Z}:= \big[k_{\mc Z}(\bz_i,\bz_j)\big]_{i,j=1}^m$ and $\bso^{\pi}(\bx;\mathbb B) := \big(\omega_1^\pi(\bx;\mathbb B),\ldots,\omega_m^\pi(\bx;\mathbb B)\big)^\top$. The matrices $\mathbb H(\bx)$ and $\mathbb G(\bx)$ collect, respectively, the inner products between grid sections and reward-shifted Bellman sections and between pairs of reward-shifted sections. Their entries are conditional expectations of kernel evaluations given $\bX=\bx$. All such conditional expectations, both here and in the off-policy continuation weights of Section~\ref{sec:ope_weights}, are estimated with a common kernel-ridge conditional weight vector
\begin{equation}\label{eq:Gamma}
\boldsymbol{\Gamma}(\bx)
:=
\left(\mathbb K_{\mc X}+N\lambda_{\Gamma}\mathbb I_N\right)^{-1}
\widetilde{\bk}_{N}(\bx),
\qquad
\widetilde{\bk}_{N}(\bx)
:= \big( k_{\mc X}(\bx_1,\bx),\ldots, k_{\mc X}(\bx_N,\bx)\big)^\top 
\end{equation}
Supplementary Material~\ref{drive:gamma_with_W} gives the entrywise definitions, plug-in formulas, and full algebraic expansion. The pointwise residual is averaged over $\mc X_\star$ to form the criterion in Section~\ref{sec:optimization}.

\subsection{Off-policy Weighting from Behavior Policy Data}\label{sec:ope_weights}
Identification of target-policy continuation expectations from behavior-policy transitions requires the following conditions.
\begin{assumption}[Offline identification and overlap]\label{ass:off_policy_identification}
For the logged trajectories generated by the stationary behavior policy $\beta$, assume:
\begin{enumerate}[label=(\roman*)]
\vspace{-1em}\item \textit{Markov property.} The observed one-step variables are generated from a time-homogeneous Markov kernel, so that the law of $(\bR_t,\bS_{t+1})$ given the observed history depends on the history only through $(\bS_t,\bA_t)$.
\vspace{-1em}\item \textit{Sequential exchangeability.} Conditional on the current state and action, the logged one-step reward and transition kernel coincide with the kernel that would be observed under an intervention assigning that action.
\vspace{-1em}\item \textit{Positivity (overlap).} $\forall \bs \in \mathcal{S}, \ \ \tx{supp}\big(\pi(\boldsymbol{\cdot}|\bs)\big) \subseteq \tx{supp}\big(\beta(\boldsymbol{\cdot}|\bs)\big)$. That is for every state $\bs$ in the evaluation region, $\pi(\ba\mid\bs)>0$ implies $\beta(\ba\mid\bs)>0$. In continuous-action settings this is understood as absolute continuity of the target action density with respect to the behavior action density.
\end{enumerate}
\end{assumption}

\noindent In dominated form, with $p^\pi(\bx'\mid\bx)$ the density of $\mc{P}_X^\pi(d\bx'\mid\bx)$, the target policy enters the Bellman target embedding through the continuation coefficient $\omega_i^\pi(\bx;\mathbb B)=\int_{\mc X}\omega_i(\bx';\mathbb B)\,p^\pi(\bx'\mid\bx)\,d\bx'$ of Proposition~\ref{prop:DBO_empirical}. Because the transition law for $\bS'$ is common under $\pi$ and $\beta$, the importance ratio for the successor action is $\eta(\bx') = \frac{p^\pi(\bx'\mid\bx)}{p^\beta(\bx'\mid\bx)}= \frac{\pi(\ba'\mid\bs')}{\beta(\ba'\mid\bs')}$ for $\bx'=(\bs'^\top,\ba'^\top)^\top$. Hence
\[
\omega_i^\pi(\bx;\mathbb B)
=
\int_{\mc X}\eta(\bx')\,\omega_i(\bx';\mathbb B)\,p^\beta(\bx'\mid\bx)\,d\bx'
\]
\noindent The plug-in construction additionally requires $\eta$ to be known or consistently estimated on the evaluation region, with fitted values bounded or truncated so that the weighted kernel averages below are well defined. This is an estimation condition rather than an identification condition.
Using state-matched samples, uLSIF \citep{kanamori2009least} estimates the successor-action density ratio $\pi(\ba'\mid\bs')/\beta(\ba'\mid\bs')$, not a ratio of policy-specific state-occupancy laws. We model this ratio as $\widehat\eta(\bx)=\sum_{\ell=1}^{b_\eta}\alpha_\ell k_{\mc X}(\bx_{\eta,\ell},\bx)$, where the $b_\eta$ kernel centers may be selected from either uLSIF sample, and estimate the coefficients by uLSIF; see Supplementary Material~\ref{app:uLSIF}. For nonterminal transitions, set $\widehat\eta_j:=\widehat\eta(\bx_j')$ and $\mathbb D_\eta:=\operatorname{diag}( \widehat\eta_1,\ldots,\widehat\eta_N)$.
To express the continuation coefficient vector in the $L$-dimensional basis of $\mathbb B$, define the basis-to-successor cross-Gram matrix $\mathbb K_{L^+}:= \big[ k_{\mc X}(\underline \bx_\ell,\bx_j')\big]_{\ell=1,j=1}^{L,N} \in\R^{L\times N}$. Using the conditional weights $\boldsymbol\Gamma(\bx)$ from \eqref{eq:Gamma}, the target-policy continuation vector has the plug-in form
\begin{align}
\widehat\bso^{\pi}(\bx;\mathbb B)
&=
\sum_{j=1}^{N}
\Gamma_j(\bx)\widehat\eta_j
\mathbb B^\top\bk_L(\bx_j')
=\mathbb B^\top\mathbb K_{L^+}\mathbb D_\eta\boldsymbol\Gamma(\bx)
:=\mathbb B^\top\boldsymbol\Phi_L(\bx)
\label{eq:Phi}
\end{align}
where $\boldsymbol\Phi_L(\bx)
:=\mathbb K_{L^+}\mathbb D_\eta\boldsymbol\Gamma(\bx)\in\R^L$. The full derivation of \eqref{eq:Phi} is in Supplementary Material~\ref{app:Off_policy_IS}.

\vspace{-1em}
\begin{remark}
The state-action kernel used in $\mathbb K_{\mc X}$, $\mathbb K_L$, and $\mathbb K_{L^+}$ need not match the return-space Mat\'{e}rn kernel. Equation~\eqref{eq:Phi} treats the density ratio as scalar importance weights, so it does not require the product $\eta(\cdot)\omega_i(\cdot)$ to lie in the state-action RKHS.
\end{remark}
\subsection{Finite-Dictionary Bellman Criterion and Numerical Optimization}
\label{sec:optimization}
\label{sec:global_ke_loss}

The population Bellman criterion in \eqref{eq:exact_population_loss} is defined over unrestricted conditional return laws and depends on the joint one-step target-policy law of $(\bR,\bX')\mid\bX=\bx$. KE-DRL replaces this population criterion with a finite-dimensional Bellman embedding criterion. For a fixed target policy $\pi$, the candidate conditional return embedding is represented by a single policy-specific coefficient matrix $\mathbb B\in\R^{L\times m}$ as in \eqref{eq:mu_Z} and \eqref{eq:hat_w}. The rows of $\mathbb B$ correspond to the selected state-action dictionary $\mc C_L=\{\underline \bx_\ell\}_{\ell=1}^L$, while the columns correspond to the return-grid atoms $\mathbf Z_m=\{\bz_i\}_{i=1}^m$. Thus $L$ controls the complexity of the state-action coefficient functions and $m$ controls the numerical resolution of the return-space expansion. Once estimated, the single matrix $\mathbb B$ is used for all evaluation points rather than being re-estimated at each $\bx$. Accordingly, the finite-dimensional criterion averages the pointwise residual over a chosen conditioning distribution and adds the coefficient-function ridge $\mathfrak R_B(\mathbb B)=\mathrm{tr}(\mathbb B^\top\mathbb K_L\mathbb B)$. Supplementary Material~\ref{app:population_finite_criterion} gives the population criterion and the corresponding RKHS norm identity.

\noindent In finite samples, choose conditioning points $\mc X_\star=\{\bx_q^\star\}_{q=1}^{M}\subseteq\mc X$ and nonnegative weights $\varrho_q$ satisfying $\sum_{q=1}^{M}\varrho_q=1$. We take $\varrho_q=1/M$ unless nonuniform weights are prespecified. The conditioning points determine where the projected Bellman equation is enforced; they may be the full training support, a random or clustered subset, or another design over the empirical support. The conditioning-set size $M$ is distinct from both the return-grid size $m$ and the state-action coefficient-basis size $L$. When all pooled transitions are used as conditioning points, $M=N$. The reported analyses may instead use a subset, as described in Section~\ref{sec:Alg_implementation}.
\noindent For each $\bx_q^\star\in\mc X_\star$, define $\bk_{L,q}:=\bk_L(\bx_q^\star),\ \boldsymbol\Phi_{L,q}:=\boldsymbol\Phi_L(\bx_q^\star),\ \mathbb H_q:= \widehat{\mathbb H}(\bx_q^\star)$, and $\  \mathbb G_q:=\widehat{\mathbb G}(\bx_q^\star)$ where $\boldsymbol\Phi_L(\bx)$ is defined in \eqref{eq:Phi}, and $\widehat{\mathbb H}(\bx)$ and $\widehat{\mathbb G}(\bx)$ are given by \eqref{eq:h_ij} and \eqref{eq:g_ij}. Define the current and target-policy continuation coefficient vectors by $u_q(\mathbb B):=\mathbb B^\top\bk_{L,q},\ v_q(\mathbb B):=\mathbb B^\top\boldsymbol\Phi_{L,q}$. Here $u_q(\mathbb B)$ is the candidate return-grid coefficient vector at the current input $\bx_q^\star$, while $v_q(\mathbb B)$ is the off-policy continuation coefficient vector under the target policy. Using \eqref{eq:gamma_with_W}, the empirical pointwise Bellman residual is
\begin{align*}
\widehat\Bgamma_{k,q}^{\,2}(\mathbb B)
&=
u_q(\mathbb B)^\top
\mathbb K_{\mc Z}
u_q(\mathbb B)
-
2u_q(\mathbb B)^\top
\mathbb H_q
v_q(\mathbb B)
+
v_q(\mathbb B)^\top
\mathbb G_q
v_q(\mathbb B)
\end{align*}

\noindent An exact minimizer of the regularized finite-grid criterion is defined by
\begin{align}
\widehat{\mathbb B}_{\mathrm{opt}}^{\pi}
&\in
\argmin_{\mathbb B\in\R^{L\times m}}
\widehat{\mc L}_{\mathrm{impl}}(\mathbb B)
\nonumber\\
\quad\widehat{\mc L}_{\mathrm{impl}}(\mathbb B)
&:=
\sum_{q=1}^{M}\varrho_q\widehat\Bgamma_{k,q}^{\,2}(\mathbb B)
+
\lambda_B\mathfrak{R}_B(\mathbb B) +
\lambda_{\mathrm{mass}}
\sum_{q=1}^{M}\varrho_q
\left\{\mathbf 1_m^\top u_q(\mathbb B)-c_{\mathrm{mass}}\right\}^2
+
\lambda_{\mathrm{neg}}
\sum_{q=1}^{M}\varrho_q
\left\|\{u_q(\mathbb B)\}_{-}\right\|_2^2
\label{eq:global_B_objective}
\end{align}
Here $\mathbf 1_m$ is the vector of ones, and $c_{\mathrm{mass}}$ is the target finite-grid coefficient mass, equal to one in the reported analyses. For $a\in\R^m$, $\{a\}_{-}:=(\max\{-a_i,0\})_{i=1}^m$ denotes the componentwise negative part. The third and fourth terms in \eqref{eq:global_B_objective} are the coefficient-mass and negative-part penalties, respectively. Setting $\lambda_{\mathrm{mass}}=\lambda_{\mathrm{neg}}=0$ gives the empirical finite-grid counterpart of the regularized Bellman criterion in \eqref{eq:population_ke_loss}.
Supplementary Material~\ref{app:population_finite_criterion} records a positive-semidefiniteness condition on the operator blocks under which the criterion is convex in $\mathbb B$. The coefficient-mass term excludes the zero candidate from attaining zero penalty when $\lambda_{\mathrm{mass}}>0$ and $c_{\mathrm{mass}}\ne0$. This term is needed because the induced Bellman embedding operator acts linearly on embeddings of finite measures. Consequently, $\mathbb B=\mathbf 0$ makes every exact finite-sample Bellman residual and the coefficient ridge equal to zero and is a minimizer of the criterion without the coefficient-mass penalty when the operator blocks are constructed from the same exact kernel sections or from a common finite feature map.

\noindent Neither the coefficient-mass penalty nor the optional negative-part penalty is a simplex constraint. Both are soft penalties evaluated only at the conditioning points $\mc X_\star$, so they do not make the finite-dictionary candidate a probability law at training or evaluation inputs. Setting $\lambda_{\mathrm{neg}}=0$ leaves the signed coefficient class unchanged. Fitted coefficients may be negative and should be interpreted as RKHS coordinates rather than probabilities; probability summaries are obtained only through regularized post-estimation recovery in Section~\ref{sec:stat_recovery}. Moreover, the residual bound in Theorem~\ref{thm:fixed_point} applies to candidates corresponding to laws in the stated class $K$. A small projected Bellman residual for an unrestricted signed coefficient matrix therefore indicates close empirical agreement with the finite-dimensional projected Bellman equation; it does not itself provide a bound on Wasserstein error.
\section{Population Fixed-Point Identification and Conditional-Embedding Error Bounds}\label{sec:theory}
\noindent The analysis distinguishes three levels. First, Theorem~\ref{thm:fixed_point} is a population result for conditional law maps represented in the class $K$; it establishes fixed-point identification and residual-to-$W_1$ control on the stated regular class. Second, Theorem~\ref{thm:pointwise_error_bound_intro} and its extensions in Supplementary Material~\ref{app:tikhonov_extensions} analyze a regularized conditional mean embedding estimator and its Bellman-target counterpart when the required input-output pairs are available. Third, Algorithm~\ref{alg:rk_drl} computes the finite-iteration estimate $\widehat{\mathbb B}^{\pi}$ from offline transitions by numerically optimizing a finite-grid projected Bellman criterion with a state-action basis. The results for the regularized conditional mean embedding estimator do not establish consistency or an end-to-end error bound for $\widehat{\mathbb B}^{\pi}$.

\subsection{Population Fixed-Point Identification}
The distributional Bellman operator is a Banach contraction under Wasserstein distance \citep{rosler1992fixed,bellemare2017distributional}. Here, return laws are compared through an integral probability metric induced by an RKHS on the return space. \footnote{When the test-function class is the unit ball of an RKHS with kernel $k$, the resulting discrepancy $\Bgamma_k$ is the maximum mean discrepancy.} The relevant question is whether this discrepancy identifies the same target-policy return law as the Wasserstein contraction. On the regular distribution class used below, $\Bgamma_k$ and $W_1$ induce the same weak topology, and $W_1$ is bounded by a H\"older power of $\Bgamma_k$ \citep{vayer2023controlling}; Supplementary Material~\ref{app:Wass_Control_MMD} states the comparison precisely. Accordingly, we use a Mat\'ern return kernel because it satisfies the Fourier-decay conditions required for the MMD-to-$W_1$ comparison and is translation invariant, which simplifies the embedding calculations. The following theorem shows that the induced discrepancy identifies the same fixed point as the Wasserstein Bellman contraction.

\begin{assumption}[Regular Bellman embedding class]\label{ass:regular_bellman_embedding_class}
Let $\mathfrak S_{b,\mathfrak m,r,s}$ be the class of laws $\mathbb P=f\,d\bz$ satisfying $\|f\|_{H^s(\R^d)}\le b$ and $\{\E_{\mathbb P}\|\bZ\|_2^r\}^{1/r}\le\mathfrak m$, as defined in Theorem~\ref{thm:mmd_wasserstein_equiv}. Here $r>1$ and $s\ge s_k/2$, where $s_k$ is the Fourier-decay order of the return kernel in condition~\ref{cond:Fourier_Decay} ($s_k=2\nu+d$ for the Mat\'ern kernel with smoothness $\nu$; see Supplementary Material~\ref{app:matern_verification}). For conditional law maps, write
\vspace{-1em}\[
D_k(\mathbb P,\mathbb Q)
:=
\sup_{\bx\in\mc X}
\Bgamma_k\{\mathbb P(\cdot\mid\bx),\mathbb Q(\cdot\mid\bx)\}\]
There exists a nonempty, norm-closed, bounded, convex set $K$ of admissible conditional mean-embedding maps in a reflexive Banach space $\mc E$ such that:
\begin{enumerate}[label=(\roman*)]
\vspace{-1em}\item every $\mu\in K$ corresponds to a measurable conditional return-law map $\mathbb P_\mu(\cdot\mid\bx)\in\mathfrak S_{b,\mathfrak m,r,s}$ for every $\bx\in\mc X$;
\vspace{-1em}\item the induced Bellman embedding operator $\mathfrak T^\pi\mu:=\mu_{\T^\pi\mathbb P_\mu}$ maps $K$ into itself. In the vector-valued RKHS template of Supplementary Material~\ref{app:vvRKHS_example}, this requires preservation of both the Sobolev-moment law class and the ambient $\mc E$-ball. Lemma~\ref{lem:invariance_identification} gives sufficient conditions for preservation of the law class, while preservation of the ambient ball remains an ambient regularity condition;
\vspace{-1em}\item on $K$, weak convergence in $\mc E$ implies pointwise weak convergence in $\mc H_{\mc Z}$, and the relative weak topology is generated by the pointwise dual evaluations used in Lemma~\ref{lemma:weak_cts_embed_operator}.
\end{enumerate}
\end{assumption}
\noindent Unlike a Banach-contraction argument, Assumption~\ref{ass:regular_bellman_embedding_class} does not impose metric compatibility between the $\mc E$-norm and the embedding discrepancy $D_k$. Existence follows from weak compactness of $K$ (norm-closed, bounded, convex subsets of reflexive spaces are weakly compact) and weak sequential continuity of $\mathfrak T^\pi$. The quantitative conclusions are stated in terms of $D_k$ and the conditional Wasserstein metric. Supplementary Material~\ref{app:vvRKHS_example} provides a conditional vector-valued RKHS construction for $(\mc E,K)$ in which nonemptiness and clauses (i) and (iii) follow from an explicit compatibility condition. It also separates the law-class and ambient-space requirements in clause (ii).
\begin{theorem}[Fixed Point of the Distributional Bellman Operator]
\label{thm:fixed_point}
Let $\mathcal{T}^\pi$ be the distributional Bellman operator under policy $\pi$, and let $\gamma \in (0,1)$ be the discount factor. Let $k_{\mc Z}$ be a $\mbox{Mat\'{e}rn}$ kernel on the $d$-dimensional return space with smoothness parameter $\nu > 1$, length-scale $\ell > 0$, variance $\sigma^2 > 0$, and feature-map Lipschitz constant
\vspace{-1em}
$$L_k=\frac{\sigma}{\ell}\sqrt{\frac{\nu}{\nu-1}}$$
Set
\vspace{-1em}
$$\rho=\frac{r-1}{d+2r}\in(0,1)$$
and let $C_*$ be the constant in the upper MMD-Wasserstein comparison of Theorem~\ref{thm:mmd_wasserstein_equiv} on $\mathfrak S_{b,\mathfrak m,r,s}$. Suppose Assumption~\ref{ass:regular_bellman_embedding_class} holds. For conditional law maps $\mathbb P,\mathbb Q$ represented by elements of $K$, write $D_k(\mathbb P,\mathbb Q)$ for the maximal embedding discrepancy in Assumption~\ref{ass:regular_bellman_embedding_class}. Then the following hold.
\begin{enumerate}[label=(\alph*)]
\vspace{-1em}\item \textit{H\"older bound.} The induced Bellman operator satisfies
\vspace{-1em}
\[
D_k(\T^\pi\mathbb P,\T^\pi\mathbb Q)
\le
\gamma L_k C_*\,D_k(\mathbb P,\mathbb Q)^\rho \]
\vspace{-4em}
\item \textit{Existence.} The operator $\mathfrak T^\pi$ has at least one fixed point $\mu^\pi\in K$, and the corresponding conditional law $\mathbb P^\pi$ satisfies $\T^\pi\mathbb P^\pi=\mathbb P^\pi$.
\vspace{-1em}\item \textit{Uniqueness.} Equipped with the maximal conditional metric $d_1(\mathbb P,\mathbb Q):=\sup_{\bx\in\mc X}W_1\{\mathbb P(\cdot\mid\bx),\mathbb Q(\cdot\mid\bx)\}$, which is finite on the regular class by the uniform moment bound, the vector-valued Wasserstein coupling contraction in the proof gives $d_1(\T^\pi\mathbb P,\T^\pi\mathbb Q)\le\gamma\,d_1(\mathbb P,\mathbb Q)$; hence the fixed point is unique among conditional law maps in the regular class.
\vspace{-1em}\item \textit{Residual bound.} For any $\mathbb Q$ represented in $K$ such that $\T^\pi\mathbb Q$ also belongs to the same regular class, the maximal conditional $W_1$ distance between $\mathbb Q$ and the unique fixed-point law $\mathbb P^\pi$ of parts (b) and (c) satisfies
\[
d_1(\mathbb Q,\mathbb P^\pi)
\le
\frac{C_*}{1-\gamma}
\left[
\sup_{\bx\in\mc X}
\Bgamma_k\{\mathbb Q(\cdot\mid\bx),(\T^\pi\mathbb Q)(\cdot\mid\bx)\}
\right]^\rho \]
\end{enumerate}
Furthermore, under the reward conditions of Lemma~\ref{lem:invariance_identification} in Supplementary Material~\ref{app:invariance_identification}, the true conditional return law $\mc V^\pi_{\bx}=\mc L\{\bZ^\pi(\bx)\}$ of \eqref{eq:return_def_new} belongs to $\mathfrak S_{b,\mathfrak m,r,s}$ for every $\bx$, and the fixed point identified above coincides with it: $\mathbb P^\pi(\cdot\mid\bx)=\mc V^\pi_{\bx}$ for every $\bx\in\mc X$.
\end{theorem}

\noindent Theorem~\ref{thm:fixed_point} is the population justification for replacing the intractable Wasserstein criterion with the computable embedding criterion. Part (a) shows that the Bellman operator is H\"older-continuous, rather than contractive, in the embedding discrepancy; the contraction that drives uniqueness in part (c) operates in the Wasserstein metric, so the two geometries play complementary roles. Part (d) is the operational conclusion: for any candidate law map in the regular class, the maximal conditional $W_1$ distance to the true return law is controlled by a power of the population embedding Bellman residual, which is precisely the quantity that the KE-DRL criterion of Section~\ref{sec:optimization} minimizes and the held-out diagnostic of Section~\ref{sec:evaluation_metric} reports. Driving the embedding residual toward zero therefore drives a law-valued candidate toward the true conditional return law in Wasserstein distance, at a rate governed by the exponent $\rho=(r-1)/(d+2r)$, which weakens as the return dimension $d$ grows or the available moment order $r$ decreases. Because the bound applies to candidates representing laws in $K$, the empirical residual of an unconstrained signed candidate remains a model-fit diagnostic rather than a certified Wasserstein error bound, as noted after \eqref{eq:global_B_objective}.

\noindent The theorem applies to absolutely continuous return laws with the stated Sobolev regularity and moment bounds. Parts (b) and (c) identify the population law, and part (d) converts a population embedding residual into Wasserstein control only for law-valued candidates represented in $K$. The embedding construction and finite-dictionary approximation remain well defined for more general laws, including mixed or atomic laws, but the MMD-to-Wasserstein bound and the fixed-point theorem are not asserted outside this regular class. Supplementary Theorem~\ref{thm:DBO-pdf} gives a density representation of the Bellman update under additional density, joint-measurability, and conditional-independence conditions.

\noindent Gaussian kernels on $\mathbb{R}^d$ do not satisfy the Fourier-decay condition~\ref{cond:Fourier_Decay} used in Theorem~\ref{thm:mmd_wasserstein_equiv}, whereas Mat\'ern kernels do. The Mat\'ern kernel is
\[
k(\bz,\bz')
=
\sigma^2\frac{2^{1-\nu}}{\Gamma(\nu)}
\left(\sqrt{2\nu}\frac{\|\bz-\bz'\|_2}{\ell}\right)^\nu
\mathcal K_\nu\left(\sqrt{2\nu}\frac{\|\bz-\bz'\|_2}{\ell}\right)
\]
where $\nu$ controls smoothness, $\ell$ is the length scale, $\mathcal K_\nu$ is the modified Bessel function of the second kind, and $\sigma^2$ is the marginal variance.
Supplementary Material~\ref{app:matern_verification} verifies the Fourier-decay and feature-map Lipschitz conditions and includes an illustration of the effects of $\nu$ and $\ell$.
\subsection{Error Bounds for a Regularized Conditional Mean Embedding Estimator}
\label{sec:tikhonov_reference}
The analysis is stated for pairs $\{(\bX_i,\bZ_i)\}_{i=1}^n$ in which $\bZ_i$, or its Bellman-target counterpart, is observed; here $n$ denotes the number of these pairs, not the number of trajectories. We use the state-action input $\bX\in\mc X=\mc S\times\mc A$ to avoid switching between $\bX$ and $(\bS,\bA)$. Throughout this subsection, $k_{\mc Z}$ is a bounded Mat\'ern kernel and $ k_{\mc X}$ is a bounded measurable kernel on $\mc X$, with separable RKHSs $\mc H_{\mc Z}$ and $\mc H_{\mc X}$; boundedness gives the finite constants $\kappa_{\mc Z}^2:=\sup_{\bz\in\mc Z}k_{\mc Z}(\bz,\bz)$ and $\kappa_{\mc X}^2:=\sup_{\bx\in\mc X} k_{\mc X}(\bx,\bx)$. Following \citet{song2009hilbert,zhang2011kernel,grunewalder2012modelling}, write $\mbC_{\mc{ZX}}:\mc H_{\mc X}\to\mc H_{\mc Z}$ for the uncentered cross-covariance operator and $\mbC_{\mc{XZ}}:=\mbC_{\mc{ZX}}^*$. Under the well-specifiedness, bounded-extension, and source conditions of Theorem~\ref{thm:pointwise_error_bound_intro}, the population target satisfies
\[
\mu_{\bZ\mid\bX=\bx}
=
\lim_{\lambda\downarrow0}
\mbC_{\mc{ZX}}(\mbC_{\mc{XX}}+\lambda\mathbb I)^{-1}
k_{\mc X}(\bx,\cdot)
\quad\text{in }\mc H_{\mc Z}
\]
When $k_{\mc X}(\bx,\cdot)\in\operatorname{Dom}(\mbC_{\mc{XX}}^\dagger)$, this identity may also be written as $\mu_{\bZ\mid\bX=\bx}=\mbC_{\mc{ZX}}\mbC_{\mc{XX}}^\dagger k_{\mc X}(\bx,\cdot)$. Let $\widehat{\mbC}_{\mc{XX}}$ and $\widehat{\mbC}_{\mc{ZX}}$ denote the $n^{-1}$-scaled empirical input covariance and output-input cross-covariance operators. For $\lambda>0$, define the regularized conditional mean embedding estimator by $\widehat\mu_{\bZ\mid\bx}:=\widehat{\mbC}_{\mc{ZX}}(\widehat{\mbC}_{\mc{XX}}+\lambda\mathbb I)^{-1} k_{\mc X}(\bx,\cdot)$. Its Gram representation is $\widehat\mu_{\bZ\mid\bx}=\sum_{i=1}^n\beta_i(\bx)k_{\mc Z}(\bZ_i,\cdot)$, where $\boldsymbol\beta(\bx)=(\mathbb K_{\mc X}+n\lambda\mathbb I_n)^{-1}\widetilde{\bk}_{\mc X}^{n}(\bx)$. Because $\boldsymbol\beta(\bx)$ depends only on the observed inputs, the same weight vector applies to any response observed at those inputs. Within this subsection, $\widehat\mu_{\bZ\mid\bx}$ denotes this estimator, which uses the required input-output pairs directly, with operator-level Tikhonov regularization and no finite-dictionary, density-ratio, or random-feature approximation; it is distinct from the finite-dictionary candidate $\widehat\mu_{\bZ\mid\bx}(\cdot\ ;\mathbb B)$.

\begin{theorem}[Pointwise Conditional-Embedding Error Bound]\label{thm:pointwise_error_bound_intro}\label{thm:Error_bound}
Fix $\bx\in\mc X$ and let $\{(\bX_i,\bZ_i)\}_{i=1}^n$ be a possibly dependent sample. Assume the standard conditional-embedding well-specifiedness condition: for every $f\in\mc H_{\mc Z}$, the conditional mean function $m_f(\cdot):=\E\{f(\bZ)\mid\bX=\cdot\}$ admits a version in $\mc H_{\mc X}$ point-wise for each $x$. Assume also that $\mbC_{\mc{ZX}}\mbC_{\mc{XX}}^{-1/2}$ extends to a bounded operator and that the source condition holds: for some $g_{\bx}\in\mc H_{\mc X}$ and $0<\tau\le1/2$, $ k_{\mc X}(\bx,\cdot)=\mbC_{\mc{XX}}^{1/2+\tau}g_{\bx}$. If, with probability at least $1-\delta$, the empirical covariance operators satisfy
\[
\|\widehat{\mbC}_{\mc{ZX}}-\mbC_{\mc{ZX}}\|_{\mathrm{HS}}\le\varepsilon_{\mc{ZX}}(n,\delta),
\qquad
\|\widehat{\mbC}_{\mc{XX}}-\mbC_{\mc{XX}}\|_{\mathrm{op}}\le\varepsilon_{\mc{XX}}(n,\delta)
\]
(Supplementary Material~\ref{app:mixing_concentration} provides concentration radii of this form for geometrically mixing sequences), then
\[
\|\widehat\mu_{\bZ\mid\bx}-\mu_{\bZ\mid\bx}\|_{\mc H_{\mc Z}}
\le
\frac{\kappa_{\mc X}}{\lambda}\varepsilon_{\mc{ZX}}(n,\delta)
+\frac{\kappa_{\mc X}^2\kappa_{\mc Z}}{\lambda^2}\varepsilon_{\mc{XX}}(n,\delta)
+c_{\bx}\lambda^\tau
\]
where $c_{\bx}=\|\mbC_{\mc{ZX}}\mbC_{\mc{XX}}^{-1/2}\|_{\mathrm{op}}\|g_{\bx}\|_{\mc H_{\mc X}}$.
\end{theorem}

\noindent The theorem separates stochastic covariance error from deterministic regularization bias and requires only the displayed covariance-concentration events. Temporal dependence and the sampling design enter through the radii $\varepsilon_{\mc{ZX}}(n,\delta)$ and $\varepsilon_{\mc{XX}}(n,\delta)$. Supplementary Material~\ref{app:mixing_concentration} provides explicit radii for a single geometrically mixing sequence as one sufficient design; other designs, including collections of independent trajectories, enter through their corresponding covariance radii. Thus the pointwise rate has the form $\varepsilon_{\mc{ZX}}/\lambda+\varepsilon_{\mc{XX}}/\lambda^2+\lambda^\tau$, matching regularized conditional-embedding analyses such as \citet{fukumizu2013kernel}. The bound makes the role of the regularization parameter explicit. Decreasing $\lambda$ reduces the regularization bias $c_{\bx}\lambda^\tau$ but inflates the stochastic terms. For covariance radii of the parametric order $n^{-1/2}$, balancing the dominant stochastic term against the bias term suggests $\lambda\asymp n^{-1/\{2(\tau+2)\}}$ and yields a pointwise rate of order $n^{-\tau/\{2(\tau+2)\}}$; smoother targets, corresponding to larger source exponents $\tau$, tolerate smaller regularization and give faster rates.

\noindent Because the Bellman criterion compares two estimated embeddings, their pointwise RKHS-norm estimation errors enter additively in the perturbation bound for the squared Bellman residual.

\begin{corollary}[Bellman-residual error bound]\label{cor:bellman_residual_error}
Suppose, in addition to the conditions of Theorem~\ref{thm:pointwise_error_bound_intro}, that $(\bX,(\T^\pi\bZ)(\bX))$ satisfies the corresponding well-specifiedness, boundedness, source, and covariance-concentration conditions, with radius $\varepsilon_{\T\mc Z,\mc X}(n,\delta)$ and the same $\varepsilon_{\mc{XX}}(n,\delta)$. Assume the two covariance events hold jointly with probability at least $1-\delta$. Then
\[
\left|
\|\widehat{\mu}_{\bZ\mid\bx}-\widehat{\mu}_{\T^\pi\bZ\mid\bx}\|_{\mc H_{\mc Z}}^2
-
\|\mu_{\bZ\mid\bx}-\mu_{\T^\pi\bZ\mid\bx}\|_{\mc H_{\mc Z}}^2
\right|
\le
4\kappa_{\mc Z}\{\mathfrak B_{\bZ}+\mathfrak B_{\T^\pi\bZ}\}
+\{\mathfrak B_{\bZ}+\mathfrak B_{\T^\pi\bZ}\}^2
\]
Here $\mathfrak B_{\bZ}$ and $\mathfrak B_{\T^\pi\bZ}$ denote the pointwise RKHS-norm error bounds for estimating the conditional embeddings of $\bZ$ and $\T^\pi\bZ$, respectively. The former is the right-hand side of Theorem~\ref{thm:pointwise_error_bound_intro}, and the latter is its Bellman-target analogue. Each combines covariance-estimation and regularization-bias terms; consequently, the displayed bound vanishes whenever both component error bounds vanish.
\end{corollary}

\noindent Supplementary Material~\ref{app:tikhonov_extensions} gives the corresponding pointwise-consistency, uniform error, and strong-consistency extensions, which share the stochastic-bias decomposition of Theorem~\ref{thm:pointwise_error_bound_intro}.

\section{Implementation, Diagnostics, and Recovery}\label{sec:implementation}
\noindent This section turns the criterion of Section~\ref{sec:optimization} into a complete procedure. Section~\ref{sec:Alg_implementation} states the estimation algorithm and describes how the fitted embedding is evaluated at new inputs. Section~\ref{sec:evaluation_metric} introduces a held-out projected Bellman residual for model assessment, and Section~\ref{sec:stat_recovery} delineates which functionals can be read directly from the fitted embedding and which require regularized post-estimation recovery.
\subsection{Estimation Algorithm and Embedding Evaluation}\label{sec:Alg_implementation}
\noindent Under Assumption~\ref{ass:reward_continuation_separation}, Algorithm~\ref{alg:rk_drl} summarizes the separated empirical procedure used in the reported analyses. Conditional-embedding and reward-shifted operator quantities are precomputed for the conditioning set, while numerical optimization updates the matrix $\mathbb B\in\R^{L\times m}$. For large return grids or many conditioning points, the matrices $\mathbb H_q$ and $\mathbb G_q$ may be constructed either from direct kernel evaluations or through random Fourier features \citep{rahimi2007random} for the Mat\'ern return kernel. When random features are used, $\mathbb K_{\mc Z}$, $\mathbb H_q$, and $\mathbb G_q$ are all constructed from the same feature map. The resulting Bellman term is therefore the finite-feature approximation to the same projected Bellman residual.
\begin{algorithm}[H]
\footnotesize
\caption{KE-DRL finite-dictionary embedding estimator}
\label{alg:rk_drl}
\begin{algorithmic}[1]
\State \textbf{Input:} Historical trajectories $\mathcal D$, pooled transitions $\{(\bx_j,\br_j,\bx_j')\}_{j=1}^{N}$, target policy $\pi$, state-action basis $\mc C_L=\{\underline \bx_\ell\}_{\ell=1}^{L}$, conditioning set $\mc X_\star=\{\bx_q^\star\}_{q=1}^{M}$, discount factor $\gamma$, return grid $\mathbf Z_m=\{\bz_i\}_{i=1}^{m}$, kernels $( k_{\mc X},k_{\mc Z})$, regularization parameters $(\lambda_{\Gamma},\lambda_B,\lambda_{\mathrm{mass}},\lambda_{\mathrm{neg}})$, target mass $c_{\mathrm{mass}}$, optimizer settings

\State \textbf{Precomputation:}
\State \hspace{1em} Compute $\mathbb K_{\mc X}=[ k_{\mc X}(\bx_j,\bx_r)]_{j,r=1}^{N}$, $\mathbb K_L=[ k_{\mc X}(\underline \bx_\ell,\underline \bx_r)]_{\ell,r=1}^{L}$, $\mathbb K_{L^+}=[ k_{\mc X}(\underline \bx_\ell,\bx_j')]_{\ell=1,j=1}^{L,N}$, and $\mathbb K_{\mc Z}=[k_{\mc Z}(\bz_i,\bz_j)]_{i,j=1}^{m}$
\State \hspace{1em} Estimate density ratios $\widehat\eta_j=\widehat\eta(\bx_j')$ and set $\mathbb D_\eta=\operatorname{diag}(\widehat\eta_1,\ldots,\widehat\eta_N)$ \hfill \texttt{(See Algorithm \ref{alg:ulsif})}

\State \textbf{Auxiliary Operators for the Conditioning Set:}
\For{$q=1,\ldots,M$}
\State \hspace{1em} $\bk_{L,q}\leftarrow\bk_L(\bx_q^\star)$
\State \hspace{1em} $\widetilde\bk_{N,q}\leftarrow( k_{\mc X}(\bx_1,\bx_q^\star),\ldots, k_{\mc X}(\bx_N,\bx_q^\star))^\top$
\State \hspace{1em} $\boldsymbol\Gamma_q\leftarrow(\mathbb K_{\mc X}+N\lambda_\Gamma\mathbb I_N)^{-1}\widetilde\bk_{N,q}$
\State \hspace{1em} $\mathbb H_q\leftarrow \tx{H}(\boldsymbol\Gamma_q,\mathbf Z_m,\gamma,\br,k_{\mc Z})$ by exact kernels or shared random features \hfill \texttt{(See Equation \eqref{eq:h_ij})}
\State \hspace{1em} $\mathbb G_q\leftarrow \tx{G}(\boldsymbol\Gamma_q,\mathbf Z_m,\gamma,\br,k_{\mc Z})$ by exact kernels or shared random features \hfill \texttt{(See Equation \eqref{eq:g_ij})}
\State \hspace{1em} $\boldsymbol\Phi_{L,q}\leftarrow\mathbb K_{L^+}\mathbb D_\eta\boldsymbol\Gamma_q$ \hfill \texttt{(See Equation \eqref{eq:Phi})}
\EndFor

\State \textbf{Optimization Step:}
\State \hspace{1em} $\widehat{\mathbb B}^{\pi}\leftarrow\operatorname{Optimize}\bigl(\{\bk_{L,q},\mathbb G_q,\mathbb H_q,\boldsymbol\Phi_{L,q},\varrho_q\}_{q=1}^{M},\mathbb K_L,\lambda_B,\lambda_{\mathrm{mass}},\lambda_{\mathrm{neg}},c_{\mathrm{mass}}\bigr)$ \hfill \texttt{(See Equation \eqref{eq:global_B_objective})}

\State \textbf{Return} $\widehat{\mathbb B}^{\pi}$ and the fitted map $\bx\mapsto(\widehat{\mathbb B}^{\pi})^\top\bk_L(\bx)$
\end{algorithmic}
\end{algorithm}

\noindent The reported implementation uses AdamW \citep{loshchilov2019decoupled} with automatic differentiation. The reported estimate $\widehat{\mathbb B}^{\pi}$ is the AdamW iterate with the smallest value of \eqref{eq:global_B_objective}, evaluated over the full conditioning set, among the prespecified evaluation iterations; it is distinct from the exact minimizer $\widehat{\mathbb B}_{\mathrm{opt}}^{\pi}$. Supplementary Material~\ref{app:optimizer_details} gives the stopping rule, mini-batch construction, numerical stabilization, and analysis-specific numerical settings.

\noindent After fitting $\widehat{\mathbb B}^{\pi}$, the estimated return embedding conditional on any state-action input $\bx_0$ is
\begin{equation}\label{eq:final_embedding_prediction}
\widehat\bso(\bx_0)
=
(\widehat{\mathbb B}^{\pi})^\top\bk_L(\bx_0)
\qquad
\widehat\mu^{\pi}(\bx_0)(\cdot)
=
\widehat\bso(\bx_0)^\top
\bk_{\mc Z}(\cdot)
\end{equation}
Equation~\eqref{eq:final_embedding_prediction} defines an element of $\mc H_{\mc Z}$. Its coefficient vector is not a vector of atom probabilities and may contain negative entries. The embedding can be used directly for RKHS test-function evaluations and other smooth functionals, whereas probability summaries require the separate recovery procedure in Section~\ref{sec:stat_recovery}.
For diagnostics that evaluate the one-step target-policy continuation at $\bx_0$, the same fitted matrix can also be combined with $\boldsymbol\Phi_L(\bx_0)$. This distinction is computational rather than conceptual. Both calculations use the same jointly estimated matrix $\widehat{\mathbb B}^{\pi}$.

\subsection{Held-out-Input Projected Bellman Residual}\label{sec:evaluation_metric}
Target-policy returns are not observed, so we assess the fitted map through a plug-in projected Bellman residual evaluated at held-out state-action inputs. The operator summaries and density-ratio estimate remain those obtained from the training transition sample. Let $\mc X_{\mathrm{test}}=\{\bx_j^{\mathrm{test}}\}_{j=1}^{N_{\mathrm{test}}}$ denote these inputs. At each held-out input, the corresponding training-sample operator estimates are evaluated while $\widehat{\mathbb B}^{\pi}$ remains fixed. The held-out-input projected Bellman risk is
\begin{align}
\widehat{\mc R}_{\mathrm{PB}}(\widehat{\mathbb B}^{\pi})
:=
\frac{1}{N_{\mathrm{test}}}
\sum_{j=1}^{N_{\mathrm{test}}}
\Big[
&\widehat u_j^\top \mathbb K_{\mc Z}\widehat u_j
-2\widehat u_j^\top\mathbb H_{j}^{\mathrm{test}}\widehat v_j
+\widehat v_j^\top\mathbb G_{j}^{\mathrm{test}}\widehat v_j
\Big]
\label{eq:generalization_metric}
\end{align}
where $\widehat u_j=(\widehat{\mathbb B}^{\pi})^\top\bk_L(\bx_j^{\mathrm{test}})$ and $\widehat v_j=(\widehat{\mathbb B}^{\pi})^\top\boldsymbol\Phi_L(\bx_j^{\mathrm{test}})$. This quantity is the squared finite-sample residual induced by the separated plug-in Bellman target embedding. Under Assumption~\ref{ass:reward_continuation_separation}, the corresponding separated population target embedding equals the paired finite-dictionary Bellman target embedding. Smaller values indicate closer fit to the projected Bellman equation on the held-out inputs, and a value of zero corresponds to exact satisfaction of that equation within the finite-dimensional representation. The quantity is not a scalar policy value and should not be interpreted as revenue, welfare, or causal performance.
Supplementary Material~\ref{app:heldout_residual_details} gives the complete test-point construction and explains why this risk differs from the same-dictionary expression that omits the reward shift.
\subsection{Embedding Functionals and Post-Estimation Recovery}\label{sec:stat_recovery}

The population conditional mean embedding represents expectations of RKHS test functions. The fitted finite-dictionary element may have signed coefficients, so it instead induces the embedding-based plug-in functional
\[
g\longmapsto
\left\langle g,\widehat\mu^{\pi}(\bx)\right\rangle_{\mc H_{\mc Z}}
=
\sum_{u=1}^m \widehat\omega_u(\bx)g(\bz_u),
\qquad g\in\mc H_{\mc Z}
\]
The functional need not be an expectation under a probability law unless a probability-valued recovery has been performed. Smooth utilities and smoothed tail summaries can be evaluated directly through it. Exact indicators, unsmoothed densities, and other sharp features require a regularized inversion. Supplementary Material~\ref{app:emb_ill_posed} gives the associated stability and smoothing-bias bounds, and Supplementary Material~\ref{subsec:expedia_density_recovery} gives the application-specific recovery construction.
\section{Simulation Study}\label{sec:sim}

We assess KE-DRL in an offline off-policy evaluation setting in which the behavior policy that generated the logged data is unknown during estimation. The estimand is the conditional kernel mean embedding of the discounted-return distribution under a fixed target policy, evaluated at state-action points that are not selected during data collection. The study assesses pointwise agreement with Monte Carlo benchmarks over specified evaluation grids. We use Gaussian, Uniform, and Logistic policy families and pair each behavior family with the other two families as targets, producing six off-policy settings. Supplementary Material~\ref{app:sim_setting} gives the complete policy parameterizations and linear-Gaussian data-generating system. For each target policy and evaluation point $(\bs,\ba)$, the Monte Carlo benchmark uses $10{,}000$ target-policy trajectories of horizon $T_{\mathrm{MC}}=500$. KE-DRL is fitted to $100$ independent behavior-policy datasets, each containing $n=300$ trajectories with $T=50$ recorded time points after a burn-in of $100$ steps. Each replicate therefore contains $N=14{,}700$ pooled transitions, with state, reward, and action dimensions $p=5$, $d=3$, and $q=1$. The transition and reward innovations are conditionally independent, so $\bR\indep\bS'\mid\bX$, which is sufficient for Assumption~\ref{ass:reward_continuation_separation}.

\noindent Figure~\ref{fig:mu_res_2} compares the estimated mean embeddings with the Monte Carlo benchmark for the Uniform behavior policy and Gaussian target policy. At the displayed state-action points, the estimated and Monte Carlo embedding functions have similar values over the plotted evaluation grids, even though the training data are generated under a different behavior policy. This comparison concerns pointwise function evaluations and is not an RKHS-norm or MMD comparison. To assess whether this agreement extends beyond a single state-action pair, we repeat the evaluation at $10$ randomly selected pairs $(\bs,\ba)$. For each pair, the error metrics are computed over $100$ independent offline-data replicates and then summarized across the $10$ pairs. Table~\ref{tab:metric_summary_across_points} reports pointwise signed bias, RMSE, MAE, and held-out-input projected Bellman risk.

\noindent Although $\widehat{\mu}_{\bZ\mid\bs,\ba}-\mu_{\bZ\mid\bs,\ba}$ is an $\mc H_{\mc Z}$-valued discrepancy, the reported scalar error metrics are computed from pointwise evaluations on the same evaluation grid for each for each state--action pair. Bias, RMSE, and MAE are the average signed error, root-mean-square error, and mean absolute error over this grid, respectively. These quantities are pointwise embedding-function evaluation errors, not RKHS-norm or MMD errors. The summaries in Table~\ref{tab:metric_summary_across_points} are averaged first over offline-data replicates and then over evaluation pairs. Supplementary Material~\ref{app:mc-repeat-target} gives the aggregation formulas and reports a separate repeated Monte Carlo reference diagnostic. A negative bias indicates that the estimated embedding function is smaller than the Monte Carlo reference on average over the evaluation grid. The held-out-input projected Bellman risk evaluates the fitted plug-in residual at held-out inputs and measures empirical agreement with the projected Bellman equation.

\noindent Supplementary Material~\ref{app:sim_results} reports the remaining behavior-target comparisons and the complete set of benchmark results.

\begin{table}[tbp]
\centering
\caption{Pointwise embedding-function evaluation errors across 10 randomly selected evaluation pairs $(\bs,\ba)$. For each pair, the metrics are computed over 100 independent offline-data replicates and then averaged over the pairs; parentheses report standard deviations across evaluation pairs. Bias, RMSE, and MAE are computed over common evaluation grids and are not RKHS-norm or MMD errors. Bellman Risk denotes held-out-input projected Bellman risk.}
\footnotesize
\label{tab:metric_summary_across_points}
\begin{tabular}{@{}ll|cccc@{}}
\toprule\toprule
\multicolumn{1}{c}{\textbf{Behavior policy}} & \multicolumn{1}{c|}{\textbf{Target Policy}} & \textbf{Bias} & \textbf{RMSE} & \textbf{MAE} & \textbf{Bellman Risk} \\ \midrule
\multirow{2}{*}{Gaussian} & Uniform  & $-0.0129$ (0.0121) & 0.0302 (0.0233) & 0.0257 (0.0191) & 0.0155 (0.0144) \\
 & Logistic & $-0.0089$ (0.0144) & 0.0276 (0.0251) & 0.0235 (0.0213) & 0.0141 (0.0128) \\ \midrule
\multirow{2}{*}{Logistic} & Uniform  & $-0.0019$ (0.0197) & 0.0304 (0.0220) & 0.0258 (0.0184) & 0.0111 (0.0090) \\
 & Gaussian & $-0.0031$ (0.0187) & 0.0296 (0.0250) & 0.0254 (0.0215) & 0.0129 (0.0114) \\ \midrule
\multirow{2}{*}{Uniform}  & Gaussian & 0.0044 (0.0125) & 0.0255 (0.0114) & 0.0216 (0.0096) & 0.0116 (0.0099) \\
 & Logistic & 0.0020 (0.0130) & 0.0255 (0.0128) & 0.0215 (0.0106) & 0.0089 (0.0068) \\
\bottomrule\bottomrule
\end{tabular}
\end{table}

\begin{figure}[tbp]
\centering
\includegraphics[width=0.70\linewidth]{plots/simulation_res/mu_summary_benchmarks_UG_3x2.png}
\caption{\setstretch{0.8}Simulation study performance metrics. Uniform behavior policy and Gaussian target policy. Pointwise evaluations of the estimated mean-embedding functions and of the Monte Carlo benchmark over the plotted evaluation grids at the benchmark target points. Shared and pair-specific settings are given in Tables \ref{tab:sim_shared_params} and \ref{tab:sim_pair_params} in Supplementary Material.}
\label{fig:mu_res_2}
\end{figure}

\section{Expedia Hotel Search Data}\label{sec:realdata}

We use a subset of the Expedia Hotel Search data from the Kaggle \emph{Expedia Personalized Sort} competition \citep{expedia_data} to illustrate estimation with multivariate rewards, continuous action summaries, and mixed one-step reward types. Because the data are observational and the action summaries are not randomly assigned, we interpret the analysis as a comparison of fitted policy-indexed conditional return summaries, neither causally nor as business recommendations. Supplementary Material~\ref{app:Exp_data} provides the variable definitions, estimation and tuning choices, support diagnostics, and raw-scale action histograms.

\noindent We index trajectories by search destination, recorded as \texttt{srch\_destination\_id}, and treat each search within a destination as one time point. For each displayed hotel list we construct a $7$-dimensional state vector $\bs_{it}$, a $3$-dimensional continuous action summary $\ba_{it}$, and a bivariate reward
\[
\br_{it}=\bigl(\text{gross revenue per night},\ \text{total clicks}\bigr)
\]
\noindent The action vector summarizes observed features of the displayed list. The average price per night, the total number of promotions, and the within-list price dispersion. These coordinates are low-dimensional descriptors of the realized display, not intervention variables assigned in a randomized experiment, and should be interpreted accordingly.

\noindent We first fit two reward-specific linear-Gaussian target policies from the logged sample. The subsequent KE-DRL evaluation conditions on these realized fitted policies, which are not re-estimated during that step, and does not propagate policy-construction uncertainty. The first policy, $\pi^{(\mathrm{rev})}$, is weighted toward larger gross revenue per night; the second, $\pi^{(\mathrm{clk})}$, is weighted toward larger total clicks. Both have the same parametric form and are evaluated under the same KE-DRL specification, so the comparison reflects differences between policy-indexed summaries rather than differences in the estimation procedure. The reported analysis uses $10{,}000$ training transitions and $4{,}000$ test transitions. The seven raw state variables include categorical components that are one-hot encoded before fitting, yielding a $28$-dimensional encoded state. For comparability, both policies use the same return kernel, regularization, return grid, state-action basis, and post-estimation recovery specification. The complete numerical specification is given in Supplementary Material~\ref{subsec:expedia_policy_eval_details}.

\noindent The one-step reward has a continuous revenue coordinate and a count-valued click coordinate. This mixed one-step structure does not by itself determine whether the discounted return law is absolutely continuous, because $\sum_{t\ge0}\gamma^t C_t$ need not be integer-valued or atomic. We do not verify the Sobolev-density conditions of Theorem~\ref{thm:fixed_point} for this application and therefore do not use the application as verification of that theorem. The conditional embedding remains well defined for a bounded return kernel, while the conditional-embedding error bounds apply only when the separate well-specifiedness, source, and covariance-concentration conditions of Theorem~\ref{thm:pointwise_error_bound_intro} hold. Separately, the reported Expedia fit uses the separated Bellman construction under Assumption~\ref{ass:reward_continuation_separation} as a working moment restriction, and the displayed conditional summaries are interpreted under this restriction.

\noindent Relative to the logged behavior, $\pi^{(\mathrm{rev})}$ shifts the display toward higher average prices and greater within-list price dispersion while leaving promotions close to the logged level. In contrast, $\pi^{(\mathrm{clk})}$ shifts toward lower prices, more promotions, and lower price dispersion. The corresponding held-out-input projected Bellman risks are $0.0513$ and $0.0478$. These values measure held-out agreement with the projected Bellman equation and are not estimates of business performance. For all three action summaries, the greedy policy actions lie within the observed marginal training ranges. This rules out extrapolation beyond the observed marginal ranges but does not establish joint overlap. Supplementary Table~\ref{tab:expedia_policy_action_summary} gives the complete action and risk summary.

\begin{figure}[tbp]
\centering
\includegraphics[width=0.70\textwidth,trim=0in 5.6in 0in 0.15in,clip]{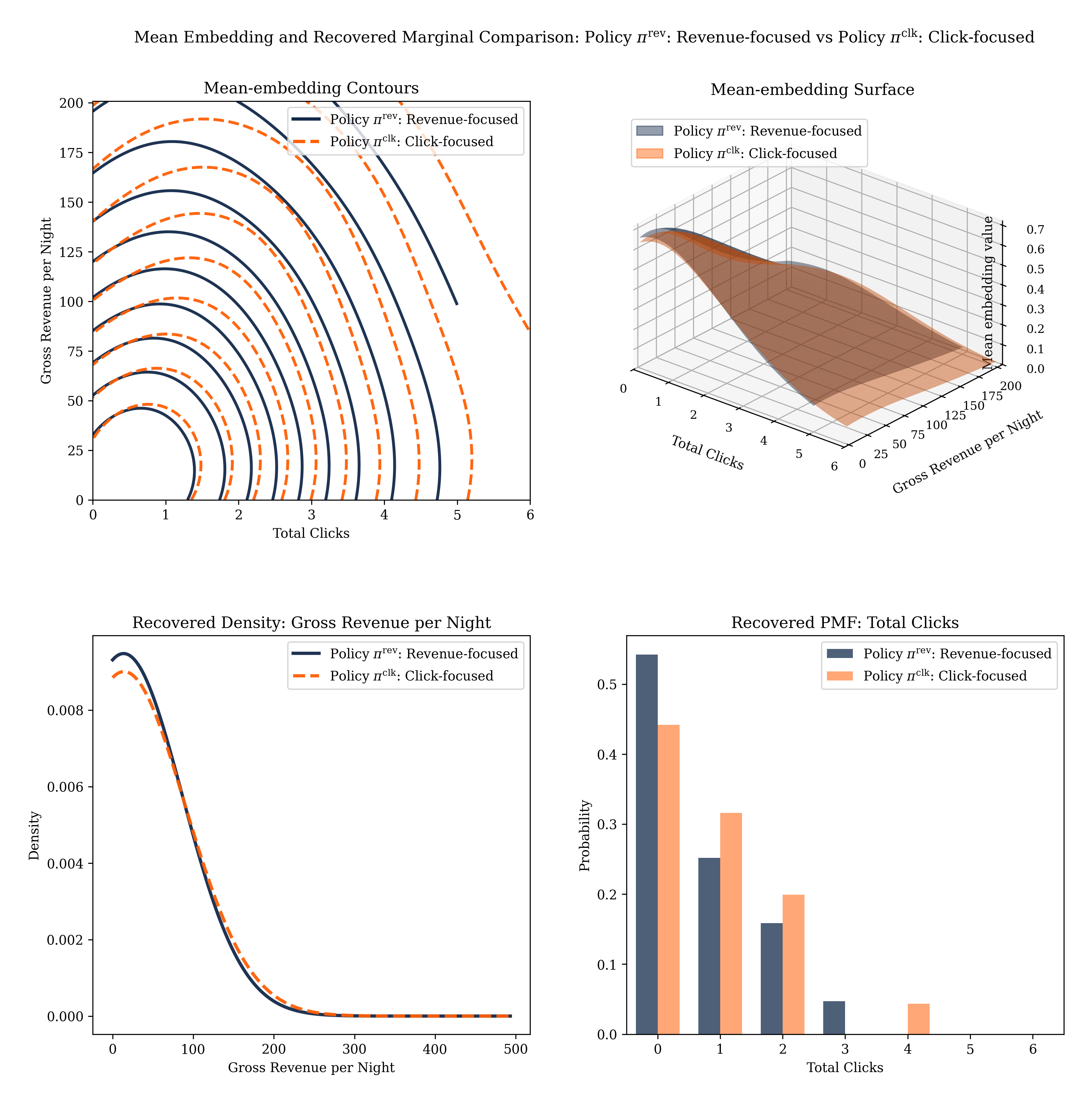}
\caption{Estimated RKHS mean embeddings for the two Expedia policies at the same selected state and their respective greedy initial actions. Left: contours; right: surfaces. The displayed values are not probability densities.}
\label{fig:expedia_res}
\end{figure}

\noindent Figure~\ref{fig:expedia_res} presents the principal distributional comparison. Both figures use the same selected state, but each conditions on the corresponding policy's greedy initial action. The contrast therefore reflects the policy-specific continuation rule and the policy-specific initial action; it is not a comparison of state-level return laws averaged over a common initial-action distribution. The two panels display the estimated RKHS mean-embedding functions over the bivariate return grid, not probability densities. In portions of the grid with larger click-return coordinates, $\pi^{(\mathrm{clk})}$ has relatively larger embedding values than $\pi^{(\mathrm{rev})}$. This pattern is consistent with the coordinatewise policy-overlap summaries in Supplementary Material~\ref{app:Exp_data}, although those summaries do not establish joint overlap.

\noindent Supplementary Figure~\ref{fig:expedia_res_full} adds recovered probability proxies under the common recovery specification. The revenue marginals overlap substantially. The click-focused proxy has a larger projected mean and more mass at or above one and two, whereas the difference at or above three is small. These are coordinate-specific differences between the conditional multivariate-return summaries, not uniform policy dominance.

\section{Discussion and Conclusion}\label{sec:conclusion}

This paper develops KE-DRL for offline policy evaluation of conditional multivariate return distributions, separating population identification, finite-dimensional embedding estimation, and post-estimation recovery. At the population level, a Mat\'ern embedding discrepancy identifies the unique Wasserstein Bellman fixed point on the stated Bellman-invariant Sobolev-moment class and converts the population embedding residual into Wasserstein control. KE-DRL then fits a finite-dictionary RKHS embedding from logged transitions through an MMD-based Bellman criterion. Smooth RKHS utilities and kernel-smoothed distribution functions can be evaluated from this embedding, whereas density- and quantile-related summaries require regularized recovery.

\noindent The result for the regularized conditional mean embedding estimator applies under the stated kernel, source, and covariance-concentration conditions. It is not an end-to-end bound for the finite-dictionary estimator because it does not jointly account for finite-grid approximation, estimated density ratios, random features, numerical optimization, and recovery. The simulations compare pointwise embedding evaluations with Monte Carlo benchmarks over specified grids, and the Expedia analysis compares conditional multivariate-return summaries for two constructed policies.

\noindent Limited overlap and highly variable importance weights can increase the variability of off-policy estimates. The reported analyses use the separated Bellman construction in Assumption~\ref{ass:reward_continuation_separation}; the paired construction preserves the joint reward-successor target but generally cannot reduce its target-target term to the same $\mathbb B$-independent $m\times m$ reward-shift matrices. High-dimensional kernel and matrix computations may require low-rank, tensorized, or random-feature approximations. Finally, sharp nonsmooth functionals require problem-specific regularization, and mixed or atomic return laws lie outside the Sobolev-density class used in the fixed-point analysis.
\JASADeclarations
\begingroup
\spacingset{\JASAReferenceSpacing}
\bibliography{bib_JASA}
\endgroup
\newpage
\fi
\ifmainonly
\else
\appendix
\addcontentsline{toc}{section}{Supplementary Material}
\setcounter{section}{0}
\setcounter{page}{1}
\setcounter{equation}{0}
\setcounter{theorem}{0}
\setcounter{table}{0}
\setcounter{figure}{0}
\setcounter{algorithm}{0}
\renewcommand{\thesection}{S\arabic{section}}
\renewcommand{\theequation}{S\arabic{equation}}
\renewcommand{\thetable}{S\arabic{table}}
\renewcommand{\thepage}{S\arabic{page}}
\renewcommand{\thefigure}{S\arabic{figure}}
\renewcommand{\thealgorithm}{S\arabic{algorithm}}
\renewcommand{\theHsection}{supp.\arabic{section}}
\renewcommand{\theHequation}{supp.\arabic{equation}}
\renewcommand{\theHtable}{supp.\arabic{table}}
\renewcommand{\theHfigure}{supp.\arabic{figure}}
\renewcommand{\theHalgorithm}{supp.\arabic{algorithm}}
\renewcommand{\theHtheorem}{supp.theorem.\arabic{theorem}}
\renewcommand{\theHclaim}{supp.claim.\arabic{theorem}}
\renewcommand{\theHproposition}{supp.proposition.\arabic{theorem}}
\renewcommand{\theHcorollary}{supp.corollary.\arabic{theorem}}
\renewcommand{\theHconjecture}{supp.conjecture.\arabic{theorem}}
\renewcommand{\theHlemma}{supp.lemma.\arabic{theorem}}
\renewcommand{\theHdefinition}{supp.definition.\arabic{theorem}}
\renewcommand{\theHremark}{supp.remark.\arabic{theorem}}
\renewcommand{\theHassumption}{supp.assumption.\arabic{theorem}}
\spacingset{\JASASupplementSpacing}
\begin{center}
    \textbf{\LARGE Supplementary Material}\\
    \textit{Vector-Valued Distributional Reinforcement Learning Policy Evaluation: A Hilbert Space Embedding Approach}
    \JASASupplementAuthorLine
\end{center}
\section{Notation Reference}\label{app:notation_reference}
\footnotesize
\begin{longtable}{@{}>{\raggedright\arraybackslash}p{0.20\textwidth}>{\raggedright\arraybackslash}p{0.77\textwidth}@{}}
\caption{Main notation, ordered by conceptual dependency. Symbols used only locally in a single supplement subsection are defined where they occur.}\\
\toprule\toprule
\textbf{Symbol} & \textbf{Meaning} \\
\midrule
\endfirsthead
\toprule\toprule
\textbf{Symbol} & \textbf{Meaning} \\
\midrule
\endhead
\bottomrule\bottomrule
\endfoot
\multicolumn{2}{@{}l}{\textbf{Data, policies, and Bellman objects}} \\
\midrule
$\mc S,\mc A,\mc R,\mc Z$ & State, action, reward, and return spaces; $\mc X=\mc S\times\mc A$, $\mc Z\subseteq\R^d$. \\
$\bX=(\bS^\top,\bA^\top)^\top$, $\bx$ & State-action input (random variable, realization); primes ($\bX'$, $\bx'$) denote generic successors, with transition-observation subscripts ($\bx_j'$) for realized successors in the sample. \\
$\pi,\beta$; $\gamma$; $\mc D$; $n,T_i,N$ & Target and behavior policies; discount factor in $(0,1)$; logged data; numbers of trajectories, recorded time points in trajectory $i$, and pooled transition rows, with $N=\sum_i(T_i-1)$. Within Section~\ref{sec:tikhonov_reference}, $n$ is used locally for the number of observed pairs $(\bX_i,\bZ_i)$. \\
$\bZ^\pi(\bx)$; $\mc V^\pi_{\bx}$ & Target-policy discounted return from $\bx$; its conditional law $\mc L\{\bZ^\pi(\bx)\}$. \\
$\T^\pi$ & Distributional Bellman operator on conditional law maps, applied before embedding. \\
$\mathcal{M}^\pi(d\br,d\bx'\mid\bx)$; $\mc{P}_X^\pi(d\bx'\mid\bx)$ & Joint target-policy one-step law of $(\bR,\bX')$ given $\bX=\bx$, and its $\bX'$-marginal, with density $p^\pi(\bx'\mid\bx)=p(\bs'\mid\bs,\ba)\pi(\ba'\mid\bs')$ when a dominated form is used; the $\bR$-marginal $F_{\bR}(\cdot\mid\bx)$ does not depend on the target policy because the successor action is drawn after the reward is realized. \\
$\mc{G}_{\mathrm{wt}}$; $\mathbb P_{\bX}$ & Weighting law over conditioning inputs in population losses; marginal input law in $L^2(\mathbb P_{\bX};\mc H_{\mc Z})$ statements. \\
\midrule
\multicolumn{2}{@{}l}{\textbf{Kernels, embeddings, and fixed-point theory}} \\
\midrule
$k_{\mc Z}$, $\mc H_{\mc Z}$; $ k_{\mc X}$, $\mc H_{\mc X}$ & Return-space (Mat\'ern) and state-action kernels with their RKHSs; $\kappa_{\mc Z}^2=\sup_\bz k_{\mc Z}(\bz,\bz)$, $\kappa_{\mc X}^2=\sup_\bx k_{\mc X}(\bx,\bx)$. \\
$\mu^\pi(\bx)$, $\mu_{\bZ\mid\bx}$ & Conditional mean embedding of the (target-policy) return law at $\bx$. \\
$\Bgamma_k$; $D_k(\mathbb P,\mathbb Q)$ & MMD induced by $k_{\mc Z}$ (pointwise or between laws); maximal conditional MMD $\sup_{\bx}\Bgamma_k\{\mathbb P(\cdot\mid\bx),\mathbb Q(\cdot\mid\bx)\}$. \\
$W_1$; $d_1(\mathbb P,\mathbb Q)$ & $1$-Wasserstein distance; maximal conditional $W_1$ over $\bx\in\mc X$. \\
$\theta_{\br,\gamma}$; $\mc{C}_{\br,\gamma}$, $\mc{C}_{\br,\gamma}^*$ & Affine Bellman map $\theta_{\br,\gamma}(\bz)=\br+\gamma\bz$; induced composition operator $(\mc{C}_{\br,\gamma}h)(\bz)=h(\br+\gamma\bz)$ on $\mc H_{\mc Z}$ and its adjoint, which maps a continuation embedding to the pushforward embedding. \\
$\mbC_{\mc{XX}},\mbC_{\mc{ZX}}$; $\widehat{\mbC}_{\mc{XX}},\widehat{\mbC}_{\mc{ZX}}$ & Uncentered population input covariance and output-input cross-covariance operators, with $\mbC_{\mc{ZX}}:\mc H_{\mc X}\to\mc H_{\mc Z}$ and $\mbC_{\mc{XZ}}:=\mbC_{\mc{ZX}}^*$; empirical counterparts. \\
$\mathfrak S_{b,\mathfrak m,r,s}$ & Sobolev-moment class of laws with density in the $H^s$-ball of radius $b$ and $r$-th moment bound $\mathfrak m$ ($\mathfrak m$ is distinct from the grid size $m$). \\
$\mc E$; $K$; $\mathfrak T^\pi$; $\mathbb P_\mu$ & Reflexive ambient space of conditional embedding maps; admissible closed convex set; induced Bellman embedding operator $\mathfrak T^\pi\mu=\mu_{\T^\pi\mathbb P_\mu}$; law map represented by $\mu\in K$. \\
$\Upsilon_{\bx'}^{\mathbb P,\mathbb Q}$ & Measurable optimal-coupling kernel between $\mathbb P(\cdot\mid\bx')$ and $\mathbb Q(\cdot\mid\bx')$. \\
$\rho$; $L_k$; $c_*, C_*$ & H\"older exponent $(r-1)/(d+2r)$ in the MMD-Wasserstein comparison; feature-map Lipschitz constant of $k_{\mc Z}$; lower and upper comparison constants. \\
$s_k$; $\tau$ & Fourier-decay order of $k_{\mc Z}$ ($s_k=2\nu+d$ for Mat\'ern); source-condition exponent in $(0,1/2]$. \\
\midrule
\multicolumn{2}{@{}l}{\textbf{Finite dictionaries and fitted embedding map}} \\
\midrule
$\mathbf Z_m=\{\bz_i\}_{i=1}^m$; $\bk_{\mc Z}(\cdot)$; $\mathbb K_{\mc Z}$ & Return grid of size $m$; vector of kernel sections $(k_{\mc Z}(\bz_1,\cdot),\ldots,k_{\mc Z}(\bz_m,\cdot))^\top$; $m\times m$ return-grid Gram matrix. \\
$\mc C_L=\{\underline\bx_\ell\}_{\ell=1}^L$; $\bk_L(\bx)$; $\mathbb K_L$ & State-action basis dictionary of size $L$; basis feature vector at $\bx$; $L\times L$ basis Gram matrix. \\
$\mathbb B$; $\omega_i(\bx;\mathbb B)$; $\bso(\bx;\mathbb B)$ & Single policy-specific $L\times m$ coefficient matrix; coefficient on atom $\bz_i$ at input $\bx$; current-input coefficient vector $\mathbb B^\top\bk_L(\bx)$. \\
$\ell_i(\bx,\cdot)$; $\bso^\pi(\bx;\mathbb B)$ & Reward-shifted kernel section $\int_{\mc R}k_{\mc Z}(\br+\gamma\bz_i,\cdot)\,F_{\bR}(d\br\mid\bx)$; target-policy continuation coefficient vector. \\
$\mathbb H(\bx),\mathbb G(\bx)$; $\widehat{\mathbb H}(\bx),\widehat{\mathbb G}(\bx)$ & $m\times m$ reward-shift inner-product matrices (cross and target-target) and their plug-in estimators. \\
$\boldsymbol\Gamma(\bx)$; $\mathbb K_{\mc X}$ & Kernel-ridge conditional weights at conditioning point $\bx$; $N\times N$ Gram matrix on pooled inputs. \\
$\eta(\bx')$; $\mathbb D_\eta$; $\mathbb K_{L^+}$; $\boldsymbol\Phi_L(\bx)$ & Successor-action density ratio (importance ratio) $\pi/\beta$; diagonal matrix of the fitted ratios $\widehat\eta_j=\widehat\eta(\bx_j')$; $L\times N$ basis-to-successor cross-Gram matrix; continuation feature $\mathbb K_{L^+}\mathbb D_\eta\boldsymbol\Gamma(\bx)$. \\
\midrule
\multicolumn{2}{@{}l}{\textbf{Empirical objective and regularization}} \\
\midrule
$\mc X_\star=\{\bx_q^\star\}_{q=1}^M$; $\varrho_q$ & Conditioning set of size $M$ ($M=N$ recovers the full empirical objective); nonnegative weights with $\sum_q\varrho_q=1$. \\
$u_q(\mathbb B),v_q(\mathbb B)$ & Current and continuation coefficient vectors $\mathbb B^\top\bk_{L,q}$ and $\mathbb B^\top\boldsymbol\Phi_{L,q}$. \\
$\widehat{\mathbb B}_{\mathrm{opt}}^{\pi}$; $\widehat{\mathbb B}^{\pi}$ & An exact minimizer of the stated empirical criterion; finite-iteration numerical estimate computed by Algorithm~\ref{alg:rk_drl}. \\
$\lambda_\Gamma,\lambda$; $\lambda_B$; $\lambda_{\mathrm{mass}},c_{\mathrm{mass}}$; $\lambda_{\mathrm{neg}}$ & Conditional-weight ridge and operator-scale regularization; coefficient ridge; coefficient-mass penalty weight and target mass; optional weight for the training-point negative-part penalty. \\
$\varepsilon_{\mc{ZX}}(n,\delta),\varepsilon_{\mc{XX}}(n,\delta)$; $c_{\bx}, c_\star$ & Covariance-operator concentration radii; pointwise and uniform source constants in the error bounds. \\
$Q_w$; $w\in\Delta_m$ & Post-estimation recovered probability law over return-grid atoms; simplex weights used only in post-estimation probability recovery. \\
\midrule
\multicolumn{2}{@{}l}{\textbf{Expedia application (Section~\ref{sec:realdata}, Supplementary Material~\ref{app:Exp_data})}} \\
\midrule
$\pi^{(\mathrm{rev})},\pi^{(\mathrm{clk})}$; $a_c^{\mathrm{gr}}(s)$ & Revenue- and click-focused target policies; clipped greedy action of policy $c$. \\
$\Pi_{\mc A},\Pi_{\mc C}$; $\bar c_i$ & Clipping to the empirical action support and the imposed recovery projection onto observed one-step count support $\mc C$; projected click-return coordinate of atom $\bz_i$. \\
$\bx_\star^\pi$; $\widehat{\boldsymbol\beta}_{\pi}(\bx)$ & Policy-specific evaluation point (median first-PC state, greedy action); fitted coefficient vector, not a probability vector out of sample. \\
$h_R,\phi_{h_R}$; $\lambda_{\mathrm{den}}$; $\lambda_2,\lambda_{\mathrm{KL}}$ & Revenue smoothing bandwidth and kernel; spectrum-scaled recovery ridge; regularization penalties toward the reference vector $w_0$. \\
$A$; $w_0$; $\widehat w$; $\widehat f_R,\widehat p_C$ & Recovery cross-Gram matrix; reference vector $w_0$ (normalized positive part of $\widehat{\boldsymbol\beta}$); recovered simplex weights; recovered revenue density and discrete click-return proxy. \\
$\mathrm{ESS}(\widehat\eta)$ & Effective sample size $(\sum_j\widehat\eta_j)^2/\sum_j\widehat\eta_j^2$ of calibrated ratio weights. \\
\end{longtable}
\normalsize

\section{Additional Mathematical Derivations}
\subsection{A Paired Bellman Target without Reward-Continuation Factorization}\label{app:paired_bellman_estimator}
Assumption~\ref{ass:reward_continuation_separation} is used only to replace the joint target in \eqref{eq:projected_pair_embedding} with the separated target of Proposition~\ref{prop:DBO_empirical}. The separated representation expresses the target through the $\mathbb B$-independent $m\times m$ reward-shift matrices $\widehat{\mathbb H}(\bx)$ and $\widehat{\mathbb G}(\bx)$ in \eqref{eq:h_ij}-\eqref{eq:g_ij} and the continuation vector $\widehat\bso^{\pi}(\bx;\mathbb B)$ in \eqref{eq:Phi}. More precisely, the assumption concerns a single joint law and equates a joint integral with the product of the corresponding marginal integrals. Equivalently, conditional on $\bX=\bx$, the $\mc H_{\mc Z}$-valued covariance between the reward-shift feature $k_{\mc Z}(\bR+\gamma\bz_i,\cdot)$ and the continuation coefficient $\omega_i(\bX';\bB)$ is zero. It does not make the reward deterministic or independent of the full future return. The reward remains random in the shifted embedding $\int_{\mc R}k_{\mc Z}(\br+\gamma\bz_i,\cdot)\,F_{\bR}(d\br\mid\bx)$.

\noindent This restriction is weaker than $\bR\indep\bX'\mid\bX=\bx$. One sufficient condition is $\omega_i(\bX';\bB)\indep\bR\mid\bX=\bx$, which may hold even when $\bR$ and $\bX'$ remain dependent through components of $\bX'$ not represented by the scalar coefficient $\omega_i(\bX';\bB)$. A stronger sufficient condition is $\bR\indep\bS'\mid\bX=\bx$. Because the successor action is drawn from $\pi(\cdot\mid\bS')$ after the reward is realized and is conditionally independent of $\bR$ given $(\bX,\bS')$, this reward-successor-state condition implies $\bR\indep\bX'\mid\bX=\bx$, and hence the factorization for every $i$ and every $\bB$. These independence conditions are sufficient but are not assumed directly.

\noindent The factorization may fail when the realized reward is included in the successor state or affects the next action through state variables observed by the policy. The paired construction below preserves the realized pair $(\bR,\bX')$ in such settings. The restriction applies only to the separated estimator used in the reported analyses and has no role in the population fixed-point result.

\noindent When this conditional moment restriction is not appropriate, for example when the realized reward is a component of the next state, the underlying conditional return-law target and projected Bellman residual remain unchanged. The relevant finite-dictionary Bellman target is the paired embedding $\widehat{\mu}_{\T^\pi\bZ\mid\bx}(\cdot\,;\mb B)$ in \eqref{eq:projected_pair_embedding}. A direct plug-in estimator replaces $\mathcal{M}^\pi(d\br,d\bx'\mid\bx)$ with the kernel-ridge weights $\Gamma_r(\bx)$ from \eqref{eq:Gamma} and the density ratios $\widehat\eta_r$ from Section~\ref{sec:ope_weights}, giving
\[
\sum_{r=1}^{N}
\Gamma_r(\bx)\,\widehat\eta_r
\sum_{i=1}^{m}
\omega_i(\bx_r';\mb B)\,
k_{\mc Z}(\br_r+\gamma\bz_i,\cdot)\]
where transition observation $r$ contributes its realized pair $(\br_r,\bx_r')$. Thus, the paired construction does not impose $\bR\indep\bX'\mid\bX$.

\noindent The paired construction is more computationally intensive. Expanding the squared $\mc H_{\mc Z}$-norm couples pairs of transitions and return atoms through $k_{\mc Z}(\br_r+\gamma\bz_i,\br_{r'}+\gamma\bz_j)$. Because the target-target term contains transition-specific coefficients $\omega_i(\bx_r';\mathbb B)$, it cannot generally be reduced to the same $\mathbb B$-independent $m\times m$ reward-shift matrices. Direct evaluation has order $N^2m^2$ cost at each conditioning point and optimization iteration unless mini-batching or low-rank kernel approximations are used. Under the separation assumption, the reward-dependent terms reduce to $\mathbb B$-independent $m\times m$ matrices, although constructing these matrices may still be computationally costly. The reported analyses use Assumption~\ref{ass:reward_continuation_separation}; the paired construction is applicable when the restriction is unsuitable and the sample size or approximation scheme makes direct evaluation feasible.

\subsection[Derivation of the projected Bellman residual]{Derivation of equation \eqref{eq:gamma_with_W}}\label{drive:gamma_with_W}
We first write the calculation at a state-action pair $(\bs,\ba)$ and then use the input notation $\bx=(\bs^\top,\ba^\top)^\top$. Using Proposition~\ref{prop:DBO_empirical} and suppressing dependence on $\mathbb B$ when no ambiguity arises,
\begin{align*}
\widehat{\Bgamma}_{k}^{\,2}(\bs,\ba;\mathbb B)
&:= \Big\|\widehat{\mu}_{\bZ|\bs,\ba}-\widehat{\mu}_{\T^\pi \bZ|\bs,\ba}\Big\|_{\mathcal{H}_{\mathcal{Z}}}^2 \\
&=
\left\|
\sum_{i=1}^m \omega_i(\bs,\ba)k_{\mc Z}(\bz_i,\cdot)
-
\sum_{j=1}^m \omega_j^\pi(\bs,\ba)\ell_j((\bs,\ba),\cdot)
\right\|_{\mc H_{\mc Z}}^2\\
&=
\sum_{i,j}\omega_i(\bs,\ba)\omega_j(\bs,\ba)
\langle k_{\mc Z}(\bz_i,\cdot),k_{\mc Z}(\bz_j,\cdot)\rangle_{\mc H_{\mc Z}}\\
&\quad
-2\sum_{i,j}\omega_i(\bs,\ba)\omega_j^\pi(\bs,\ba)
\langle k_{\mc Z}(\bz_i,\cdot),\ell_j((\bs,\ba),\cdot)\rangle_{\mc H_{\mc Z}}\\
&\quad
+\sum_{i,j}\omega_i^\pi(\bs,\ba)\omega_j^\pi(\bs,\ba)
\langle \ell_i((\bs,\ba),\cdot),\ell_j((\bs,\ba),\cdot)\rangle_{\mc H_{\mc Z}}
\end{align*}
Here
$\omega_j^\pi(\bs,\ba)=\E_{p^\pi(\bS',\bA'\mid\bs,\ba)}\{\omega_j(\bS',\bA')\}$ and
$\ell_j((\bs,\ba),\cdot)=\E\{k_{\mc Z}(\gamma\bz_j+\bR,\cdot)\mid\bs,\ba\}$.
The reproducing property gives $\langle k_{\mc Z}(\bz_i,\cdot),k_{\mc Z}(\bz_j,\cdot)\rangle_{\mc H_{\mc Z}}=k_{\mc Z}(\bz_i,\bz_j)$, and hence
\begin{align*}
\widehat{\Bgamma}_{k}^{\,2}(\bs,\ba;\mathbb B)
&=
\sum_{i,j}\omega_i(\bs,\ba)\omega_j(\bs,\ba)k_{\mc Z}(\bz_i,\bz_j)\\
&\quad
-2\sum_{i,j}\omega_i(\bs,\ba)
\E_{p^\pi(\bS',\bA'\mid\bs,\ba)}\{\omega_j(\bS',\bA')\}
\E\{k_{\mc Z}(\bz_i,\gamma\bz_j+\bR)\mid\bs,\ba\}\\
&\quad
+\sum_{i,j}
\E_{p^\pi(\bS',\bA'\mid\bs,\ba)}\{\omega_i(\bS',\bA')\}
\E_{p^\pi(\bS',\bA'\mid\bs,\ba)}\{\omega_j(\bS',\bA')\}\\
&\qquad\qquad\times
\E\{k_{\mc Z}(\gamma\bz_i+\bR,\gamma\bz_j+\widetilde{\bR})\mid\bs,\ba\}
\end{align*}
where $\bR$ and $\widetilde{\bR}$ are conditionally independent draws from the reward law given $(\bs,\ba)$ in the last term. Returning to $\bx=(\bs^\top,\ba^\top)^\top$, define
\[
h_{ij}(\bx)
:=
\left\langle k_{\mc Z}(\bz_i,\cdot),\ell_j(\bx,\cdot)\right\rangle_{\mc H_{\mc Z}}
=
\E\{k_{\mc Z}(\bz_i,\gamma\bz_j+\bR)\mid\bX=\bx\}
\]
and, for conditionally independent reward draws $\bR,\widetilde{\bR}\mid\bX=\bx$,
\[
g_{ij}(\bx)
:=
\left\langle \ell_i(\bx,\cdot),\ell_j(\bx,\cdot)\right\rangle_{\mc H_{\mc Z}}
=
\E\{k_{\mc Z}(\gamma\bz_i+\bR,\gamma\bz_j+\widetilde{\bR})\mid\bX=\bx\}
\]
Collecting the coefficients into
$\bso(\bx;\mathbb B)$ and $\bso^\pi(\bx;\mathbb B)$ gives the matrix expression in \eqref{eq:gamma_with_W}.

\noindent For computation, estimate the reward conditional mean embedding by
\[
\widehat{\mu}_{\bR\mid\bx}(\cdot)
=
\sum_{r=1}^{N}\Gamma_r(\bx)k_{\mc Z}(\br_r,\cdot)
\]
using the state-action kernel-ridge weights in \eqref{eq:Gamma}. Translation invariance of the return kernel gives $k_{\mc Z}(\bz_i,\gamma\bz_j+\br)=k_{\mc Z}(\bz_i-\gamma\bz_j,\br)$ for every $\br\in\mc R$, so that $h_{ij}(\bx)=\langle k_{\mc Z}(\bz_i-\gamma\bz_j,\cdot),\mu_{\bR\mid\bx}\rangle_{\mc H_{\mc Z}}$, where $\mu_{\bR\mid\bx}$ denotes the conditional mean embedding of the reward law. Its plug-in version is
\begin{align}
\widehat h_{ij}(\bx)
&=
\left\langle
k_{\mc Z}(\bz_i-\gamma\bz_j,\cdot),
\widehat{\mu}_{\bR\mid\bx}
\right\rangle_{\mc H_{\mc Z}}
= \sum_{r=1}^{N}\Gamma_r(\bx)k_{\mc Z}(\br_r,\bz_i-\gamma\bz_j)
\nonumber\\
&=
\widetilde{\bk}_{N}(\bx)^\top
\left(\mathbb K_{\mc X}+N\lambda_{\Gamma}\mathbb I_N\right)^{-1}
\bk_{\bR}^{(ij)}
\label{eq:h_ij}
\end{align}
where $\bk_{\bR}^{(ij)}=(k_{\mc Z}(\br_1,\bz_i-\gamma\bz_j),\ldots,k_{\mc Z}(\br_N,\bz_i-\gamma\bz_j))^\top$. Similarly, the plug-in estimator of $g_{ij}(\bx)$ is
\begin{align}
\widehat g_{ij}(\bx)
&=
\left\langle
\sum_{r=1}^N\Gamma_r(\bx)k_{\mc Z}(\gamma\bz_i+\br_r,\cdot),
\sum_{r'=1}^N\Gamma_{r'}(\bx)k_{\mc Z}(\gamma\bz_j+\br_{r'},\cdot)
\right\rangle_{\mc H_{\mc Z}}
\nonumber\\
&=
\boldsymbol{\Gamma}(\bx)^\top
\mathbb K_{\gamma\bz_i+\bR,\gamma\bz_j+\bR}
\boldsymbol{\Gamma}(\bx)
\label{eq:g_ij}
\end{align}
where the $(r,r')$ entry of $\mathbb K_{\gamma\bz_i+\bR,\gamma\bz_j+\bR}$ is
$k_{\mc Z}(\gamma\bz_i+\br_r,\gamma\bz_j+\br_{r'})$. The same conditional weights are used in both matrices, and $\mathbb G(\bx)$ is positive semidefinite because it is a Gram matrix of the estimated shifted sections.

\subsection{Population Finite-Dictionary Criterion and Ridge Identity}\label{app:population_finite_criterion}
Let $\mc{G}_{\mathrm{wt}}$ be a weighting distribution on $\mc X$. In the empirical criterion, $\mc{G}_{\mathrm{wt}}$ is approximated by the weighted empirical distribution on the conditioning set. If the scientific target concerns a particular region of the state-action space, the weighting distribution may emphasize that region, subject to the overlap condition in Section~\ref{sec:ope_weights}. The population finite-dictionary criterion is
\begin{equation}\label{eq:population_ke_loss}
\mc L_{\mathrm{B}}(\mathbb B)
:=
\int_{\mc X}
\Bgamma_k^{\,2}(\bx;\mathbb B)
d\mc{G}_{\mathrm{wt}}(\bx)
+
\lambda_B
\mathfrak{R}_B(\mathbb B),
\qquad
\mathfrak{R}_B(\mathbb B)
:= \mathrm{tr}\left(\mathbb B^\top\mathbb K_L\mathbb B\right)
\end{equation}
The first term averages the squared projected Bellman residual in \eqref{eq:gamma_with_W}. The second controls the combined RKHS norm of the $m$ coefficient functions. Using the $L$ basis functions in $\mc C_L$,
\begin{equation}\label{eq:rkhs_B_norm}
\sum_{i=1}^{m}
\left\|
\sum_{\ell=1}^{L}
b_{\ell i} k_{\mc X}(\underline \bx_\ell,\cdot)
\right\|_{\mc H_{\mc X}}^2
=
\mathrm{tr}
\left(
\mathbb B^\top\mathbb K_L\mathbb B
\right)
\end{equation}

\noindent For every conditioning point $q$, suppose the residual block matrix $\left[\begin{smallmatrix}\mathbb K_{\mc Z}&-\mathbb H_q\\-\mathbb H_q^\top&\mathbb G_q\end{smallmatrix}\right]$ is positive semidefinite. This holds when $\mathbb K_{\mc Z}$, $\mathbb H_q$, and $\mathbb G_q$ are Gram blocks constructed from the same exact kernel sections or the same finite feature map; see \eqref{eq:h_ij} and \eqref{eq:g_ij}. Under this condition, the Bellman, ridge, and coefficient-mass terms are convex quadratics in $\mathbb B$, while the optional negative-part term is convex and piecewise quadratic; the empirical criterion \eqref{eq:global_B_objective} is therefore a convex program in $\mathbb B$.

\subsection[Off-policy weight derivation]{Off-policy Weight Derivation for Equation \eqref{eq:Phi}}\label{app:Off_policy_IS}
We first derive the continuation weights using the explicit state-action notation $(\bs,\ba)$ and then use the compact input notation $\bx=(\bs^\top,\ba^\top)^\top$. Primes, as in $\bX'$ and $\bx'$, denote generic successor inputs under a transition law; realized successors carry observation subscripts, as in $\bx_j'$.

\noindent For a fixed return-grid atom $\bz_i$, the target-policy continuation coefficient is
\begin{align*}
\omega_i^\pi(\bs,\ba;\mathbb B)
&=
\E_{p^\pi(\bS',\bA'\mid\bs,\ba)}
\{\omega_i(\bS',\bA';\mathbb B)\}
\\
&=
\int_{\mc S}p(\bs'\mid\bs,\ba)
\left\{
\int_{\mc A}\pi(\ba'\mid\bs')\omega_i(\bs',\ba';\mathbb B)\,d\ba'
\right\}
d\bs'
\end{align*}
Since the transition law for $\bS'$ is common under the target and behavior policies,
\[
p^\pi(\bs',\ba'\mid\bs,\ba)
=
p(\bs'\mid\bs,\ba)\pi(\ba'\mid\bs')
\qquad
p^\beta(\bs',\ba'\mid\bs,\ba)
=
p(\bs'\mid\bs,\ba)\beta(\ba'\mid\bs')
\]
Under the positivity condition,
\[
\eta(\bs',\ba')
:=
\frac{p^\pi(\bs',\ba'\mid\bs,\ba)}{p^\beta(\bs',\ba'\mid\bs,\ba)}
=
\frac{\pi(\ba'\mid\bs')}{\beta(\ba'\mid\bs')}
\]
and therefore
\begin{align*}
\omega_i^\pi(\bs,\ba;\mathbb B)
&=
\int_{\mc S}p(\bs'\mid\bs,\ba)
\int_{\mc A}\beta(\ba'\mid\bs')
\frac{\pi(\ba'\mid\bs')}{\beta(\ba'\mid\bs')}
\omega_i(\bs',\ba';\mathbb B)\,d\ba'\,d\bs'
\\
&=
\E_{p^\beta(\bS',\bA'\mid\bs,\ba)}
\{\eta(\bS',\bA')\omega_i(\bS',\bA';\mathbb B)\}
\end{align*}

\noindent Let the pooled transition sample be
\[
\{(\bs_j,\ba_j,\br_j,\bs_j',\ba_j')\}_{j=1}^N,
\qquad
\bx_j=(\bs_j^\top,\ba_j^\top)^\top,
\quad
\bx_j'=((\bs_j')^\top,(\ba_j')^\top)^\top
\]
The empirical conditional mean embedding yields the following plug-in estimator of the conditional expectation under the behavior policy:
\[
\widehat\E_{p^\beta(\bS',\bA'\mid\bs,\ba)}\{g(\bS',\bA')\}
=
\sum_{j=1}^{N}\Gamma_j(\bs,\ba)g(\bs_j',\ba_j')
\]
where
\[
\boldsymbol\Gamma(\bs,\ba)
:=
\left(\mathbb K_{\mc X}+N\lambda_\Gamma\mathbb I_N\right)^{-1}
\widetilde\bk_{\mc X}^{N}(\bs,\ba)
\]
and $\widetilde\bk_{\mc X}^{N}(\bs,\ba)=\{ k_{\mc X}(\bx_1,(\bs^\top,\ba^\top)^\top),\ldots, k_{\mc X}(\bx_N,(\bs^\top,\ba^\top)^\top)\}^\top$. Taking $g(\bs',\ba')=\widehat\eta(\bs',\ba')\omega_i(\bs',\ba';\mathbb B)$ gives
\begin{align*}
\widehat\omega_i^\pi(\bs,\ba;\mathbb B)
&=
\sum_{j=1}^{N}
\Gamma_j(\bs,\ba)\widehat\eta(\bs_j',\ba_j')
\omega_i(\bs_j',\ba_j';\mathbb B)
\\
&=
\sum_{j=1}^{N}
\Gamma_j(\bs,\ba)\widehat\eta_j
\sum_{\ell=1}^{L}b_{\ell i} k_{\mc X}(\underline \bx_\ell,\bx_j')
\end{align*}
Thus only the values $\widehat\eta(\bx_j')\omega_i(\bx_j';\mathbb B)$ at the observed successor inputs are required; the product $\eta(\cdot)\omega_i(\cdot)$ need not belong to the state-action RKHS.

\noindent Stacking the $m$ coefficients gives
\begin{align*}
\widehat\bso^\pi(\bs,\ba;\mathbb B)
&=
\sum_{j=1}^{N}
\Gamma_j(\bs,\ba)\widehat\eta_j
\mathbb B^\top\bk_L(\bx_j')
\\
&=
\mathbb B^\top
\left\{
\sum_{j=1}^{N}
\Gamma_j(\bs,\ba)\widehat\eta_j
\bk_L(\bx_j')
\right\}
\end{align*}
Using $\bx=(\bs^\top,\ba^\top)^\top$, define
\[
\mathbb K_{L^+}
:=
\big[ k_{\mc X}(\underline \bx_\ell,\bx_j')\big]_{\ell=1,j=1}^{L,N}
\qquad
\mathbb D_\eta
:=
\operatorname{diag}(\widehat\eta_1,\ldots,\widehat\eta_N)
\]
The vector inside braces has $\ell$th component
\[
\sum_{j=1}^{N}
 k_{\mc X}(\underline \bx_\ell,\bx_j')
\widehat\eta_j
\Gamma_j(\bx)
=
\left[\mathbb K_{L^+}\mathbb D_\eta\boldsymbol\Gamma(\bx)\right]_\ell
\]
Thus
\[
\boldsymbol\Phi_L(\bx)
:=
\mathbb K_{L^+}\mathbb D_\eta\boldsymbol\Gamma(\bx)
\in\R^L
\]
and
\[
\widehat\bso^\pi(\bx;\mathbb B)
=
\mathbb B^\top\boldsymbol\Phi_L(\bx)
=
\mathbb B^\top\mathbb K_{L^+}\mathbb D_\eta\boldsymbol\Gamma(\bx)
\]
which is equation~\eqref{eq:Phi}.

\subsection{Pushforward Mean Embeddings for the Mat\'ern Kernel}\label{app:pushforward_operator}
\noindent This subsection connects the conditional return embedding to the distributional Bellman recursion through the pushforward of the return law under the Bellman map and establishes the pushforward mean-embedding identities used in Section~\ref{sec:bellman_discrepancy}. 
. The Bellman target at input $\bx$ has the affine form $\bR+\gamma\bZ(\bX')$. Rather than estimating the target density directly, we embed the law of this Bellman target in $\mc H_{\mc Z}$. For any candidate return process $\bZ=\{\bZ(\bx):\bx\in\mc X \}$, let $\Q_\bZ(\cdot\mid\bx):=\mc L\big({\bZ(\bx)}\big)$ denote its conditional return-law map. The conditional mean embedding of the candidate law at $\bx$ is $\mu_{\bZ\mid\bx}(\cdot):=\int_{\mc Z}
k_{\mc Z}(\bz,\cdot)\ d\Q_\bZ(\bz\mid\bx)$.

\noindent For a reward value $\br\in\mc R$ and the discount $\gamma\in(0,1)$, define the affine Bellman map and the associated pushforward measure
\begin{equation}\label{eq:pushforward_map}
\theta_{\br,\gamma}(\bz):=\br+\gamma\bz\ ,
\qquad
(\theta_{\br,\gamma})_{\#}\Q:=\Q\circ\theta_{\br,\gamma}^{-1}
\end{equation}
for any probability law $\Q$ on $\mc Z$; that is, $(\theta_{\br,\gamma})_{\#}\Q$ is the law of $\br+\gamma\bZ$ when $\bZ\sim\Q$. In this notation, conditioning on fixed one-step values $(\bR,\bX')=(\br,\bx')$, the law of the Bellman target is exactly the pushforward $(\theta_{\br,\gamma})_{\#}\Q_\bZ(\cdot\mid\bx')$, and the conditional law produced by the distributional Bellman operator is the mixture
\begin{equation}\label{eq:pushforward_bellman_law_supp}
(\T^\pi\Q_\bZ)(\cdot\mid\bx)
=
\int_{\mc R\times\mc X}
(\theta_{\br,\gamma})_{\#}\Q_\bZ(\cdot\mid\bx')\
\mathcal{M}^\pi(d\br,d\bx'\mid\bx)
\end{equation}
The corresponding \emph{pushforward mean embedding} is the mean embedding $\mu_{(\theta_{\br,\gamma})_{\#}\Q}$ of the pushforward law. A change of variables gives, for every $g\in\mc H_{\mc Z}$,
\[
\big\langle g,\ \mu_{(\theta_{\br,\gamma})_{\#}\Q}\big\rangle_{\mc H_{\mc Z}}
=
\E_{\Q}\{g(\br+\gamma\bZ)\},
\qquad\text{so that}\qquad
\mu_{(\theta_{\br,\gamma})_{\#}\Q}(\cdot)
=
\int_{\mc Z}k_{\mc Z}(\br+\gamma\bz,\cdot)\ d\Q(\bz)
\]
In words, the Bellman update transports each continuation return $\bz$ to $\br+\gamma\bz$ before embedding, replacing the kernel section $k_{\mc Z}(\bz,\cdot)$ by the reward-shifted section $k_{\mc Z}(\br+\gamma\bz,\cdot)$. 
For a reward value $\br\in\mc R$ and the discount $\gamma\in(0,1)$, define the affine Bellman map and the associated pushforward measure
\begin{equation}\label{eq:pushforward_map}
\theta_{\br,\gamma}(\bz):=\br+\gamma\bz\ ,
\qquad
(\theta_{\br,\gamma})_{\#}\Q:=\Q\circ\theta_{\br,\gamma}^{-1}
\end{equation}

Recall the affine Bellman map $\theta_{\br,\gamma}(\bz)=\br+\gamma\bz$ in \eqref{eq:pushforward_map} and the composition operator $\mc{C}_{\br,\gamma}g:=g\circ\theta_{\br,\gamma}$. Throughout, $k_{\mc Z}$ is a Mat\'ern kernel on $\R^d$ with smoothness $\nu>1$. Its RKHS $\mc H_{\mc Z}$ is norm-equivalent to the Sobolev space $H^{\nu+d/2}(\R^d)$ \citep{wendland2004scattered}: there exist constants $0<C_1\le C_2<\infty$ such that
\begin{equation}\label{eq:sobolev_equivalence}
C_1\|f\|_{H^{\nu+d/2}}\le\|f\|_{\mc H_{\mc Z}}\le C_2\|f\|_{H^{\nu+d/2}}
\qquad\text{for all }f\in\mc H_{\mc Z}=H^{\nu+d/2}(\R^d)
\end{equation}

\begin{lemma}[Pushforward mean embedding]\label{lem:pushforward_operator}
Let $k_{\mc Z}$ be a Mat\'ern kernel on $\R^d$ with smoothness $\nu>1$, and let $\gamma\in(0,1]$. Then:
\begin{enumerate}[label=(\roman*)]
\vspace{-1em}\item For every $\br\in\R^d$, the composition operator $\mc{C}_{\br,\gamma}:\mc H_{\mc Z}\to\mc H_{\mc Z}$ is well defined and bounded, with
\[
\|\mc{C}_{\br,\gamma}\|_{\mathrm{op}}
\le
c_\gamma:=\frac{C_2}{C_1}\,\gamma^{-d/2}
\]
where $C_1,C_2$ are the equivalence constants in \eqref{eq:sobolev_equivalence}. The bound does not depend on $\br$.
\vspace{-1em}\item For every Borel probability law $\Q$ on $\R^d$, the pushforward law $(\theta_{\br,\gamma})_{\#}\Q$ is embeddable and its mean embedding is the image of $\mu_{\Q}$ under the adjoint of the composition operator:
\[
\mu_{(\theta_{\br,\gamma})_{\#}\Q}
=
\mc{C}_{\br,\gamma}^{*}\,\mu_{\Q},
\qquad
\big\|\mu_{(\theta_{\br,\gamma})_{\#}\Q}\big\|_{\mc H_{\mc Z}}\le c_\gamma\|\mu_{\Q}\|_{\mc H_{\mc Z}}
\]
\end{enumerate}
\end{lemma}

\begin{proof}
(i) Write $s=\nu+d/2$ and let $g\in\mc H_{\mc Z}=H^{s}(\R^d)$. Decompose $\theta_{\br,\gamma}$ as the dilation $\bz\mapsto\gamma\bz$ followed by the translation $\bz\mapsto\bz+\br$. Translation leaves the modulus of the Fourier transform unchanged, so it is an isometry of $H^{s}$. For the dilation, let $h(\bz):=g(\gamma\bz)$, so that $\widehat h(\bu)=\gamma^{-d}\widehat g(\bu/\gamma)$. Substituting $\mathbf v=\bu/\gamma$,
\[
\|h\|_{H^s}^2
=
\int_{\R^d}\gamma^{-2d}\,|\widehat g(\bu/\gamma)|^2(1+\|\bu\|_2^2)^s\,d\bu
=
\gamma^{-d}
\int_{\R^d}|\widehat g(\mathbf v)|^2(1+\gamma^2\|\mathbf v\|_2^2)^s\,d\mathbf v
\le
\gamma^{-d}\|g\|_{H^s}^2
\]
where the inequality uses $\gamma\le1$ together with $s\ge0$. Hence $\|\mc{C}_{\br,\gamma}g\|_{H^s}\le\gamma^{-d/2}\|g\|_{H^s}$, and the norm equivalence \eqref{eq:sobolev_equivalence} yields the chain
\[
\|\mc{C}_{\br,\gamma}g\|_{\mc H_{\mc Z}}
\le
C_2\|\mc{C}_{\br,\gamma}g\|_{H^{s}}
\le
C_2\,\gamma^{-d/2}\|g\|_{H^{s}}
\le
\frac{C_2}{C_1}\,\gamma^{-d/2}\|g\|_{\mc H_{\mc Z}}
\]
which is the stated operator bound, uniformly in $\br$.

(ii) Since $k_{\mc Z}$ is bounded, both $\mu_{\Q}$ and $\mu_{(\theta_{\br,\gamma})_{\#}\Q}$ exist as Bochner integrals in $\mc H_{\mc Z}$. For every $g\in\mc H_{\mc Z}$, the change-of-variables formula for pushforward measures and the reproducing property give
\[
\big\langle g,\ \mu_{(\theta_{\br,\gamma})_{\#}\Q}\big\rangle_{\mc H_{\mc Z}}
=
\int_{\R^d}g\ d\big\{(\theta_{\br,\gamma})_{\#}\Q\big\}
=
\int_{\R^d}g(\br+\gamma\bz)\ d\Q(\bz)
=
\big\langle \mc{C}_{\br,\gamma}g,\ \mu_{\Q}\big\rangle_{\mc H_{\mc Z}}
=
\big\langle g,\ \mc{C}_{\br,\gamma}^{*}\mu_{\Q}\big\rangle_{\mc H_{\mc Z}}
\]
Since $g$ is arbitrary, $\mu_{(\theta_{\br,\gamma})_{\#}\Q}=\mc{C}_{\br,\gamma}^{*}\mu_{\Q}$, and the norm bound follows from $\|\mc{C}_{\br,\gamma}^{*}\|_{\mathrm{op}}=\|\mc{C}_{\br,\gamma}\|_{\mathrm{op}}\le c_\gamma$.
\end{proof}

\noindent Lemma~\ref{lem:pushforward_operator} has two consequences used below. First, the Bellman target embedding in \eqref{eq:pushforward_bellman_eb} is well defined whenever the candidate embeddings are uniformly bounded because the integrand $\mc{C}_{\br,\gamma}^{*}\mu_{\bZ\mid\bx'}$ is bounded in norm by $c_\gamma\sup_{\bx'}\|\mu_{\bZ\mid\bx'}\|_{\mc H_{\mc Z}}$. Second, for every $g\in\mc H_{\mc Z}$ and reward value $\br$, the shifted and rescaled function $g_{\br,\gamma}:=\mc{C}_{\br,\gamma}g=g(\br+\gamma\,\cdot)$ belongs to $\mc H_{\mc Z}$ and satisfies $\|g_{\br,\gamma}\|_{\mc H_{\mc Z}}\le c_\gamma\|g\|_{\mc H_{\mc Z}}$ uniformly in $\br$. This bound provides the domination used in Lemma~\ref{lemma:weak_cts_embed_operator}. Together, these facts show that the population Bellman residual in \eqref{eq:exact_bellman_discrepancy} is the $\mc H_{\mc Z}$-norm distance between the candidate embedding at $\bx$ and its Bellman target embedding, the $\mathcal{M}^\pi$-average of $\mc{C}_{\br,\gamma}^{*}\mu_{\bZ\mid\bx'}$. It is not a discrepancy between two separately estimated distributions.

\noindent The operator bound uses the Mat\'ern kernel only through the norm equivalence in \eqref{eq:sobolev_equivalence}. Translation is an isometry of $H^s$, and the dilation argument requires only $\gamma\le1$ and $s\ge0$. The same argument and the bound $c_\gamma=(C_2/C_1)\gamma^{-d/2}$ therefore apply to any translation-invariant kernel whose RKHS is norm-equivalent to $H^s(\R^d)$ for some $s\ge0$. Existence of the two mean embeddings in part (ii) requires only a bounded measurable kernel; the adjoint representation additionally uses boundedness of the composition operator established in part (i).

\noindent Gaussian kernels also satisfy the composition-operator conclusion. Their spectral densities are radial and nonincreasing, so the spectral characterization of the translation-invariant RKHS norm gives the bound $\gamma^{-d/2}$ under dilation \citep{wendland2004scattered}. They are excluded from the fixed-point analysis only because their exponentially decaying Fourier transforms do not satisfy condition~\ref{cond:Fourier_Decay}, which is required for the MMD-Wasserstein comparison in Theorem~\ref{thm:mmd_wasserstein_equiv}. Finally, Lemma~\ref{lem:pushforward_operator} itself requires only $\nu>0$. The stronger restriction $\nu>1$ is used to obtain the finite feature-map Lipschitz constant $L_k$ in Theorem~\ref{thm:fixed_point}.

\subsection{Comparison of MMD and Wasserstein Distance} \label{app:Wass_Control_MMD}

Let $\mathcal P(\mc Z)$ denote the set of Borel probability measures on a metric space $(\mc Z,d)$. For a class $\mc F$ of bounded measurable test functions, the associated integral probability metric is
\[
\Bgamma_{\mc F}(\mathbb P,\mathbb Q)
:=
\sup_{f\in\mc F}
\left|
\int_{\mc Z} f\,d\mathbb P
-
\int_{\mc Z} f\,d\mathbb Q
\right|
\]
Two choices of $\mc F$ are important here. If $\mc F$ is the set of $1$-Lipschitz functions, then $\Bgamma_{\mc F}$ is the $1$-Wasserstein distance $W_1$ by the Kantorovich-Rubinstein duality \citep{Dudley_2002}. If $\mc F$ is the unit ball of an RKHS $\mc H_{\mc Z}$ with kernel $k_{\mc Z}$, then $\Bgamma_{\mc F}$ is the maximum mean discrepancy, denoted $\Bgamma_k$ \citep{gretton2006kernel,smola2007hilbert}. In this case,
\[
\Bgamma_k(\mathbb P,\mathbb Q)
=
\left\|
\int_{\mc Z}k_{\mc Z}(\cdot,\bz)\,d\mathbb P(\bz)
-
\int_{\mc Z}k_{\mc Z}(\cdot,\bz)\,d\mathbb Q(\bz)
\right\|_{\mc H_{\mc Z}}
\]
When $k_{\mc Z}$ is characteristic, $\Bgamma_k(\mathbb P,\mathbb Q)=0$ if and only if $\mathbb P=\mathbb Q$. The result below states the two comparisons used in the fixed-point argument: the lower comparison follows from Lipschitz continuity of the kernel feature map, while the upper H\"older comparison follows from the regular-density result of \citet{vayer2023controlling}.

\begin{theorem}[MMD and Wasserstein comparison] \label{thm:mmd_wasserstein_equiv}
Let $\mc Z\subseteq\R^d$ be equipped with the Euclidean distance, and let
$k_{\mc Z}(\bz,\bz')=\psi(\bz-\bz')$ be a continuous, bounded, translation-invariant, positive definite kernel on $\R^d$ with RKHS $\mc H_{\mc Z}$. Assume the following conditions.

\begin{enumerate}[label=(\Roman*)]
\vspace{-1em}\item \label{asmp:1-w-MDD} \textit{Feature-map Lipschitz continuity.}
There exists $L_k<\infty$ such that, for all $\bz,\bz'\in\mc Z$,
\[
\|k_{\mc Z}(\cdot,\bz)-k_{\mc Z}(\cdot,\bz')\|_{\mc H_{\mc Z}}
\le
L_k\|\bz-\bz'\|_2
\]

\vspace{-1em}\item \label{cond:Fourier_Decay} \textit{Polynomial Fourier decay.}
Let $\widehat\psi$ denote the Fourier transform of $\psi$. Assume $\psi\in L^1(\R^d)$, $\widehat\psi(\bu)>0$ for every $\bu\in\R^d$, and there exists $s_k>0$ such that
\[
\widehat\psi(\bu)^{-1}
=
O(\|\bu\|_2^{s_k})
\qquad
\text{as }\|\bu\|_2\to\infty
\]

\vspace{-1em}\item \textit{Characteristic kernel.}
The kernel $k_{\mc Z}$ is characteristic.

\vspace{-1em}\item \textit{Regular model class.}
Fix $r>1$, $b>0$, $\mathfrak m>0$, and $s\ge s_k/2$. Define
\[
\mathfrak S_{b,\mathfrak m,r,s}
:=
\left\{
\mathbb P\in\mathcal P(\R^d):
\mathbb P=f\,d\bz,\
\|f\|_{H^s(\R^d)}\le b,\
M_r[\mathbb P]\le \mathfrak m
\right\}
\]
where
\[
M_r[\mathbb P]
:=
\left(
\int_{\R^d}\|\bz\|_2^r\,d\mathbb P(\bz)
\right)^{1/r}
\]
\end{enumerate}

Then there exist constants $c_*>0$ and $C_*>0$, depending only on the displayed kernel and model-class constants, such that for all $\mathbb P,\mathbb Q\in\mathfrak S_{b,\mathfrak m,r,s}$,
\begin{equation}\label{eq:W_gamma_equiv}
c_*\,\Bgamma_k(\mathbb P,\mathbb Q)
\le
W_1(\mathbb P,\mathbb Q)
\le
C_*\,\Bgamma_k(\mathbb P,\mathbb Q)^\rho,
\qquad
\rho=\frac{r-1}{d+2r}
\end{equation}
In particular, on $\mathfrak S_{b,\mathfrak m,r,s}$, convergence in $\Bgamma_k$ and convergence in $W_1$ are equivalent.
\end{theorem}

\begin{proof}
For the lower comparison, let $\Upsilon(\mathbb P,\mathbb Q)$ denote the set of couplings of $\mathbb P$ and $\mathbb Q$. By the feature-map Lipschitz assumption, for any coupling $\upsilon\in\Upsilon(\mathbb P,\mathbb Q)$,
\begin{align*}
\Bgamma_k(\mathbb P,\mathbb Q)
&=
\left\|
\int_{\mc Z\times\mc Z}
\{k_{\mc Z}(\cdot,\bz)-k_{\mc Z}(\cdot,\widetilde{\bz})\}
d\upsilon(\bz,\widetilde{\bz})
\right\|_{\mc H_{\mc Z}}\\
&\le
\int_{\mc Z\times\mc Z}
\|k_{\mc Z}(\cdot,\bz)-k_{\mc Z}(\cdot,\widetilde{\bz})\|_{\mc H_{\mc Z}}
d\upsilon(\bz,\widetilde{\bz})\\
&\le
L_k
\int_{\mc Z\times\mc Z}\|\bz-\widetilde{\bz}\|_2\,d\upsilon(\bz,\widetilde{\bz})
\end{align*}
Taking the infimum over couplings gives $\Bgamma_k(\mathbb P,\mathbb Q)\le L_k W_1(\mathbb P,\mathbb Q)$, so the first inequality holds with $c_*=1/L_k$.

\noindent For the upper comparison, apply Theorem 15 of \citet{vayer2023controlling} with $p=1$. Under the Fourier-decay condition and the Sobolev/moment constraints defining $\mathfrak S_{b,\mathfrak m,r,s}$, that theorem gives a constant $C_*>0$ such that
\[
W_1(\mathbb P,\mathbb Q)
\le
C_*\,\Bgamma_k(\mathbb P,\mathbb Q)^{(r-1)/(d+2r)}
\]
for every $\mathbb P,\mathbb Q\in\mathfrak S_{b,\mathfrak m,r,s}$. Combining the two inequalities proves \eqref{eq:W_gamma_equiv}. The final statement follows immediately from the two-sided comparison.
\end{proof}

\noindent For the Mat\'ern class with smoothness $\nu$ and length scale $\ell$, the spectral density is proportional to $(2\nu/\ell^2+\|\bu\|_2^2)^{-(\nu+d/2)}$ \citep{wendland2004scattered}, so $\widehat\psi(\bu)^{-1}=O(\|\bu\|_2^{2\nu+d})$ and the Fourier condition \ref{cond:Fourier_Decay} holds with
\[
s_k=2\nu+d
\]
Accordingly, the comparison is applied to densities with Sobolev smoothness $s\ge s_k/2=\nu+d/2$, the smoothness order of the Mat\'ern RKHS. Strict positivity of the spectral density also makes the Mat\'ern kernel characteristic \citep{sriperumbudur2010hilbert}, which verifies condition (III). Gaussian kernels are characteristic, but their Fourier transforms decay exponentially and therefore do not satisfy the polynomial reciprocal-growth condition used in the upper H\"older bound. Supplementary Material~\ref{app:matern_verification} verifies conditions \ref{asmp:1-w-MDD}-(IV) for the Mat\'ern family, including the monotonicity result used to obtain $L_k$.
\subsection{Fixed Points of Weakly Sequentially Continuous Mappings} \label{app:Fixed_Point_Holder_Maps}
\begin{lemma}[{\citet{arino1984fixedpoint}}]\label{lem:AGP}
Let $K$ be a nonempty, weakly compact, convex subset of a Banach space $X$, and let $T:K\to K$ be weakly sequentially continuous. Then $T$ has a fixed point in $K$.
\end{lemma}
\noindent In a reflexive Banach space, every nonempty, closed, bounded, convex set is weakly compact. Mazur's theorem gives weak closedness, and the Banach-Alaoglu theorem together with reflexivity gives relative weak compactness (Kakutani's theorem). By the Eberlein-\v{S}mulian theorem, weak compactness is equivalent to weak sequential compactness, as required by Lemma~\ref{lem:AGP}. Hence the lemma applies to the set $K$ in Assumption~\ref{ass:regular_bellman_embedding_class}; metric contractivity of $T$ in the ambient norm is not needed for existence.
\begin{remark}\label{rem:barroso}
Theorem~3.1 of \citet{BARROSO2023_holderfixedpoint} provides an alternative that does not require $K$ to be bounded. A weakly sequentially continuous self-map of a closed convex subset of a reflexive space that is $\alpha$-H\"older Lipschitz in the ambient norm, $\alpha\in(0,1)$, has a fixed point if and only if it has a bounded orbit. Applying that result would additionally require a H\"older comparison between the ambient norm and the embedding discrepancy $D_k$ on $K$. Because this condition is restrictive for supremum-type discrepancies in infinite dimensions, the proof of Theorem~\ref{thm:fixed_point} uses the weak-compactness result in Lemma~\ref{lem:AGP}.
\end{remark}

\subsection{Weak Sequential Continuity of the Bellman Operator on Conditional Embeddings}
\begin{lemma}[Weak sequential continuity of the induced embedding operator]\label{lemma:weak_cts_embed_operator}
Let $\mc X:=\mc S\times\mc A$, and let $k_{\mc Z}$ be a bounded continuous Mat\'ern kernel on $\mc Z$ with RKHS $\mc H_{\mc Z}$ and $\sup_{\bz\in\mc Z} k_{\mc Z}(\bz,\bz)\le \sigma^2<\infty$. Let $\mc E$ be the reflexive Banach space of admissible conditional embedding maps used in Theorem~\ref{thm:fixed_point}, where each $\mu\in K\subset\mc E$ corresponds to a regular conditional law $\mathbb P_\mu(\cdot\mid\bx)$ satisfying
\[
\mu(\bx)
=
\int_{\mc Z} k_{\mc Z}(\cdot,\bz)\,d\mathbb P_\mu(\bz\mid\bx),
\qquad
\bx\in\mc X
\]
Assume that $K$ is invariant under the induced Bellman embedding operator $\mathfrak T^\pi$ and that embeddings in $K$ are uniformly bounded pointwise:
\[
\sup_{\mu\in K}\sup_{\bx\in\mc X}\|\mu(\bx)\|_{\mc H_{\mc Z}}\le \sigma
\]
Assume also that weak convergence in $\mc E$ implies pointwise weak convergence in $\mc H_{\mc Z}$ and that the relative weak topology on $K$ is generated by the pointwise dual evaluations $\mu\mapsto\langle g,\mu(\bx)\rangle_{\mc H_{\mc Z}}$, $g\in\mc H_{\mc Z}$ and $\bx\in\mc X$.

For every $g\in\mc H_{\mc Z}$ and reward value $\br$, write $g_{\br,\gamma}:=\mc{C}_{\br,\gamma}g=g(\br+\gamma\,\cdot)$, and let $\mathcal{M}^\pi(d\br,d\bx'\mid\bx)$ denote the joint target-policy one-step law of $(\bR,\bX')$ given $\bX=\bx$. By Lemma~\ref{lem:pushforward_operator}, $g_{\br,\gamma}\in\mc H_{\mc Z}$ with the uniform bound $\|g_{\br,\gamma}\|_{\mc H_{\mc Z}}\le c_\gamma\|g\|_{\mc H_{\mc Z}}$ for every $\br$; no separate envelope hypothesis is required.
Then $\mathfrak T^\pi:K\to K$ is weakly sequentially continuous. That is, if $\mu_n\rightharpoonup\mu$ weakly in $\mc E$, with $\mu_n,\mu\in K$, then $\mathfrak T^\pi\mu_n\rightharpoonup\mathfrak T^\pi\mu$ weakly in $\mc E$.
\end{lemma}

\begin{proof}
For $\mu\in K$, the induced Bellman embedding at the current input $\bx=(\bs^\top,\ba^\top)^\top$ is
\[
(\mathfrak T^\pi\mu)(\bx)
:=
\int_{\mc R\times\mc X}\int_{\mc Z}
k_{\mc Z}(\cdot,\br+\gamma\bz)\,
d\mathbb P_\mu(\bz\mid\bx')
\mathcal{M}^\pi(d\br,d\bx'\mid\bx)
\]
where $\mathcal{M}^\pi$ keeps the realized reward and successor input paired. Let $\mu_n\rightharpoonup\mu$ in $\mc E$, with all elements in $K$. To prove weak sequential continuity on $K$, by the assumed relative weak topology it is enough to show that, for every fixed $\bx\in\mc X$ and every $g\in\mc H_{\mc Z}$,
\[
\left\langle g,(\mathfrak T^\pi\mu_n)(\bx)\right\rangle_{\mc H_{\mc Z}}
\longrightarrow
\left\langle g,(\mathfrak T^\pi\mu)(\bx)\right\rangle_{\mc H_{\mc Z}}
\]
Using the reproducing property and Fubini's theorem,
\begin{align*}
\left\langle g,(\mathfrak T^\pi\mu_n)(\bx)\right\rangle_{\mc H_{\mc Z}}
&=
\int_{\mc R\times\mc X}\int_{\mc Z}
g(\br+\gamma\bz)\,
d\mathbb P_{\mu_n}(\bz\mid\bx')
\mathcal{M}^\pi(d\br,d\bx'\mid\bx)\\
&=
\int_{\mc R\times\mc X}
\left\langle g_{\br,\gamma},\mu_n(\bx')\right\rangle_{\mc H_{\mc Z}}
\mathcal{M}^\pi(d\br,d\bx'\mid\bx)
\end{align*}
For fixed $(\bx',\br)$, membership $g_{\br,\gamma}\in\mc H_{\mc Z}$ (Lemma~\ref{lem:pushforward_operator}) and the pointwise weak convergence $\mu_n(\bx')\rightharpoonup\mu(\bx')$ in $\mc H_{\mc Z}$ imply
\[
\left\langle g_{\br,\gamma},\mu_n(\bx')\right\rangle_{\mc H_{\mc Z}}
\longrightarrow
\left\langle g_{\br,\gamma},\mu(\bx')\right\rangle_{\mc H_{\mc Z}}
\]
Moreover, the uniform embedding bound on $K$ and the uniform operator bound of Lemma~\ref{lem:pushforward_operator} give the domination
\[
\left|
\left\langle g_{\br,\gamma},\mu_n(\bx')\right\rangle_{\mc H_{\mc Z}}
\right|
\le
\|g_{\br,\gamma}\|_{\mc H_{\mc Z}}\|\mu_n(\bx')\|_{\mc H_{\mc Z}}
\le
\sigma c_\gamma\|g\|_{\mc H_{\mc Z}}
\]
The right-hand side is a finite constant and is therefore integrable under $\mathcal{M}^\pi(d\br,d\bx'\mid\bx)$ for every current input $\bx$. The integrand is jointly measurable in $(\br,\bx')$: for $\nu>0$ the Mat\'ern RKHS embeds continuously into the bounded continuous functions, so $(\br,\bz)\mapsto g(\br+\gamma\bz)$ is bounded and jointly continuous, and $\bx'\mapsto\mathbb P_{\mu_n}(\cdot\mid\bx')$ is a measurable probability kernel, so $(\br,\bx')\mapsto\int g(\br+\gamma\bz)\,d\mathbb P_{\mu_n}(\bz\mid\bx')$ is measurable. Therefore dominated convergence yields
\[
\left\langle g,(\mathfrak T^\pi\mu_n)(\bx)\right\rangle_{\mc H_{\mc Z}}
\longrightarrow
\left\langle g,(\mathfrak T^\pi\mu)(\bx)\right\rangle_{\mc H_{\mc Z}}
\]
for every $\bx\in\mc X$ and $g\in\mc H_{\mc Z}$. By the assumed characterization of the relative weak topology on $K$, this pointwise weak convergence of all dual evaluations is exactly $\mathfrak T^\pi\mu_n\rightharpoonup\mathfrak T^\pi\mu$ in $\mc E$.
\end{proof}

\subsection{A Conditional Vector-Valued RKHS Construction for Assumption~\ref{ass:regular_bellman_embedding_class}}\label{app:vvRKHS_example}

This subsection gives a conditional construction of the pair $(\mc E,K)$. The vector-valued RKHS and the compatibility condition below establish nonemptiness and the topological clauses of Assumption~\ref{ass:regular_bellman_embedding_class}. The ambient part of Bellman invariance remains a model-specific assumption. Let $\mc E$ be the vector-valued RKHS of $\mc H_{\mc Z}$-valued maps on $\mc X$ induced by the operator-valued kernel
\[
\mathbf K(\bx,\bx')
:=
 k_{\mc X}(\bx,\bx')\,\mathrm{Id}_{\mc H_{\mc Z}}
\]
with $ k_{\mc X}$ bounded and continuous. The space $\mc E$ is Hilbert and therefore reflexive. Point evaluations are bounded: for every $\bx\in\mc X$ and $g\in\mc H_{\mc Z}$, the map $\mu\mapsto\langle g,\mu(\bx)\rangle_{\mc H_{\mc Z}}=\langle\mu,\mathbf K(\cdot,\bx)g\rangle_{\mc E}$ is a bounded linear functional. Thus, weak convergence $\mu_n\rightharpoonup\mu$ in $\mc E$ implies $\mu_n(\bx)\rightharpoonup\mu(\bx)$ in $\mc H_{\mc Z}$ for every $\bx$, which gives the first part of clause (iii).

The linear span of $\{\mathbf K(\cdot,\bx)g:\bx\in\mc X,\ g\in\mc H_{\mc Z}\}$ is dense in $\mc E$. On norm-bounded subsets of a Hilbert space, weak convergence is equivalent to convergence against any total family of functionals. Therefore, on the bounded set $K$ below, the relative weak topology is generated by the pointwise dual evaluations $\mu\mapsto\langle g,\mu(\bx)\rangle_{\mc H_{\mc Z}}$. This gives the second part of clause (iii).

\noindent For the admissible set, fix a radius $R_K>0$ and impose the following compatibility condition: there exists a measurable conditional law map $\mathbb P_0$ with $\mathbb P_0(\cdot\mid\bx)\in\mathfrak S_{b,\mathfrak m,r,s}$ for every $\bx$, whose embedding map $\mu_0(\bx):=\mu_{\mathbb P_0(\cdot\mid\bx)}$ belongs to $\mc E$ and satisfies $\|\mu_0\|_{\mc E}\le R_K$. Define
\[
K
:=
\left\{
\mu\in\mc E:
\begin{array}{l}
\|\mu\|_{\mc E}\le R_K,\ \text{and there exists a measurable conditional law map }\mathbb P_\mu,\\
\mu(\bx)=\mu_{\mathbb P_\mu(\cdot\mid\bx)},\quad
\mathbb P_\mu(\cdot\mid\bx)\in\mathfrak S_{b,\mathfrak m,r,s}\ \text{for every }\bx\in\mc X
\end{array}
\right\}
\]
The compatibility condition gives $\mu_0\in K$, so $K$ is nonempty. The set $K$ is convex because the norm ball and the class $\mathfrak S_{b,\mathfrak m,r,s}$ are convex and the embedding map is affine in the law: for $t\in[0,1]$ and laws $\mathbb P_1=f_1\,d\bz$ and $\mathbb P_2=f_2\,d\bz$ in $\mathfrak S_{b,\mathfrak m,r,s}$, the mixture $t\mathbb P_1+(1-t)\mathbb P_2$ has density $tf_1+(1-t)f_2$ with $\|tf_1+(1-t)f_2\|_{H^s}\le tb+(1-t)b=b$, its $r$-th moment satisfies $\int\|\bz\|_2^r\,d\{t\mathbb P_1+(1-t)\mathbb P_2\}=t\,M_r[\mathbb P_1]^r+(1-t)\,M_r[\mathbb P_2]^r\le\mathfrak m^r$, and $\mu_{t\mathbb P_1+(1-t)\mathbb P_2}=t\mu_{\mathbb P_1}+(1-t)\mu_{\mathbb P_2}$. It is bounded by construction.

\noindent To prove norm closedness, let $\mu_n\to\mu$ in $\mc E$ with $\mu_n\in K$, and let $\mathbb P_n$ denote corresponding measurable conditional law maps. The argument proceeds in five steps.

\noindent\textit{Step 1: pointwise norm convergence.} For every $\bx\in\mc X$ and $g\in\mc H_{\mc Z}$, the reproducing identity gives
\[
\langle g,\mu_n(\bx)-\mu(\bx)\rangle_{\mc H_{\mc Z}}
=
\langle\mu_n-\mu,\mathbf K(\cdot,\bx)g\rangle_{\mc E},
\]
while
\[
\|\mathbf K(\cdot,\bx)g\|_{\mc E}^2
=
\langle g,\mathbf K(\bx,\bx)g\rangle_{\mc H_{\mc Z}}
=
 k_{\mc X}(\bx,\bx)\|g\|_{\mc H_{\mc Z}}^2.
\]
Consequently,
\[
\|\mu_n(\bx)-\mu(\bx)\|_{\mc H_{\mc Z}}
\le
 k_{\mc X}(\bx,\bx)^{1/2}\|\mu_n-\mu\|_{\mc E}
\longrightarrow0
\]

\noindent\textit{Step 2: tightness and subsequential law limits.} Fix $\bx\in\mc X$. Since $\sup_nM_r[\mathbb P_n(\cdot\mid\bx)]\le\mathfrak m$ with $r>1$, Markov's inequality gives $\sup_n\mathbb P_n(\{\bz:\|\bz\|_2>t\}\mid\bx)\le(\mathfrak m/t)^{r}$ for every $t>0$, so the family $\{\mathbb P_n(\cdot\mid\bx)\}_{n\ge1}$ is tight. By Prokhorov's theorem, every subsequence contains a further subsequence along which $\mathbb P_{n_k}(\cdot\mid\bx)$ converges weakly to some law $\mathbb P^{(\bx)}$.

\noindent\textit{Step 3: the subsequential limit lies in $\mathfrak S_{b,\mathfrak m,r,s}$.} Along the subsequence of Step 2, the densities $f_{n_k}$ of $\mathbb P_{n_k}(\cdot\mid\bx)$ lie in the closed ball of radius $b$ in $H^s(\R^d)$, which is weakly compact; pass to a further subsequence with $f_{n_k}\rightharpoonup f$ weakly in $H^s(\R^d)$. For every $\varphi\in C_c^\infty(\R^d)$, the functional $h\mapsto\int\varphi h\,d\bz$ is bounded on $H^s(\R^d)$, so $\int\varphi f_{n_k}\,d\bz\to\int\varphi f\,d\bz$; on the other hand, $\int\varphi\,d\mathbb P_{n_k}(\cdot\mid\bx)\to\int\varphi\,d\mathbb P^{(\bx)}$ by weak convergence of laws. Hence $\int\varphi\,d\mathbb P^{(\bx)}=\int\varphi f\,d\bz$ for every $\varphi\in C_c^\infty(\R^d)$, so $\mathbb P^{(\bx)}$ has Lebesgue density $f$. Weak lower semicontinuity of the $H^s$ norm gives $\|f\|_{H^s}\le\liminf_k\|f_{n_k}\|_{H^s}\le b$, and the portmanteau theorem applied to the nonnegative lower semicontinuous function $\bz\mapsto\|\bz\|_2^r$ gives $M_r[\mathbb P^{(\bx)}]^r\le\liminf_kM_r[\mathbb P_{n_k}(\cdot\mid\bx)]^r\le\mathfrak m^r$. Hence $\mathbb P^{(\bx)}\in\mathfrak S_{b,\mathfrak m,r,s}$.

\noindent\textit{Step 4: identification of the limit and convergence of the full sequence.} Mat\'ern RKHS functions are bounded and continuous, so weak convergence of laws gives, for every $g\in\mc H_{\mc Z}$,
\[
\langle g,\mu_{n_k}(\bx)\rangle_{\mc H_{\mc Z}}
=
\int g\,d\mathbb P_{n_k}(\cdot\mid\bx)
\longrightarrow
\int g\,d\mathbb P^{(\bx)}
=
\langle g,\mu_{\mathbb P^{(\bx)}}\rangle_{\mc H_{\mc Z}}
\]
that is, $\mu_{n_k}(\bx)\rightharpoonup\mu_{\mathbb P^{(\bx)}}$ weakly in $\mc H_{\mc Z}$. By Step 1, $\mu_{n_k}(\bx)\to\mu(\bx)$ in norm, and the norm limit coincides with the weak limit, so $\mu_{\mathbb P^{(\bx)}}=\mu(\bx)$. Because the kernel is characteristic, $\mathbb P^{(\bx)}$ is the unique law with embedding $\mu(\bx)$. All subsequential limits therefore coincide, and the full sequence converges weakly: $\mathbb P_n(\cdot\mid\bx)\Rightarrow\mathbb P_\mu(\cdot\mid\bx):=\mathbb P^{(\bx)}$, with $\mu(\bx)=\mu_{\mathbb P_\mu(\cdot\mid\bx)}$ and $\mathbb P_\mu(\cdot\mid\bx)\in\mathfrak S_{b,\mathfrak m,r,s}$ for every $\bx\in\mc X$.

\noindent\textit{Step 5: measurability of the limit law map.} Each map $\bx\mapsto\mathbb P_n(\cdot\mid\bx)$ is measurable into $\mathcal P(\R^d)$ equipped with the weak topology, which is a Polish space. By Step 4 these maps converge pointwise on $\mc X$, and a pointwise limit of measurable maps into a metric space is measurable. Hence $\bx\mapsto\mathbb P_\mu(\cdot\mid\bx)$ is a measurable conditional law map. Therefore $\mu\in K$, and $K$ is norm closed.

\noindent A convex norm-closed set is weakly closed by Mazur's theorem, and the closed ball of a reflexive space is weakly compact by Kakutani's theorem; a weakly closed subset of a weakly compact set is weakly compact. Therefore $K$ is weakly compact. Under the stated compatibility condition, clauses (i) and (iii) of Assumption~\ref{ass:regular_bellman_embedding_class} hold.

\noindent In this construction, clause (ii) separates into requirements for the law class and the ambient space. Lemma~\ref{lem:invariance_identification} verifies preservation of the law class under the stated reward conditions. The ambient requirement is that the Bellman update $\bx\mapsto\int_{\mc R\times\mc X}\mc{C}_{\br,\gamma}^{*}\,\mu(\bx')\,\mathcal{M}^\pi(d\br,d\bx'\mid\bx)$ belongs to $\mc E$ and has $\mc E$-norm at most $R_K$. This is a joint regularity condition on the transition kernel and reward law: relative to $ k_{\mc X}$, the map $\bx\mapsto \mathcal{M}^\pi(\cdot\mid\bx)$ must be sufficiently smooth for the updated embedding map to remain in the ball. The condition is assumed because it depends on both the transition kernel and the chosen state-action kernel.

\noindent Joint error analysis for the full finite-dictionary estimator in Algorithm~\ref{alg:rk_drl} and construction of a finite-dimensional parameterization that defines a probability law at every conditioning point are left for future work.

\subsection{Invariance of the Regular Class and Identification of the Fixed Point}\label{app:invariance_identification}

The following lemma gives sufficient conditions stated in terms of the one-step reward law for the law-class part of invariance clause (ii) in Assumption~\ref{ass:regular_bellman_embedding_class}. Under the same conditions, it identifies the fixed point of Theorem~\ref{thm:fixed_point} with the true conditional return law. The Bellman update convolves the rescaled continuation law with the conditional reward density. Because convolution with a probability measure is a Fourier multiplier of modulus at most one, the reward density supplies the required Sobolev smoothness. If the continuation law has an $H^s$ density, rescaling by $\bz\mapsto\gamma\bz$ can inflate its norm by a factor of $\gamma^{-(s+d/2)}$. The proof does not require such a density because convolution with the reward density yields the stated bound for the updated density.

\begin{lemma}[Bellman invariance and identification]\label{lem:invariance_identification}
Let $\gamma\in(0,1)$, $r>1$, $s\ge0$, $b>0$, and $\mathfrak m>0$. Assume:
\begin{enumerate}[label=(\alph*)]
\vspace{-1em}\item for every $\bx\in\mc X$ and $\mc{P}_X^\pi(\cdot\mid\bx)$-almost every $\bx'$, the conditional one-step reward law $F_{\bR}(\cdot\mid\bx,\bx')$ admits a Lebesgue density $f_{\bR}(\cdot\mid\bx,\bx')$ on $\R^d$ that is jointly measurable in $(\br,\bx,\bx')$, and the density kernel $(\bx,\bx')\mapsto f_{\bR}(\cdot\mid\bx,\bx')$ is strongly measurable as a map into the separable space $H^s(\R^d)$, with
\[
\sup_{\bx,\bx'}\|f_{\bR}(\cdot\mid\bx,\bx')\|_{H^s(\R^d)}\le b
\]
\vspace{-2em}\item $\displaystyle\sup_{\bx\in\mc X}\left\{\E\big(\|\bR\|_2^r\mid\bX=\bx\big)\right\}^{1/r}\le(1-\gamma)\mathfrak m$, where the reward moment does not depend on the target policy by the structure of $\mathcal{M}^\pi$;
\vspace{-1em}\item conditional on $(\bX,\bX')=(\bx,\bx')$, the continuation variable entering the Bellman update is independent of the one-step reward, as in Theorem~\ref{thm:DBO-pdf}.
\end{enumerate}
Then the following hold.
\begin{enumerate}[label=(\roman*)]
\vspace{-1em}\item \textit{Invariance.} For any conditional law map $\Q$ with $\sup_{\bx}M_r[\Q(\cdot\mid\bx)]\le \mathfrak m$, the updated map satisfies $(\T^\pi\Q)(\cdot\mid\bx)\in\mathfrak S_{b,\mathfrak m,r,s}$ for every $\bx\in\mc X$. In particular, $\T^\pi$ maps $\mathfrak S_{b,\mathfrak m,r,s}$-valued conditional law maps into $\mathfrak S_{b,\mathfrak m,r,s}$-valued conditional law maps.
\vspace{-1em}\item \textit{Identification.} The true conditional return law map $\bx\mapsto\mc V^\pi_{\bx}=\mc L\{\bZ^\pi(\bx)\}$ of \eqref{eq:return_def_new} belongs to $\mathfrak S_{b,\mathfrak m,r,s}$ for every $\bx$, satisfies the distributional Bellman equation \eqref{eq:DBO_new}, and is the unique fixed point of $\T^\pi$ among conditional law maps with uniformly bounded first moments. Consequently, the fixed point $\mathbb P^\pi$ of Theorem~\ref{thm:fixed_point} coincides with $\mc V^\pi$.
\end{enumerate}
\end{lemma}

\begin{proof}
(i) Fix $\bx\in\mc X$. Conditional on $\bX'=\bx'$, and using the conditional independence in (c), the law of the Bellman target $\bR+\gamma\bZ(\bx')$ is the convolution
\[
f_{\bR}(\cdot\mid\bx,\bx')\ast\Q_{\gamma,\bx'},
\qquad
\Q_{\gamma,\bx'}:=\mc L\{\gamma\bZ(\bx')\}
\]
which admits the density $h_{\bx,\bx'}(\by):=\int_{\R^d}f_{\bR}(\by-\bw\mid\bx,\bx')\,d\Q_{\gamma,\bx'}(\bw)$. Translations act continuously on $H^s(\R^d)$; therefore, condition (a), the measurable probability-kernel property of the conditional law map $\Q$, and measurability of parameterized Bochner integrals imply that $\bx'\mapsto h_{\bx,\bx'}$ is strongly $H^s$-measurable and that its mixture below is well defined. On the Fourier side, $\widehat h_{\bx,\bx'}(\bu)=\widehat f_{\bR}(\bu\mid\bx,\bx')\,\varphi_{\gamma,\bx'}(\bu)$, where $\varphi_{\gamma,\bx'}$ is the characteristic function of $\Q_{\gamma,\bx'}$ and satisfies $|\varphi_{\gamma,\bx'}|\le1$ pointwise. Hence
\[
\|h_{\bx,\bx'}\|_{H^s}
\le
\|f_{\bR}(\cdot\mid\bx,\bx')\|_{H^s}
\le b
\]
The updated law is the mixture $(\T^\pi\Q)(\cdot\mid\bx)=\int h_{\bx,\bx'}\,\mc{P}_X^\pi(d\bx'\mid\bx)$, so Minkowski's integral inequality in $H^s$ gives $\|f_{\T^\pi\Q}(\cdot\mid\bx)\|_{H^s}\le\sup_{\bx'}\|h_{\bx,\bx'}\|_{H^s}\le b$. For the moments, Minkowski's inequality in $L^r$ yields, without any independence requirement,
\[
M_r[(\T^\pi\Q)(\cdot\mid\bx)]
\le
\left\{\E\big(\|\bR\|_2^r\mid\bx\big)\right\}^{1/r}
+\gamma\sup_{\bx'}M_r[\Q(\cdot\mid\bx')]
\le
(1-\gamma)\mathfrak m+\gamma \mathfrak m=\mathfrak m
\]
Together the two displays give $(\T^\pi\Q)(\cdot\mid\bx)\in\mathfrak S_{b,\mathfrak m,r,s}$.

(ii) For the moments of the true return, Minkowski's inequality applied to $\bZ^\pi(\bx)=\sum_{t\ge0}\gamma^t\bR_t$ and condition (b) give
\[
M_r[\mc V^\pi_{\bx}]
\le
\sum_{t\ge0}\gamma^t\sup_{\bx}\left\{\E\big(\|\bR\|_2^r\mid\bx\big)\right\}^{1/r}
\le
\frac{(1-\gamma)\mathfrak m}{1-\gamma}=\mathfrak m
\]
For the density, decompose $\bZ^\pi(\bx)=\bR_0+\gamma\bZ^\pi(\bX_1)$ and condition on the successor input $\bX_1=\bx'$; by the Markov structure and (c), $\bR_0$ and $\bZ^\pi(\bx')$ are conditionally independent given $(\bx,\bx')$, so the argument of part (i), applied with $\Q=\mc V^\pi$ -- an application that requires only the uniform moment bound established in the preceding display -- shows that the updated law $(\T^\pi\mc V^\pi)(\cdot\mid\bx)$ admits a density with $H^s$ norm at most $b$. Since $\mc V^\pi=\T^\pi\mc V^\pi$ by the distributional Bellman equation, the same bound holds for $\mc V^\pi_{\bx}$ itself. The fixed-point property is the distributional Bellman equation \eqref{eq:DBO_new} \citep{rosler1992fixed,bellemare2017distributional}. Uniqueness among conditional law maps with uniformly bounded first moments follows from the coupling contraction $d_1(\T^\pi\mathbb P,\T^\pi\Q)\le\gamma\,d_1(\mathbb P,\Q)$ established in the uniqueness step of the proof of Theorem~\ref{thm:fixed_point}; that coupling argument is self-contained and does not invoke the present lemma, so no circularity arises, and $d_1$ is finite on the stated set. As $\mathbb P^\pi$ and $\mc V^\pi$ are both fixed points in $\mathfrak S_{b,\mathfrak m,r,s}$, and $M_1\le M_r\le \mathfrak m$ on the class, they coincide.
\end{proof}

\subsection{The Mat\'ern Return Kernel: Verification of Conditions}\label{app:matern_verification}

This subsection verifies the Mat\'ern-kernel properties used in Theorems~\ref{thm:fixed_point} and~\ref{thm:mmd_wasserstein_equiv}.

\begin{figure}[t]
    \centering
    \includegraphics[width=\textwidth]{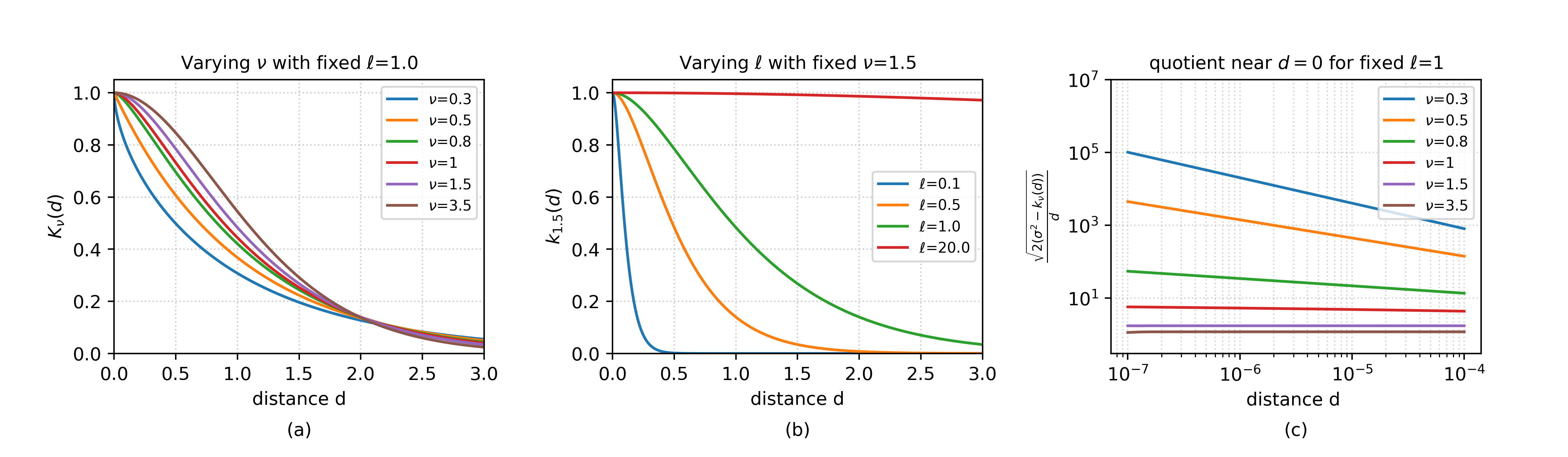}
    \caption{The Mat\'ern kernel as a function of the distance $t=\|\bz-\bz'\|$: (a) varying smoothness $\nu$ with fixed length scale $\ell=1$; (b) varying $\ell$ with fixed $\nu=1.5$; and (c) the quotient $\sqrt{2\{\sigma^2-k_\nu(t)\}}/t$ near $t=0$ for fixed $\ell=1$. The supremum in panel (c) defines the feature-map Lipschitz constant $L_k$ used in Theorem~\ref{thm:fixed_point}.}
    \label{fig:matern}
\end{figure}

\noindent\textit{Condition \ref{asmp:1-w-MDD}, feature-map Lipschitz continuity.} The Mat\'ern kernel is
\[
k_{\nu}(t)
=
\sigma^2
\frac{2^{1-\nu}}{\Gamma(\nu)}
\left(\sqrt{2\nu}\frac{t}{\ell}\right)^\nu
\mathcal K_{\nu}\left(\sqrt{2\nu}\frac{t}{\ell}\right)
\]
where $t=\|\bz-\widetilde{\bz}\|$, $\nu>0$ is the smoothness parameter, $\ell>0$ is the length scale, $\sigma^2>0$ is the variance parameter, and $\mathcal K_\nu$ is the modified Bessel function of the second kind. The squared feature-map distance satisfies
\[
\begin{aligned}
\bigl\|k_\nu(\cdot,\bz)-k_\nu(\cdot,\widetilde{\bz})\bigr\|_{\mc H_{\mc Z}}^2
&=k_\nu(\bz,\bz)+k_\nu(\widetilde{\bz},\widetilde{\bz})
-2k_\nu(\bz,\widetilde{\bz})
\\
&=2\{\sigma^2-k_\nu(\|\bz-\widetilde{\bz}\|)\}
\end{aligned}
\]
Therefore,
\begin{align*}
L_k
&:=
\sup_{\bz\ne\widetilde{\bz}}
\frac{\bigl\|k_\nu(\cdot,\bz)-k_\nu(\cdot,\widetilde{\bz})\bigr\|_{\mc H_{\mc Z}}}
{\|\bz-\widetilde{\bz}\|}
\\
&=
\sup_{t>0}
\frac{\sqrt{2\{\sigma^2-k_\nu(t)\}}}{t}
=
\frac{2\sigma\sqrt{\nu}}{\ell}
\sqrt{
\sup_{u>0}
\frac{
1-\frac{2^{1-\nu}}{\Gamma(\nu)}u^\nu\mathcal K_\nu(u)
}{u^2}
}
\end{align*}
where $u:=\sqrt{2\nu}t/\ell$. Let $c_\nu:=2^{1-\nu}/\Gamma(\nu)$ and $B_\nu(u):=\{1-c_\nu u^\nu\mathcal K_\nu(u)\}/u^2$. Lemma~\ref{lem:Bnu_monotone} below shows that $B_\nu$ is nonincreasing on $(0,\infty)$. For $\nu>1$,
\begin{align*}
\sup_{u>0}B_\nu(u)
&=
\lim_{u\to0}
\frac{1-c_\nu u^\nu\mathcal K_\nu(u)}{u^2}
\\
&\overset{\text{L'H\^opital}}{=}
\lim_{u\to0}
\frac{-c_\nu\{\nu u^{\nu-1}\mathcal K_\nu(u)+u^\nu(d/du)\mathcal K_\nu(u)\}}{2u}
\\
&=
\lim_{u\to0}
\frac{c_\nu u^{\nu-1}\mathcal K_{\nu-1}(u)}{2}
=
\frac{c_\nu}{2c_{\nu-1}}
=
\frac{1}{4(\nu-1)}
\end{align*}
The third equality uses
\[
\frac{d}{du}\mathcal K_{\nu}(u)
=-\frac{\nu}{u}\mathcal K_{\nu}(u)-\mathcal K_{\nu-1}(u)
\]
\citep[Section~9.6.26]{abramowitz1948handbook}. The fourth uses the small-argument asymptotic
\[
\mathcal K_{\nu-1}(u)
\sim\frac{1}{2}\Gamma(\nu-1)(u/2)^{-(\nu-1)},
\qquad u\to0
\]
which is valid for $\nu>1$ \citep[Section~9.6.9]{abramowitz1948handbook}. The final equality uses $\Gamma(\nu)=(\nu-1)\Gamma(\nu-1)$. Hence
\[
L_k=\frac{\sigma}{\ell}\sqrt{\frac{\nu}{\nu-1}}
\]

\begin{lemma}[Monotonicity of the Mat\'ern quotient]\label{lem:Bnu_monotone}
For every $\nu>0$, the function $t\mapsto\{\sigma^2-k_\nu(t)\}/t^2$ is nonincreasing on $(0,\infty)$, where $k_\nu$ denotes the Mat\'ern kernel with variance $\sigma^2$.
\end{lemma}

\begin{proof}
The Mat\'ern correlation $\psi_\nu(t):=k_\nu(t)/\sigma^2$ is a valid isotropic correlation function in every dimension $d$, so by Schoenberg's theorem the function $x\mapsto\psi_\nu(\sqrt x)$ is completely monotone on $[0,\infty)$. The Bernstein-Widder theorem therefore gives a probability measure $\pi_\nu$ on $[0,\infty)$ such that
\[
\psi_\nu(t)=\int_0^\infty e^{-t^2u}\,d\pi_\nu(u)
\]
that is, the Mat\'ern kernel is a scale mixture of Gaussian kernels. Consequently,
\[
\frac{\sigma^2-k_\nu(t)}{t^2}
=
\sigma^2\int_0^\infty\frac{1-e^{-t^2u}}{t^2}\,d\pi_\nu(u)
\]
For fixed $u>0$, the map $x\mapsto\{1-e^{-xu}\}/x$ is nonincreasing on $(0,\infty)$: its derivative has the sign of $g(x):=xue^{-xu}-1+e^{-xu}$, and $g(0)=0$ with $g'(x)=-xu^2e^{-xu}\le0$. Substituting $x=t^2$ and integrating over $\pi_\nu$ preserves monotonicity.
\end{proof}

\noindent\textit{Condition \ref{cond:Fourier_Decay}, polynomial Fourier decay.} The Mat\'ern spectral density on $\R^d$ is
\[
\widehat\psi(\bu)
=
\sigma^2\frac{2^d\pi^{d/2}\Gamma(\nu+d/2)(2\nu)^\nu}{\Gamma(\nu)\ell^{2\nu}}
\left(\frac{2\nu}{\ell^2}+\|\bu\|_2^2\right)^{-(\nu+d/2)}
\]
\citep{wendland2004scattered}, which is integrable, strictly positive, and satisfies $\widehat\psi(\bu)^{-1}=O(\|\bu\|_2^{2\nu+d})$; hence condition \ref{cond:Fourier_Decay} holds with $s_k=2\nu+d$, and the regular class requires Sobolev smoothness $s\ge\nu+d/2$.

\noindent\textit{Condition (III), characteristic kernel.} Strict positivity of $\widehat\psi$ everywhere implies that the Mat\'ern kernel is characteristic on all Borel probability measures \citep{sriperumbudur2010hilbert}.

\noindent\textit{Condition (IV), regular model class.} Membership of the return laws in $\mathfrak S_{b,\mathfrak m,r,s}$ with $s\ge\nu+d/2$ is guaranteed by the reward conditions of Lemma~\ref{lem:invariance_identification}; the invariance of the class under $\T^\pi$ is part (i) of that lemma. Finally, the norm equivalence \eqref{eq:sobolev_equivalence} between $\mc H_{\mc Z}$ and $H^{\nu+d/2}(\R^d)$ underlies the pushforward operator bound of Lemma~\ref{lem:pushforward_operator}.

\subsection{Covariance-Operator Concentration under Mixing}\label{app:mixing_concentration}

Theorem~\ref{thm:pointwise_error_bound_intro} assumes high-probability radii $\varepsilon_{\mc{ZX}}(n,\delta)$ and $\varepsilon_{\mc{XX}}(n,\delta)$ for the empirical covariance operators. This subsection derives explicit radii for a single sequence under geometric absolute regularity, giving one sampling design that satisfies the rate conditions in Corollaries~\ref{corr:pointwise_consistency_intro} and~\ref{cor:uniform_strong_consistency_intro}. Recall that a strictly stationary sequence $\{W_i\}_{i\ge1}$ is $\beta$-mixing (absolutely regular) with coefficients $\beta(j)$, and geometrically $\beta$-mixing if $\beta(j)\le\beta_0\vartheta^j$ for some $\beta_0>0$ and $\vartheta\in(0,1)$; geometric $\beta$-mixing implies geometric $\alpha$-mixing.

\begin{lemma}[Concentration of empirical covariance operators]\label{lem:mixing_concentration}
Let $\{(\bX_i,\bZ_i)\}_{i=1}^n$ be strictly stationary and geometrically $\beta$-mixing with constants $(\beta_0,\vartheta)$, and let $ k_{\mc X}$ and $k_{\mc Z}$ be bounded kernels with
\[
\kappa_{\mc X}^2=\sup_\bx k_{\mc X}(\bx,\bx),
\qquad
\kappa_{\mc Z}^2=\sup_\bz k_{\mc Z}(\bz,\bz)
\]
Define the block length
\[
a_n:=\max\left\{1,\left\lceil
\frac{\log(4n\beta_0/\delta)}{\log(1/\vartheta)}
\right\rceil\right\}
\]
Then there is a universal constant $c>0$ such that, for every $\delta\in(0,1)$ and every $n\ge2a_n$, with probability at least $1-\delta$,
\[
\|\widehat{\mbC}_{\mc{ZX}}-\mbC_{\mc{ZX}}\|_{\mathrm{HS}}
\le
c\,\kappa_{\mc X}\kappa_{\mc Z}
\sqrt{\frac{a_n\log(8/\delta)}{n}}
=:
\varepsilon_{\mc{ZX}}(n,\delta)
\]
and the same bound with $\kappa_{\mc Z}$ replaced by $\kappa_{\mc X}$ holds for $\|\widehat{\mbC}_{\mc{XX}}-\mbC_{\mc{XX}}\|_{\mathrm{HS}}$, which dominates the operator norm. In particular $\varepsilon_{\mc{ZX}}(n,\delta)$ and $\varepsilon_{\mc{XX}}(n,\delta)$ are of order $\sqrt{\log(n/\delta)\log(8/\delta)/n}$ up to constants depending only on $(\kappa_{\mc X},\kappa_{\mc Z},\beta_0,\vartheta)$.
\end{lemma}

\begin{proof}
Write $\xi_i:= k_{\mc Z}(\bZ_i,\cdot)\otimes k_{\mc X}(\bX_i,\cdot)-\mbC_{\mc{ZX}}$, a strictly stationary, mean-zero sequence in the separable Hilbert space of Hilbert-Schmidt operators from $\mc H_{\mc X}$ to $\mc H_{\mc Z}$, with $\|\xi_i\|_{\mathrm{HS}}\le2\kappa_{\mc X}\kappa_{\mc Z}=:b$ almost surely, and $\widehat{\mbC}_{\mc{ZX}}-\mbC_{\mc{ZX}}=n^{-1}\sum_{i=1}^n\xi_i$. Partition $\{1,\ldots,n\}$ into $2\mu_n$ consecutive blocks of length $a_n$ (discarding at most $2a_n$ terminal points, which contributes at most $2a_nb/n$ to the average and is absorbed into $c$), and let $O_1,\ldots,O_{\mu_n}$ and $E_1,\ldots,E_{\mu_n}$ denote the odd and even block sums. By Berbee's coupling lemma \citep{berbee1979random}, applied successively as in \citet{yu1994rates}, there exist independent random blocks $\widetilde O_1,\ldots,\widetilde O_{\mu_n}$, each distributed as $O_1$, with
\[
\mathbb P\big(\widetilde O_j\ne O_j\ \text{for some }j\big)\le\mu_n\beta(a_n)\le n\beta_0\vartheta^{a_n}\le\delta/4
\]
by the choice of $a_n$. Each $\widetilde O_j$ is a mean-zero element of a separable Hilbert space with $\|\widetilde O_j\|_{\mathrm{HS}}\le a_nb$, so the Hoeffding-type inequality for sums of independent, bounded, Hilbert-space-valued random elements \citep{Pinelis1992} yields, with probability at least $1-\delta/4$,
\[
\Big\|\sum_{j=1}^{\mu_n}\widetilde O_j\Big\|_{\mathrm{HS}}
\le
c_0\,a_nb\sqrt{\mu_n\log(8/\delta)}
\]
for a universal constant $c_0$. The same argument applies to the even blocks. On the intersection of the four events (two couplings, two deviation bounds), which has probability at least $1-\delta$,
\[
\Big\|\frac1n\sum_{i=1}^n\xi_i\Big\|_{\mathrm{HS}}
\le
\frac{2c_0a_nb\sqrt{\mu_n\log(8/\delta)}}{n}+\frac{2a_nb}{n}
\le
c\,b\sqrt{\frac{a_n\log(8/\delta)}{n}}
\]
using $\mu_n\le n/(2a_n)$. The statement for $\widehat{\mbC}_{\mc{XX}}$ is identical with $\xi_i= k_{\mc X}(\bX_i,\cdot)\otimes k_{\mc X}(\bX_i,\cdot)-\mbC_{\mc{XX}}$, and $\|\cdot\|_{\mathrm{op}}\le\|\cdot\|_{\mathrm{HS}}$.
\end{proof}

\begin{remark}
With $\varepsilon_{\mc{ZX}}(n,\delta)\asymp\varepsilon_{\mc{XX}}(n,\delta)\asymp\sqrt{\log(n/\delta)\log(8/\delta)/n}$, the conditions of Corollary~\ref{corr:pointwise_consistency_intro} hold for any $\lambda_n\to0$ with $\lambda_n^2\sqrt{n}/\log n\to\infty$, and those of Corollary~\ref{cor:uniform_strong_consistency_intro} hold with, for example, $\delta_n=n^{-2}$ and any such $\lambda_n$. Under weaker, merely summable $\alpha$-mixing, \citet{MollenhauerKlusSchutteKoltai2022} establish $\|\widehat{\mbC}-\mbC\|_{\mathrm{HS}}=O_P(n^{-1/2})$ together with deviation inequalities that can be substituted for Lemma~\ref{lem:mixing_concentration}; only the form of the radii changes.
\end{remark}

\subsection{Mean-Embedding Inversion is Ill-Posed}\label{app:emb_ill_posed}
Fix $\bx=(\bs^\top,\ba^\top)^\top$, and let $g:\mc Z\to\mathbb R$ be a measurable test function that is integrable under the conditional return law at $\bx$ but need not belong to $\mc H_{\mc Z}$. Define the expectation-type target functional
\[
\theta_g(\bx):=\E\{g(\bZ)\mid\bX=\bx\}.
\]
Indicator functions are typical examples. Density evaluation at a point is more singular and can instead be approximated by a localized smooth test function, whose bandwidth contributes to the approximation bias. To approximate $\theta_g(\bx)$ from a fitted embedding, first form a regularized RKHS proxy for $g$ on the return grid. Let $h_z\in\mc H_{\mc Z}$ denote the population regularized proxy at regularization level $\lambda>0$ and $\widehat h_z$ its empirical counterpart with coefficient vector $\mathbf c^{(g)}=(\mathbb K_{\mc Z} + m\lambda \mathbb{I})^{-1}\mathbf g$, where $\mathbf g=\{g(\bz_1),\ldots,g(\bz_m)\}^\top$. The plug-in estimate is $\widehat\theta_g(\bx):=\langle\widehat h_z,\widehat\mu_{\bZ\mid\bX=\bx}\rangle_{\mc H_{\mc Z}}$. Adding and subtracting $\langle h_z,\widehat\mu\rangle$ and $\langle h_z,\mu\rangle$ and applying the Cauchy-Schwarz inequality gives
\begin{align*}
\Big|\widehat\theta_g(\bx)-\theta_g(\bx)\Big|
& \le \Big|\bigl\langle \widehat{h}_z-h_z,\widehat{\mu}_{\bZ\mid(\bs,\ba)}\bigr\rangle_{\mathcal{H}_{\mathcal{Z}}}\Big| + \Big|\bigl\langle {h}_z,\widehat{\mu}_{\bZ\mid(\bs,\ba)}-\mu_{\bZ\mid(\bs,\ba)}\bigr\rangle_{\mathcal{H}_{\mathcal{Z}}}\Big|
+
\beta_{\mathrm{sm}}(\lambda)
\\
& \le
\underbrace{\|\widehat{h}_z-h_z\|_{\mathcal{H}_{\mathcal{Z}}}\ \|\widehat{\mu}_{\bZ\mid(\bs,\ba)}\|_{\mathcal{H}_{\mathcal{Z}}}}_{\tx{proxy estimation error}}\\
&\qquad+
\underbrace{\|{h}_z\|_{\mathcal{H}_{\mathcal{Z}}}}_{\tx{stability factor}}\ \underbrace{\|\widehat{\mu}_{\bZ\mid(\bs,\ba)}-\mu_{\bZ\mid(\bs,\ba)}\|_{\mathcal{H}_{\mathcal{Z}}}}_{\tx{embedding error}}
+
\underbrace{\beta_{\mathrm{sm}}(\lambda)}_{\tx{smoothing bias}}
\end{align*}
where $\beta_{\mathrm{sm}}(\lambda):=\big|\langle h_z,\mu_{\bZ\mid\bX=\bx}\rangle_{\mc H_{\mc Z}}-\theta_g(\bx)\big|$ is the deterministic bias from replacing $g$ with its regularized proxy. A small embedding error can still produce a substantial error in the recovered functional when the stability factor $\|h_z\|_{\mc H_{\mc Z}}$ is large, reflecting the ill-posed inverse problem in embedding inversion \citep{invregularization}.

\noindent The stability factor has the following spectral expression on the return grid:
\[
\|\widehat{h}_z \|^2_{\mathcal{H}_{\mathcal{Z}}}
= \mathbf{c}^{(g)\top} \mathbb K_{\mc Z} \ \mathbf{c}^{(g)}
= \mathbf{g}^\top (\mathbb K_{\mc Z} + m\lambda \mathbb{I})^{-1} \mathbb K_{\mc Z} (\mathbb K_{\mc Z} + m\lambda \mathbb{I})^{-1} \mathbf{g}
\]
Let $\mathbb K_{\mc Z} = \mathbb{U}  \mathbb{\Lambda}  \mathbb{U}^\top$ with $ \mathbb{\Lambda} = \tx{diag}(\delta_1, \dots, \delta_m)$, so that $(\mathbb K_{\mc Z} + m\lambda \mathbb{I})^{-1}=\mathbb{U} \  \tx{diag}\big( \frac{1}{\delta_i+m\lambda}\big)\ \mathbb{U}^\top$. Then
\[
\|\widehat{h}_z \|^2_{\mathcal{H}_{\mathcal{Z}}}
= \sum_{i=1}^m\frac{\delta_i}{(\delta_i+m\lambda)^2}\ \widetilde{g}_i^{\,2}
\qquad\tx{where}\quad\widetilde{g}_i= (\mathbb{U}^\top\mathbf g)_i
\]

\noindent The spectral identity above shows that the finite-grid proxy is sensitive to small positive eigenvalues. If $\widetilde g_i\ne0$, the $i$th contribution approaches $\widetilde g_i^{\,2}/\delta_i$ as $\lambda\downarrow0$. Hence an indicator, or an increasingly localized smooth approximation to density evaluation, can require a proxy with large RKHS norm when its grid evaluations project appreciably onto small-eigenvalue directions.

\noindent Keeping $\lambda>0$ suppresses these directions but increases the approximation bias $\beta_{\mathrm{sm}}(\lambda)$; the choice of $\lambda$ trades these two effects off against one another.

\subsection{Extensions of the Error Bound for the Regularized Conditional Mean Embedding Estimator}\label{app:tikhonov_extensions}
This subsection derives pointwise consistency, a uniform error bound, and uniform strong consistency for the regularized conditional mean embedding estimator. Its scope is this estimator rather than the full finite-dictionary estimator in Algorithm~\ref{alg:rk_drl}.

\begin{definition}[Strong consistency in $L^2(\mathbb P_{\bX};\mc H_{\mc Z})$]
A sequence $\{\widehat{\mu}^{(n)}_{\bZ\mid\bx}\}_{n\ge1}$ is strongly consistent in $L^2(\mathbb P_{\bX};\mc H_{\mc Z})$ if
\[
\int_{\mc X}\|\widehat{\mu}^{(n)}_{\bZ\mid\bx}-\mu_{\bZ\mid\bx}\|^2_{\mc H_{\mc Z}}\,d\mathbb P_{\bX}(\bx)
\overset{a.s.}{\longrightarrow}0
\]
\end{definition}

\begin{corollary}[Pointwise Consistency]\label{corr:pointwise_consistency_intro}
Under the conditions of Theorem~\ref{thm:pointwise_error_bound_intro}, if $\delta_n\to0$, $\lambda_n\to0$, $\varepsilon_{\mc{ZX}}(n,\delta_n)/\lambda_n\to0$, and $\varepsilon_{\mc{XX}}(n,\delta_n)/\lambda_n^2\to0$, then for each fixed $\bx\in\mc X$,
\[
\|\widehat{\mu}_{\bZ\mid\bx}-\mu_{\bZ\mid\bx}\|_{\mc H_{\mc Z}}
\overset{p}{\longrightarrow}0
\]
\end{corollary}

\begin{theorem}[Uniform Error Bound]\label{thm:uniform_error_bound_intro}\label{thm:unif_error_bound}
Assume the conditions of Theorem~\ref{thm:pointwise_error_bound_intro} hold for every $\bx\in\mc X$, with the source condition holding uniformly:
\[
\sup_{\bx\in\mc X}\|g_{\bx}\|_{\mc H_{\mc X}}<\infty
\]
Define
\[
c_\star:=\|\mbC_{\mc{ZX}}\mbC_{\mc{XX}}^{-1/2}\|_{\mathrm{op}}
\sup_{\bx\in\mc X}\|g_{\bx}\|_{\mc H_{\mc X}}
\]
Then, for any $\delta\in(0,1)$, on the covariance-concentration event of Theorem~\ref{thm:pointwise_error_bound_intro}, which has probability at least $1-\delta$ and does not depend on the evaluation point,
\[
\sup_{\bx\in\mc X}\|\widehat\mu_{\bZ\mid\bx}-\mu_{\bZ\mid\bx}\|_{\mc H_{\mc Z}}
\le
\frac{\kappa_{\mc X}}{\lambda}\varepsilon_{\mc{ZX}}(n,\delta)
+\frac{\kappa_{\mc X}^2\kappa_{\mc Z}}{\lambda^2}\varepsilon_{\mc{XX}}(n,\delta)
+c_\star\lambda^\tau
\]
No covering-number, total-boundedness, or Lipschitz condition is required because the covariance-operator event holds simultaneously for every evaluation point.
\end{theorem}

\begin{corollary}[Strong Consistency]\label{cor:uniform_strong_consistency_intro}\label{cor:unif_strong_consistency}
Assume the conditions of Theorem~\ref{thm:uniform_error_bound_intro} for every $n$, with $\lambda_n>0$ and $\delta_n\in(0,1)$ satisfying $\sum_n\delta_n<\infty$, $\lambda_n\to0$,
\[
\frac{\varepsilon_{\mc{ZX}}(n,\delta_n)}{\lambda_n}\to0,
\qquad
\frac{\varepsilon_{\mc{XX}}(n,\delta_n)}{\lambda_n^2}\to0
\]
Then $\sup_{\bx\in\mc X}\|\widehat\mu_{\bZ\mid\bx}-\mu_{\bZ\mid\bx}\|_{\mc H_{\mc Z}}\to0$ almost surely. Consequently, if $\mathbb P_{\bX}$ is supported on $\mc X$, then $\widehat{\mu}^{(n)}_{\bZ\mid\bx}$ is strongly consistent in $L^2(\mathbb P_{\bX};\mc H_{\mc Z})$.
\end{corollary}
\section{Proofs}
\subsection{Density of the Bellman-Updated Return}\label{proof:DBO-pdf}
\begin{theorem}[Density of the Bellman-updated return]\label{thm:DBO-pdf}
Fix $\bx=(\bs^\top,\ba^\top)^\top\in\mc X$ and let $\gamma\in(0,1)$. Let $\mathcal{M}^\pi(d\br,d\bx'\mid\bx)$ denote the joint target-policy one-step law of $(\bR,\bX')$ conditional on $\bX=\bx$, and let $\mc{P}_X^\pi(d\bx'\mid\bx)$ denote its $\bX'$-marginal. Assume that, for $\mc{P}_X^\pi(\cdot\mid\bx)$-almost every $\bx'$, the candidate continuation law $\Q_\bZ(\cdot\mid\bx')$ admits a Lebesgue density $f_\bZ(\cdot\mid\bx')$ on $\R^d$, and that $(\bz,\bx')\mapsto f_\bZ(\bz\mid\bx')$ admits a jointly measurable version. Assume also that, conditional on $(\bX,\bX')=(\bx,\bx')$, the continuation draw is independent of the one-step reward $\bR$. Then the conditional law of $(\T^\pi\bZ)(\bx)=\bR+\gamma\bZ(\bX')$ admits the density
\begin{equation}\label{eq:DBO-density-x}
f_{\T^\pi\bZ}(\by\mid\bx)
=
\int_{\mc X}
\int_{\mc R}
\gamma^{-d}
f_\bZ\left(
\frac{\by-\br}{\gamma}
\mid
\bx'
\right)
\mathcal{M}^\pi(d\br,d\bx'\mid\bx)
\end{equation}
Equivalently, suppose the joint one-step law admits the disintegration $\mathcal{M}^\pi(d\br,d\bx'\mid\bx)=F_{\bR}(d\br\mid\bx,\bs')\,\mc{P}_X^\pi(d\bx'\mid\bx)$, where the conditional reward law depends on $\bx'=(\bs'^\top,\ba'^\top)^\top$ only through $\bs'$ and does not depend on the target policy. Suppose the successor action is drawn independently of the reward given the successor state and $\mc{P}_X^\pi(d\bx'\mid\bx)$ has density $p^\pi(\bs',\ba'\mid\bs,\ba)=p(\bs'\mid\bs,\ba)\pi(\ba'\mid\bs')$. Then
\begin{equation}\label{eq:DBO-density-sa}
f_{\T^\pi\bZ}(\by\mid\bs,\ba)
=
\int_{\mc S}
\int_{\mc A}
\int_{\mc R}
\gamma^{-d}
f_\bZ\left(
\frac{\by-\br}{\gamma}
\mid
\bs',\ba'
\right)
F_{\bR}(d\br\mid\bs,\ba,\bs')\,
p^\pi(\bs',\ba'\mid\bs,\ba)
\,d\ba'\,d\bs'
\end{equation}
If the joint one-step law factorizes as $\mathcal{M}^\pi(d\br,d\bx'\mid\bx)=F_{\bR}(d\br\mid\bx)\,\mc{P}_X^\pi(d\bx'\mid\bx)$, equation~\eqref{eq:DBO-density-x} reduces to the corresponding product-form integral.
\end{theorem}
\begin{proof}
Fix $\bx\in\mc X$ and let $\mathcal{M}^\pi(d\br,d\bx'\mid\bx)$ be the joint target-policy law of the immediate reward and successor input. Conditional on $(\bR,\bX')=(\br,\bx')$, the Bellman target has the pushforward law $(\theta_{\br,\gamma})_{\#}\Q_\bZ(\cdot\mid\bx')$ defined in \eqref{eq:pushforward_map}. We show that its mixture over $\mathcal{M}^\pi$ admits the stated density. Let $A\subseteq\R^d$ be an arbitrary Borel set. Conditioning on the paired one-step variables $(\bR,\bX')$ and using the assumed conditional independence of the continuation draw and one-step reward given $(\bX,\bX')$ gives
\begin{align*}
\mathbb P\{\bR+\gamma\bZ(\bX')\in A\mid\bX=\bx\}
&=
\int_{\mc R\times\mc X}
\left[
\int_{\R^d}
\one\{\br+\gamma\bz\in A\}
f_\bZ(\bz\mid\bx')\,d\bz
\right]
\mathcal{M}^\pi(d\br,d\bx'\mid\bx)
\end{align*}
For fixed $(\br,\bx')$, the affine change of variables $\by=\theta_{\br,\gamma}(\bz)=\br+\gamma\bz$, with Jacobian $\gamma^{-d}$ since $\gamma>0$, gives
\[
\int_{\R^d}
\one\{\br+\gamma\bz\in A\}
f_\bZ(\bz\mid\bx')\,d\bz
=
\int_{A}
\gamma^{-d}
f_\bZ\left(\frac{\by-\br}{\gamma}\mid\bx'\right)d\by
\]
The integrand $(\by,\br,\bx')\mapsto\gamma^{-d}f_\bZ\{(\by-\br)/\gamma\mid\bx'\}$ is jointly measurable by the assumed jointly measurable version of the conditional density, and it is nonnegative. Tonelli's theorem therefore permits the interchange of the $\by$-integral and the $\mathcal{M}^\pi$-integral:
\[
\mathbb P\{\bR+\gamma\bZ(\bX')\in A\mid\bX=\bx\}
=
\int_{A}
\left[
\int_{\mc R\times\mc X}
\gamma^{-d}
f_{\bZ}\left(\frac{\by-\br}{\gamma}\mid\bx'\right)
\mathcal{M}^\pi(d\br,d\bx'\mid\bx)
\right]d\by
\]
Since $A$ is arbitrary, the bracketed function is a Lebesgue density of the conditional law of $(\T^\pi\bZ)(\bx)$, which proves \eqref{eq:DBO-density-x}. The argument requires neither smoothness of $f_\bZ$ nor differentiation under the integral sign. In state-action notation, disintegrating the joint law as
\[
\mathcal{M}^\pi(d\br,d\bs',d\ba'\mid\bs,\ba)
=
F_{\bR}(d\br\mid\bs,\ba,\bs')\,p(\bs'\mid\bs,\ba)\pi(\ba'\mid\bs')\,d\bs'\,d\ba'
\]
gives \eqref{eq:DBO-density-sa}. If $\mathcal{M}^\pi(d\br,d\bx'\mid\bx)=F_{\bR}(d\br\mid\bx)\,\mc{P}_X^\pi(d\bx'\mid\bx)$, the display reduces to the product form stated after the theorem.
\end{proof}
\subsection{Proof of Proposition \ref{prop:DBO_empirical} {\small (Finite-Dictionary Bellman Target Embedding)}} \label{proof:DBO_empirical}
\begin{proof}
Throughout the proof the candidate matrix $\mathbb B\in\R^{L\times m}$ is fixed, all identities between $\mc H_{\mc Z}$-valued quantities are equalities in $\mc H_{\mc Z}$, and all expectations of $\mc H_{\mc Z}$-valued maps are Bochner integrals. Since both kernels are bounded, set
\[
\kappa_{\mc Z}:=\sup_{\by\in\mc Z}k_{\mc Z}(\by,\by)^{1/2}<\infty,
\qquad
\kappa_{\mc X}:=\sup_{\bu\in\mc X} k_{\mc X}(\bu,\bu)^{1/2}<\infty
\]
Recall from \eqref{eq:mu_Z}-\eqref{eq:hat_w} that the finite-dictionary candidate embedding at an arbitrary input $\bu\in\mc X$ is
\[
\widehat\mu_{\bZ\mid\bu}(\cdot\ ;\mathbb B)
=\sum_{i=1}^{m}\omega_i(\bu;\mathbb B)\,k_{\mc Z}(\bz_i,\cdot),
\qquad
\omega_i(\bu;\mathbb B)=\sum_{\ell=1}^{L}b_{\ell i}\, k_{\mc X}(\underline\bx_\ell,\bu)
\]
the RKHS embedding of the finite signed measure $\Q_{\bu}:=\sum_{i=1}^{m}\omega_i(\bu;\mathbb B)\,\delta_{\bz_i}$ supported on the return grid $\mathbf Z_m$. The proof has four steps: derive the adjoint image of one kernel section, apply this identity to the finite dictionary, verify Bochner integrability, and invoke Assumption~\ref{ass:reward_continuation_separation} to obtain the separated target in \eqref{eq:mu_TZ}.

\medskip
\noindent\textit{Step 1: adjoint image of a kernel section.}
Fix a reward value $\br\in\mc R$ and recall from \eqref{eq:pushforward_map} the affine Bellman map $\theta_{\br,\gamma}(\bz)=\br+\gamma\bz$ and the composition operator $\mc{C}_{\br,\gamma}g:=g\circ\theta_{\br,\gamma}$, which is bounded on $\mc H_{\mc Z}$ with $\|\mc{C}_{\br,\gamma}\|_{\mathrm{op}}\le c_\gamma$ uniformly in $\br$ by Lemma~\ref{lem:pushforward_operator}(i). Its adjoint maps kernel sections to reward-shifted kernel sections:
\begin{equation}\label{eq:adjoint_shift}
\mc{C}_{\br,\gamma}^{*}\,k_{\mc Z}(\bz,\cdot)
=
k_{\mc Z}(\br+\gamma\bz,\cdot)
\qquad\text{for every }\bz\in\mc Z
\end{equation}
Indeed, for arbitrary $g\in\mc H_{\mc Z}$, the definition of the adjoint and the reproducing property give
\[
\big\langle g,\ \mc{C}_{\br,\gamma}^{*}k_{\mc Z}(\bz,\cdot)\big\rangle_{\mc H_{\mc Z}}
=
\big\langle \mc{C}_{\br,\gamma}g,\ k_{\mc Z}(\bz,\cdot)\big\rangle_{\mc H_{\mc Z}}
=
(\mc{C}_{\br,\gamma}g)(\bz)
=
g(\br+\gamma\bz)
=
\big\langle g,\ k_{\mc Z}(\br+\gamma\bz,\cdot)\big\rangle_{\mc H_{\mc Z}}
\]
Since $g$ is arbitrary and both candidate images lie in $\mc H_{\mc Z}$, \eqref{eq:adjoint_shift} follows. Identity \eqref{eq:adjoint_shift} is the exact atomic counterpart of the change of variables used in Theorem~\ref{thm:DBO-pdf}: when a law $\Q$ on $\mc Z$ has Lebesgue density $f$, Lemma~\ref{lem:pushforward_operator}(ii) and the substitution $\by=\br+\gamma\bz$, $d\by=\gamma^{d}\,d\bz$, give
\[
\mu_{(\theta_{\br,\gamma})_{\#}\Q}
=
\int_{\R^d} k_{\mc Z}(\by,\cdot)\,\gamma^{-d}f\left(\frac{\by-\br}{\gamma}\right)d\by
=
\int_{\R^d} k_{\mc Z}(\br+\gamma\bz,\cdot)\,f(\bz)\,d\bz
\]
for the atomic measure $\Q_{\bu}$, the same transport acts on each atom $\bz_i$ separately and without any approximation.

\medskip
\noindent\textit{Step 2: pushforward of the finite dictionary.}
Because $\mc{C}_{\br,\gamma}^{*}$ is a bounded linear operator, applying it to the finite expansion of $\widehat\mu_{\bZ\mid\bu}(\cdot\ ;\mathbb B)$ and using \eqref{eq:adjoint_shift} termwise gives, for every $\bu\in\mc X$ and every $\br\in\mc R$,
\begin{equation}\label{eq:pushforward_dictionary}
\mc{C}_{\br,\gamma}^{*}\,\widehat\mu_{\bZ\mid\bu}(\cdot\ ;\mathbb B)
=
\sum_{i=1}^{m}\omega_i(\bu;\mathbb B)\,\mc{C}_{\br,\gamma}^{*}k_{\mc Z}(\bz_i,\cdot)
=
\sum_{i=1}^{m}\omega_i(\bu;\mathbb B)\,k_{\mc Z}(\br+\gamma\bz_i,\cdot)
\end{equation}
The induced Bellman embedding operator is linear on embeddings of finite signed measures: the pushforward of $\Q_{\bu}$ under $\theta_{\br,\gamma}$ is $\sum_{i=1}^m\omega_i(\bu;\mathbb B)\delta_{\br+\gamma\bz_i}$, and \eqref{eq:pushforward_dictionary} is its embedding. Substituting the candidate embedding into the Bellman target embedding in \eqref{eq:pushforward_bellman_eb} and integrating under the joint one-step law $\mathcal{M}^\pi(d\br,d\bx'\mid\bx)$ gives
\begin{equation}\label{eq:pair_embedding_derived}
\begin{aligned}
\widehat{\mu}_{\T^\pi\bZ\mid\bx}(\cdot\ ;\mathbb B)
&=
\int_{\mc R\times\mc X}
\mc{C}_{\br,\gamma}^{*}\,\widehat\mu_{\bZ\mid\bx'}(\cdot\ ;\mathbb B)\,
\mathcal{M}^\pi(d\br,d\bx'\mid\bx)
\\
&=
\int_{\mc R\times\mc X}
\sum_{i=1}^{m}
\omega_i(\bx';\mathbb B)\,
k_{\mc Z}(\br+\gamma\bz_i,\cdot)\,
\mathcal{M}^\pi(d\br,d\bx'\mid\bx)
\end{aligned}
\end{equation}
which is \eqref{eq:projected_pair_embedding}. This identity uses only the linearity in \eqref{eq:pushforward_dictionary} and retains the joint one-step law of $(\bR,\bX')$; it does not use independence between the reward and successor input.

\medskip
\noindent\textit{Step 3: Bochner integrability.}
We verify that the Bochner integral in \eqref{eq:pair_embedding_derived} is well defined and can be taken termwise. First, by the reproducing property and boundedness of $k_{\mc Z}$,
\[
\big\|k_{\mc Z}(\br+\gamma\bz_i,\cdot)\big\|_{\mc H_{\mc Z}}
=
k_{\mc Z}(\br+\gamma\bz_i,\br+\gamma\bz_i)^{1/2}
\le
\kappa_{\mc Z}
\qquad\text{for every }\br\in\mc R
\]
Second, by \eqref{eq:hat_w} and the Cauchy-Schwarz inequality $| k_{\mc X}(\bu,\bu')|\le k_{\mc X}(\bu,\bu)^{1/2} k_{\mc X}(\bu',\bu')^{1/2}\le\kappa_{\mc X}^2$ for all $\bu,\bu'\in\mc X$,
\[
\big|\omega_i(\bx';\mathbb B)\big|
\le
\sum_{\ell=1}^{L}|b_{\ell i}|\,\big| k_{\mc X}(\underline\bx_\ell,\bx')\big|
\le
\kappa_{\mc X}^{2}\sum_{\ell=1}^{L}|b_{\ell i}|
=:\bar\omega_i(\mathbb B)<\infty
\qquad\text{for every }\bx'\in\mc X
\]
Hence each summand in \eqref{eq:pair_embedding_derived} is an $\mc H_{\mc Z}$-valued map of $(\br,\bx')$ with $\|\omega_i(\bx';\mathbb B)k_{\mc Z}(\br+\gamma\bz_i,\cdot)\|_{\mc H_{\mc Z}}\le\bar\omega_i(\mathbb B)\,\kappa_{\mc Z}$ everywhere. Each summand is also Bochner measurable: $\bx'\mapsto\omega_i(\bx';\mathbb B)$ is measurable because $ k_{\mc X}$ is a measurable kernel, the feature map $\by\mapsto k_{\mc Z}(\by,\cdot)$ is continuous since
\[
\big\|k_{\mc Z}(\by,\cdot)-k_{\mc Z}(\by',\cdot)\big\|_{\mc H_{\mc Z}}^{2}
=
k_{\mc Z}(\by,\by)-2k_{\mc Z}(\by,\by')+k_{\mc Z}(\by',\by')
\longrightarrow0
\qquad\text{as }\by'\to\by
\]
for the (continuous) Mat\'ern kernel, and $\mc H_{\mc Z}$ is separable, so strong measurability follows from the Pettis measurability theorem. Consequently the Bochner integral of each summand under $\mathcal{M}^\pi(\cdot\mid\bx)$ exists, and by linearity of the Bochner integral over the finite sum,
\begin{equation}\label{eq:sum_exchange}
\widehat{\mu}_{\T^\pi\bZ\mid\bx}(\cdot\ ;\mathbb B)
=
\sum_{i=1}^{m}
\int_{\mc R\times\mc X}
\omega_i(\bx';\mathbb B)\,
k_{\mc Z}(\br+\gamma\bz_i,\cdot)\
\mathcal{M}^\pi(d\br,d\bx'\mid\bx)
\end{equation}

\medskip
\noindent\textit{Step 4: conditional moment factorization.}
Fix $i\in\{1,\ldots,m\}$. Assumption~\ref{ass:reward_continuation_separation} states that the conditional covariance in $\mc H_{\mc Z}$ between the scalar continuation coefficient $\omega_i(\bX';\mathbb B)$ and the reward-shift feature $k_{\mc Z}(\bR+\gamma\bz_i,\cdot)$ vanishes given $\bX=\bx$, so that
\[
\int_{\mc R\times\mc X}
\omega_i(\bx';\mathbb B)\,
k_{\mc Z}(\br+\gamma\bz_i,\cdot)\
\mathcal{M}^\pi(d\br,d\bx'\mid\bx)
=
\underbrace{\int_{\mc X}\omega_i(\bx';\mathbb B)\,\mc{P}_X^\pi(d\bx'\mid\bx)}_{=\ \omega_i^\pi(\bx;\mathbb B)}
\ \underbrace{\int_{\mc R}k_{\mc Z}(\br+\gamma\bz_i,\cdot)\,F_{\bR}(d\br\mid\bx)}_{=\ \ell_i(\bx,\cdot)}
\]
Both factors are well defined: the scalar factor is finite since $|\omega_i(\bx';\mathbb B)|\le\bar\omega_i(\mathbb B)$, and the $\mc H_{\mc Z}$-valued factor is a Bochner integral of a map bounded in norm by $\kappa_{\mc Z}$, so $\|\ell_i(\bx,\cdot)\|_{\mc H_{\mc Z}}\le\kappa_{\mc Z}$. The quantities $\omega_i^\pi(\bx;\mathbb B)$ and $\ell_i(\bx,\cdot)$ are exactly those defined in the statement of the proposition. Substituting the display into \eqref{eq:sum_exchange} gives
\[
\widehat{\mu}_{\T^\pi\bZ\mid\bx}(\cdot\ ;\mathbb B)
=
\sum_{i=1}^{m}\omega_i^\pi(\bx;\mathbb B)\,\ell_i(\bx,\cdot)
\]
which is \eqref{eq:mu_TZ}.

\medskip
\noindent When $\mc{P}_X^\pi(d\bx'\mid\bx)$ has density $p^\pi(\bx'\mid\bx)$, the continuation coefficient has the dominated representation $\omega_i^\pi(\bx;\mathbb B)=\int_{\mc X}\omega_i(\bx';\mathbb B)\,p^\pi(\bx'\mid\bx)\,d\bx'$, which is used for off-policy estimation in Section~\ref{sec:ope_weights}. Equation~\eqref{eq:mu_TZ} is the finite-dictionary analogue of integrating the Bellman density \eqref{eq:DBO-density-x} against a kernel section, with each return-grid atom transported from $\bz_i$ to $\br+\gamma\bz_i$ before embedding.
\end{proof}

\subsection[Proof of the fixed-point theorem]{Proof of Theorem \ref{thm:fixed_point} {\small(Fixed Point of the Distributional Bellman Operator)}}\label{proof:fixed_point}
\noindent\textit{Organization of the proof.}
We first establish the H\"older bound (a) for the induced embedding operator. We then prove existence (b) by verifying the hypotheses of Lemma~\ref{lem:AGP}: the set $K$ is weakly compact and convex, and $\mathfrak T^\pi$ is weakly sequentially continuous by Lemma~\ref{lemma:weak_cts_embed_operator}. We prove uniqueness (c) using the $\gamma$-contraction property of the distributional Bellman operator in the maximal conditional Wasserstein metric, derive the residual bound (d), and conclude with the identification statement via Lemma~\ref{lem:invariance_identification}.

\begin{proof}
Let $\mathbb P$ and $\mathbb Q$ denote two conditional return-law maps represented by elements of $K$. Thus $\mathbb P(\cdot\mid\bs,\ba)$ is the law of $\bZ\mid\bs,\ba$, and $\mathbb Q(\cdot\mid\bs,\ba)$ is the law of $\widetilde{\bZ}\mid\bs,\ba$. By Assumption~\ref{ass:regular_bellman_embedding_class}, these conditional laws and their Bellman updates remain in the regular class $\mathfrak S_{b,\mathfrak m,r,s}$ for every conditioning input. We write $\T^\pi\mathbb P$ and $\T^\pi\mathbb Q$ for the corresponding conditional laws after applying the distributional Bellman operator. By the definition of the RKHS discrepancy,
\begin{align*}
\Bgamma_k\{(\T^\pi\mathbb P)(\cdot\mid\bs,\ba),(\T^\pi\mathbb Q)(\cdot\mid\bs,\ba)\}
&=
\bigl\|\mu_{\T^\pi\mathbb P}(\bs,\ba)-\mu_{\T^\pi\mathbb Q}(\bs,\ba)\bigr\|_{\mc H_{\mc Z}}\\
&=
\left\|
\int k_{\mc Z}(\cdot,\bz)\,d(\T^\pi\mathbb P)(\bz\mid\bs,\ba)
-
\int k_{\mc Z}(\cdot,\bz)\,d(\T^\pi\mathbb Q)(\bz\mid\bs,\ba)
\right\|_{\mc H_{\mc Z}}
\end{align*}

\noindent We next write the mean embedding of the Bellman update using the joint one-step law. Let $\mathcal{M}^\pi(d\br,d\bs',d\ba'\mid\bs,\ba)$ denote the target-policy law of $(\bR,\bS',\bA')$ given $(\bS,\bA)=(\bs,\ba)$. Then
\begin{align*}
\mu_{\T^\pi\mathbb P}(\bs,\ba)
=
(\mathfrak T^\pi\mu_{\mathbb P})(\bs,\ba)
&=
\int_{\mc R\times\mc S\times\mc A}\int_{\mc Z}
k_{\mc Z}(\cdot,\br+\gamma\widetilde{\bz})\,
d\mathbb P(\widetilde{\bz}\mid\bs',\ba')
\mathcal{M}^\pi(d\br,d\bs',d\ba'\mid\bs,\ba)\end{align*}
The same expression holds with $\mathbb Q$ in place of $\mathbb P$.

\noindent For each next input $(\bs',\ba')$, let $\Upsilon_{\bs',\ba'}$ be an optimal coupling between $\mathbb P(\cdot\mid\bs',\ba')$ and $\mathbb Q(\cdot\mid\bs',\ba')$; a jointly measurable selection $(\bs',\ba')\mapsto\Upsilon_{\bs',\ba'}$ exists by the measurable-selection result for optimal transport plans \citep[Corollary~5.22]{villani2009optimal}, so the integrals below are well defined. Then, for each fixed current input $(\bs,\ba)$,
\begin{align}
\Bgamma_k\{(\T^\pi\mathbb P)(\cdot\mid\bs,\ba),(\T^\pi\mathbb Q)(\cdot\mid\bs,\ba)\} &=
\bigl\|
(\mathfrak T^\pi\mu_{\mathbb P})(\bs,\ba)
-
(\mathfrak T^\pi\mu_{\mathbb Q})(\bs,\ba)
\bigr\|_{\mc H_{\mc Z}}
\nonumber\\
& =
\Bigg\|
\E_{\mathcal{M}^\pi(\cdot\mid\bs,\ba)}
\int
\{k_{\mc Z}(\cdot,\bR+\gamma\bz)-k_{\mc Z}(\cdot,\bR+\gamma\widetilde{\bz})\}
d\Upsilon_{\bS',\bA'}(\bz,\widetilde{\bz})
\Bigg\|_{\mc H_{\mc Z}}
\nonumber\\
&\le
\E_{\mathcal{M}^\pi(\cdot\mid\bs,\ba)}
\int
\bigl\|
k_{\mc Z}(\cdot,\bR+\gamma\bz)-k_{\mc Z}(\cdot,\bR+\gamma\widetilde{\bz})
\bigr\|_{\mc H_{\mc Z}}
d\Upsilon_{\bS',\bA'}(\bz,\widetilde{\bz})
\nonumber\\
&\le
\gamma L_k
\E_{\mathcal{M}^\pi(\cdot\mid\bs,\ba)}
\int
\|\bz-\widetilde{\bz}\|
d\Upsilon_{\bS',\bA'}(\bz,\widetilde{\bz})
\nonumber\\
& =
\gamma L_k
\E_{\mathcal{M}^\pi(\cdot\mid\bs,\ba)}
W_1\{\mathbb P(\cdot\mid\bS',\bA'),\mathbb Q(\cdot\mid\bS',\bA')\}
\nonumber\\
& \le
\gamma L_k
\sup_{\bs,\ba}
W_1\{\mathbb P(\cdot\mid\bs,\ba),\mathbb Q(\cdot\mid\bs,\ba)\}
\nonumber\\
& \le
\gamma L_k C_*
\left[
\sup_{\bs,\ba}
\Bgamma_k\{\mathbb P(\cdot\mid\bs,\ba),\mathbb Q(\cdot\mid\bs,\ba)\}
\right]^\rho
\label{eq:ineq_Cstar_Lk}
\end{align}
Taking the supremum over the current input on the left-hand side gives the operator-level H\"older bound
\begin{equation}\label{eq:operator_holder_sup}
D_k(\T^\pi\mathbb P,\T^\pi\mathbb Q)
\le
\gamma L_k C_*
D_k(\mathbb P,\mathbb Q)^\rho \end{equation}
Here the power can be placed outside the supremum because $0<\rho<1$ and $\sup_x a_x^\rho=(\sup_x a_x)^\rho$ for nonnegative $a_x$.
The first inequality follows from Jensen's inequality and the triangle inequality in the RKHS. The second uses the Lipschitz continuity of the Mat\'ern feature map.\footnote{The Mat\'ern kernel is $k_{\nu}(t) = \sigma^2
\frac{2^{1-\nu}}{\Gamma(\nu)}
\left(\sqrt{2\nu}\frac{t}{\ell}\right)^\nu
\mathcal K_{\nu}\left(\sqrt{2\nu}\frac{t}{\ell}\right)$ where $t=\|\bz-\widetilde{\bz}\|$ is the distance in the return space, $\nu>0$ controls smoothness, $\ell>0$ is the length scale, $\sigma^2>0$ is the variance parameter, and $\mathcal K_\nu$ is the modified Bessel function of the second kind. Since $\bigl\|k_\nu(\cdot,\bz)-k_\nu(\cdot,\widetilde{\bz})\bigr\|_{\mc H_{\mc Z}}^2 = k_\nu(\bz,\bz)+k_\nu(\widetilde{\bz},\widetilde{\bz})-2k_\nu(\bz,\widetilde{\bz})
= 2\{\sigma^2-k_\nu(\|\bz-\widetilde{\bz}\|)\}$ we have
\begin{align*}
L_k:=
\sup_{\bz\ne\widetilde{\bz}}
\frac{\bigl\|k_\nu(\cdot,\bz)-k_\nu(\cdot,\widetilde{\bz})\bigr\|_{\mc H_{\mc Z}}}
{\|\bz-\widetilde{\bz}\|}
=
\sup_{t>0}
\frac{\sqrt{2\{\sigma^2-k_\nu(t)\}}}{t}=
\frac{2\sigma\sqrt{\nu}}{\ell}
\sqrt{
\sup_{u>0}
\frac{
1-\frac{2^{1-\nu}}{\Gamma(\nu)}u^\nu\mathcal K_\nu(u)
}{u^2}
}
\end{align*}
where $u:=\sqrt{2\nu}t/\ell$. Let $c_\nu:=2^{1-\nu}/\Gamma(\nu)$ and $B_\nu(u):=\{1-c_\nu u^\nu\mathcal K_\nu(u)\}/u^2$. For $u>0$, $B_\nu(u)$ is nonincreasing by Lemma~\ref{lem:Bnu_monotone} in Supplementary Material~\ref{app:matern_verification}, and therefore
\begin{align*}
\sup_{u>0}B_\nu(u)
&=
\lim_{u\to0}
\frac{1-c_\nu u^\nu\mathcal K_\nu(u)}{u^2}\overset{\text{L'H\^opital}}{=}
\lim_{u\to0}
\frac{-c_\nu\{\nu u^{\nu-1}\mathcal K_\nu(u)+u^\nu(d/du)\mathcal K_\nu(u)\}}{2u}
\\
&=
\lim_{u\to0}
\frac{c_\nu u^{\nu-1}\mathcal K_{\nu-1}(u)}{2}
=
\frac{c_\nu}{2c_{\nu-1}}
=
\frac{1}{4(\nu-1)}
\end{align*}
The third equality uses the derivative identity
\[
\frac{d}{du}\mathcal K_{\nu}(u)
=
-\frac{\nu}{u}\mathcal K_{\nu}(u)-\mathcal K_{\nu-1}(u)
\]
\citep[\S9.6.26]{abramowitz1948handbook}. The fourth uses the small-argument asymptotic $\mathcal K_{\nu-1}(u)\sim\tfrac12\Gamma(\nu-1)(u/2)^{-(\nu-1)}$ as $u\to0$, valid since $\nu>1$ \citep[\S9.6.9]{abramowitz1948handbook}, and the final equality uses $\Gamma(\nu)=(\nu-1)\Gamma(\nu-1)$. Thus $L_k=\sigma\ell^{-1}\sqrt{\nu/(\nu-1)}$. The same calculation, together with the proof of the monotonicity lemma, is recorded in Supplementary Material~\ref{app:matern_verification}.}
 Specifically, for $\nu>1$,
\[
\|k_{\mc Z}(\cdot,\bz)-k_{\mc Z}(\cdot,\widetilde{\bz})\|_{\mc H_{\mc Z}}
\le
L_k\|\bz-\widetilde{\bz}\|
\qquad
L_k=\frac{\sigma}{\ell}\sqrt{\frac{\nu}{\nu-1}}
\]
The equality after the second inequality follows from optimality of the coupling for $W_1$. The final inequality is the upper MMD-to-Wasserstein comparison in Theorem~\ref{thm:mmd_wasserstein_equiv}, applied on $\mathfrak S_{b,\mathfrak m,r,s}$. Therefore, $\mathfrak T^\pi$ is H\"older-Lipschitz on the embedded law class with exponent $\rho\in(0,1)$. It is not a Banach contraction because the right-hand side has exponent $\rho<1$.

\noindent\textit{Existence.}
Work on the nonempty, closed, bounded, convex invariant set $K\subset\mc E$ of admissible conditional return-law embeddings in Assumption~\ref{ass:regular_bellman_embedding_class}. The ambient space $\mc E$ is reflexive, so $K$ is weakly compact by Kakutani's theorem; the Eberlein-\v{S}mulian theorem gives the equivalent weak sequential compactness used in Lemma~\ref{lem:AGP}. The operator $\mathfrak T^\pi$ maps $K$ into itself and is weakly sequentially continuous by Lemma~\ref{lemma:weak_cts_embed_operator}. Neither the H\"older bound \eqref{eq:operator_holder_sup} nor compatibility between the $\mc E$-norm and $D_k$ is required for this existence argument. The H\"older bound is a separate stability statement and is not used in the existence, uniqueness, residual-bound, or identification arguments below.
Therefore, by Lemma~\ref{lem:AGP}, $\mathfrak T^\pi$ has at least one fixed point in $K$. That is, there exists $\mu^\pi\in K$ such that
\[
\mathfrak T^\pi\mu^\pi=\mu^\pi
\]
Because $K$ consists of embeddings of probability laws and the Mat\'ern kernel is characteristic on the regular class, this embedding fixed point corresponds to a conditional return law $\mathbb P^\pi$ satisfying $\T^\pi\mathbb P^\pi=\mathbb P^\pi$. Equivalently, if $\bZ^\pi$ has conditional law $\mathbb P^\pi$, then $\T^\pi\bZ^\pi\overset{D}{=}\bZ^\pi$.

\noindent\textit{Uniqueness.}
Suppose there exist two fixed points $\bZ^{*,\pi}_1$ and $\bZ^{*,\pi}_2$ satisfying the distributional Bellman equation whose conditional laws lie in $\mathfrak S_{b,\mathfrak m,r,s}$ at every conditioning input, as holds for fixed points represented in $K$ by Assumption~\ref{ass:regular_bellman_embedding_class}(i). Let $\mathbb P_1(\cdot\mid\bs,\ba)$ and $\mathbb P_2(\cdot\mid\bs,\ba)$ denote their conditional laws at a fixed $(\bs,\ba)$, and define the corresponding Bellman-updated laws analogously. Since each $\bZ^{*,\pi}_i$ is a fixed point in distribution,
\begin{equation}\label{eq:FP_UQN_gamma=0_revised}
\Bgamma_k\{(\T^\pi\mathbb P_i)(\cdot\mid\bs,\ba),\mathbb P_i(\cdot\mid\bs,\ba)\}
=0
\qquad
i\in\{1,2\}
\end{equation}
By the upper inequality in Theorem~\ref{thm:mmd_wasserstein_equiv}, \eqref{eq:FP_UQN_gamma=0_revised} also implies
\begin{equation}\label{eq:W_self_zero_revised}
W_1\{(\T^\pi\mathbb P_i)(\cdot\mid\bs,\ba),\mathbb P_i(\cdot\mid\bs,\ba)\}
=0
\qquad
i\in\{1,2\}
\end{equation}
so both fixed points are fixed points under the Wasserstein metric as well.

\noindent Suppose, toward a contradiction, that the two fixed points are distinct in law at some $(\bs,\ba)$. Because $k_{\mc Z}$ is characteristic, $\Bgamma_k$ is a metric on the regular class, and hence
\begin{equation}\label{eq:MMD_pos_revised}
\mathbb P_1(\cdot\mid\bs,\ba)\ne\mathbb P_2(\cdot\mid\bs,\ba)
\quad\Longleftrightarrow\quad
\Bgamma_k\{\mathbb P_1(\cdot\mid\bs,\ba),\mathbb P_2(\cdot\mid\bs,\ba)\}>0
\end{equation}
Consider the maximal conditional Wasserstein distance
\[
d_1(\mathbb P,\mathbb Q)
:=
\sup_{(\bs,\ba)}
W_1\{\mathbb P(\cdot\mid\bs,\ba),\mathbb Q(\cdot\mid\bs,\ba)\}
\]
This quantity is finite on the regular class. Since $r>1$, the moment constraint gives $M_1[\mathbb P(\cdot\mid\bs,\ba)]\le M_r[\mathbb P(\cdot\mid\bs,\ba)]\le \mathfrak m$ for every conditioning input. The independent coupling of the two marginals then gives $W_1\{\mathbb P(\cdot\mid\bs,\ba),\mathbb Q(\cdot\mid\bs,\ba)\}\le M_1[\mathbb P(\cdot\mid\bs,\ba)]+M_1[\mathbb Q(\cdot\mid\bs,\ba)]\le2\mathfrak m$. Thus $d_1(\mathbb P_1,\mathbb P_2)\le2\mathfrak m<\infty$. For policy evaluation, the same coupling argument as in the scalar case gives the vector-valued contraction in $d_1$. Couple the one-step reward and next input identically for the two Bellman updates and, conditional on each next input $\bX'$, use an optimal coupling of the two continuation returns under the Euclidean norm on $\R^d$. Then, for every current input $\bx$,
\begin{align*}
W_1\{(\T^\pi\mathbb P_1)(\cdot\mid\bx),(\T^\pi\mathbb P_2)(\cdot\mid\bx)\}
&\le
\gamma\,\E\!\left[
W_1\{\mathbb P_1(\cdot\mid\bX'),\mathbb P_2(\cdot\mid\bX')\}
\mid \bX=\bx
\right]\\
&\le
\gamma\,d_1(\mathbb P_1,\mathbb P_2)\end{align*}
Taking the supremum over $\bx$ gives $d_1(\T^\pi\mathbb P_1,\T^\pi\mathbb P_2)\le\gamma d_1(\mathbb P_1,\mathbb P_2)$. Since the two laws are fixed points,
\[
d_1(\mathbb P_1,\mathbb P_2)
=
d_1(\T^\pi\mathbb P_1,\T^\pi\mathbb P_2)
\le
\gamma d_1(\mathbb P_1,\mathbb P_2)
\]
Since $\gamma\in(0,1)$, this forces $d_1(\mathbb P_1,\mathbb P_2)=0$. Hence $W_1\{\mathbb P_1(\cdot\mid\bs,\ba),\mathbb P_2(\cdot\mid\bs,\ba)\}=0$ for every $(\bs,\ba)$, so $\mathbb P_1(\cdot\mid\bs,\ba)=\mathbb P_2(\cdot\mid\bs,\ba)$ for every $(\bs,\ba)$. By the equivalence \eqref{eq:MMD_pos_revised}, this contradicts the supposition that the two laws differ at some $(\bs,\ba)$. We conclude that $\bZ^{*,\pi}_1\overset{D}{=}\bZ^{*,\pi}_2$ at every state-action pair, and therefore the fixed point is unique.

\noindent\textit{Residual bound.}
Let $\mathbb Q$ be any candidate conditional law represented in $K$ and let $\mathbb P^\pi$ denote the unique fixed-point law. Both laws lie in the regular class, so $d_1(\mathbb Q,\mathbb P^\pi)\le2\mathfrak m<\infty$ by the moment argument above. Since $\T^\pi\mathbb P^\pi=\mathbb P^\pi$,
\begin{align*}
d_1(\mathbb Q,\mathbb P^\pi)
&\le
\sup_{\bx\in\mc X}
W_1\{\mathbb Q(\cdot\mid\bx),(\T^\pi\mathbb Q)(\cdot\mid\bx)\}
+
d_1(\T^\pi\mathbb Q,\T^\pi\mathbb P^\pi)\\
&\le
\sup_{\bx\in\mc X}
W_1\{\mathbb Q(\cdot\mid\bx),(\T^\pi\mathbb Q)(\cdot\mid\bx)\}
+
\gamma d_1(\mathbb Q,\mathbb P^\pi)
\end{align*}
Because $\gamma\in(0,1)$ and $d_1(\mathbb Q,\mathbb P^\pi)\le2\mathfrak m<\infty$, the last display rearranges to $(1-\gamma)\,d_1(\mathbb Q,\mathbb P^\pi)\le\sup_{\bx\in\mc X}W_1\{\mathbb Q(\cdot\mid\bx),(\T^\pi\mathbb Q)(\cdot\mid\bx)\}$. Applying the upper comparison of Theorem~\ref{thm:mmd_wasserstein_equiv} pointwise in $\bx$, which is valid because $\mathbb Q(\cdot\mid\bx)$ and $(\T^\pi\mathbb Q)(\cdot\mid\bx)$ lie in the same regular class, and using $\sup_{\bx}a_{\bx}^\rho=(\sup_{\bx}a_{\bx})^\rho$ for nonnegative $a_{\bx}$ and $0<\rho<1$, gives
\[
d_1(\mathbb Q,\mathbb P^\pi)
\le
\frac{C_*}{1-\gamma}
\left[
\sup_{\bx\in\mc X}
\Bgamma_k\{\mathbb Q(\cdot\mid\bx),(\T^\pi\mathbb Q)(\cdot\mid\bx)\}
\right]^\rho
\]
This is the residual-to-fixed-point bound in the theorem. If the right-hand side is zero, then $d_1(\mathbb Q,\mathbb P^\pi)=0$, and the Wasserstein uniqueness argument above implies $\mathbb Q(\cdot\mid\bx)=\mathbb P^\pi(\cdot\mid\bx)$ for every $\bx\in\mc X$.

\noindent\textit{Identification.}
Under the reward conditions of Lemma~\ref{lem:invariance_identification}, part (ii) shows that the true conditional return law map $\mc V^\pi$ belongs to $\mathfrak S_{b,\mathfrak m,r,s}$ for every $\bx$ and satisfies the distributional Bellman equation. It is therefore a fixed point of $\T^\pi$ at $d_1$-distance at most $2\mathfrak m$ from $\mathbb P^\pi$. The contraction argument gives $d_1(\mc V^\pi,\mathbb P^\pi)=0$, so $\mathbb P^\pi(\cdot\mid\bx)=\mc V^\pi_{\bx}$ for every $\bx\in\mc X$.
\end{proof}

\subsection{Proof of Theorem \ref{thm:pointwise_error_bound_intro} {\small(Pointwise Error Bound for the Conditional Mean Embedding)}} \label{proof:pointwise_error_bound_intro}

\begin{proof}
\noindent\textit{Notation and standing conventions.}
Write $\mc X=\mc S\times\mc A$ and abbreviate the feature sections by $\phi(\bx):= k_{\mc X}(\bx,\cdot)\in\mc H_{\mc X}$ and $\psi(\bz):=k_{\mc Z}(\bz,\cdot)\in\mc H_{\mc Z}$. For $u\in\mc H_{\mc Z}$ and $v\in\mc H_{\mc X}$, let $u\otimes v:\mc H_{\mc X}\to\mc H_{\mc Z}$ denote the rank-one operator $(u\otimes v)h:=\langle v,h\rangle_{\mc H_{\mc X}}\,u$, which is Hilbert-Schmidt with $\|u\otimes v\|_{\mathrm{HS}}=\|u\|_{\mc H_{\mc Z}}\|v\|_{\mc H_{\mc X}}$. The uncentered population covariance operators are the Bochner integrals
\[
\mbC_{\mc{XX}}:=\E\big[\phi(\bX)\otimes\phi(\bX)\big]:\mc H_{\mc X}\to\mc H_{\mc X},
\qquad
\mbC_{\mc{ZX}}:=\E\big[\psi(\bZ)\otimes\phi(\bX)\big]:\mc H_{\mc X}\to\mc H_{\mc Z}
\]
which exist because the integrands are almost surely bounded in Hilbert-Schmidt norm by $\kappa_{\mc X}^2$ and $\kappa_{\mc X}\kappa_{\mc Z}$, respectively. Their empirical counterparts are
\[
\widehat{\mbC}_{\mc{XX}}:=\frac1n\sum_{i=1}^n\phi(\bX_i)\otimes\phi(\bX_i),
\qquad
\widehat{\mbC}_{\mc{ZX}}:=\frac1n\sum_{i=1}^n\psi(\bZ_i)\otimes\phi(\bX_i)
\]
The operator $\mbC_{\mc{XX}}$ is self-adjoint and positive: for every $h\in\mc H_{\mc X}$, $\langle h,\mbC_{\mc{XX}}h\rangle_{\mc H_{\mc X}}=\E\{h(\bX)^2\}\ge0$ by the reproducing property, and the same computation with the empirical measure shows that $\widehat{\mbC}_{\mc{XX}}$ is self-adjoint and positive as well. The estimator under study and its regularized population counterpart are
\[
\widehat\mu_{\bZ\mid\bx}
:=\widehat{\mbC}_{\mc{ZX}}\big(\widehat{\mbC}_{\mc{XX}}+\lambda\mathbb I\big)^{-1}\phi(\bx),
\qquad
\mu^\lambda_{\bZ\mid\bx}
:=\mbC_{\mc{ZX}}\big(\mbC_{\mc{XX}}+\lambda\mathbb I\big)^{-1}\phi(\bx)
\]
Both are well defined for every $\lambda>0$ because the regularized operators are boundedly invertible, with spectra in $[\lambda,\infty)$. The unregularized composition $\mbC_{\mc{ZX}}\mbC_{\mc{XX}}^{-1}$ need not exist on all of $\mc H_{\mc X}$ in continuous domains. Following \citet{fukumizu2013kernel}, we therefore work with the regularized map $\mu^\lambda_{\bZ\mid\bx}$. Under the well-specifiedness condition in Theorem~\ref{thm:pointwise_error_bound_intro}, for every $f\in\mc H_{\mc Z}$ the conditional mean function $m_f:=\E\{f(\bZ)\mid\bX=\cdot\,\}$ admits a version in $\mc H_{\mc X}$, and $\mu_{\bZ\mid\bx}\in\mc H_{\mc Z}$ denotes the corresponding population conditional mean embedding, characterized by
\begin{equation}\label{eq:pw_cme_char}
\langle f,\ \mu_{\bZ\mid\bx}\rangle_{\mc H_{\mc Z}}
=
m_f(\bx)
=
\langle m_f,\ \phi(\bx)\rangle_{\mc H_{\mc X}}
\qquad\text{for all }f\in\mc H_{\mc Z}
\end{equation}
where the second equality is the reproducing property; because $m_f\in\mc H_{\mc X}$, the evaluation at the fixed conditioning point $\bx$ is exact and no almost-sure qualification is needed.

\medskip
\noindent\textit{Step 0: elementary operator bounds.}
The following four facts are used repeatedly.
\begin{enumerate}[label=(\alph*)]
\vspace{-0.5em}\item\label{it:pw_feat} \emph{Feature bounds.} $\|\phi(\bx)\|_{\mc H_{\mc X}}= k_{\mc X}(\bx,\bx)^{1/2}\le\kappa_{\mc X}$ for every $\bx\in\mc X$, and $\|\psi(\bz)\|_{\mc H_{\mc Z}}=k_{\mc Z}(\bz,\bz)^{1/2}\le\kappa_{\mc Z}$ for every $\bz\in\mc Z$.
\vspace{-0.5em}\item\label{it:pw_cov} \emph{Covariance bounds.} By the triangle inequality for Bochner integrals and the rank-one identity $\|\psi(\bz)\otimes\phi(\bx)\|_{\mathrm{HS}}=\|\psi(\bz)\|\,\|\phi(\bx)\|$,
\[
\|\mbC_{\mc{ZX}}\|_{\mathrm{op}}
\le
\|\mbC_{\mc{ZX}}\|_{\mathrm{HS}}
\le
\E\,\big\|\psi(\bZ)\otimes\phi(\bX)\big\|_{\mathrm{HS}}
\le
\kappa_{\mc X}\kappa_{\mc Z}
\]
and identically $\|\widehat{\mbC}_{\mc{ZX}}\|_{\mathrm{op}}\le\kappa_{\mc X}\kappa_{\mc Z}$ almost surely; the same argument gives $\|\mbC_{\mc{XX}}\|_{\mathrm{op}}\le\kappa_{\mc X}^2$.
\vspace{-0.5em}\item\label{it:pw_res} \emph{Resolvent bounds.} If $A$ is self-adjoint and positive on a Hilbert space and $\lambda>0$, then $\sigma(A+\lambda\mathbb I)\subseteq[\lambda,\infty)$, so by the spectral mapping theorem
\[
\big\|(A+\lambda\mathbb I)^{-1}\big\|_{\mathrm{op}}
=
\sup_{t\in\sigma(A)}\frac{1}{t+\lambda}
\le
\frac1\lambda
\]
This applies to both $A=\widehat{\mbC}_{\mc{XX}}$ and $A=\mbC_{\mc{XX}}$.
\vspace{-0.5em}\item\label{it:pw_norms} \emph{Norm domination.} $\|\cdot\|_{\mathrm{op}}\le\|\cdot\|_{\mathrm{HS}}$ for every Hilbert-Schmidt operator.
\end{enumerate}

\medskip
\noindent\textit{Step 1: bias-variance decomposition.}
By the triangle inequality in $\mc H_{\mc Z}$,
\begin{equation}\label{eq:pw_decomp}
\big\|\widehat\mu_{\bZ\mid\bx}-\mu_{\bZ\mid\bx}\big\|_{\mc H_{\mc Z}}
\le
\underbrace{\big\|\widehat\mu_{\bZ\mid\bx}-\mu^\lambda_{\bZ\mid\bx}\big\|_{\mc H_{\mc Z}}}_{\text{sampling error}}
+
\underbrace{\big\|\mu^\lambda_{\bZ\mid\bx}-\mu_{\bZ\mid\bx}\big\|_{\mc H_{\mc Z}}}_{\text{regularization bias}}
\end{equation}
The sampling error is stochastic and is controlled on the covariance-concentration event; the regularization bias is deterministic and is controlled by the source condition.

\medskip
\noindent\textit{Step 2: sampling error.}
Adding and subtracting $\mbC_{\mc{ZX}}(\widehat{\mbC}_{\mc{XX}}+\lambda\mathbb I)^{-1}\phi(\bx)$,
\begin{equation}\label{eq:pw_T1T2}
\widehat\mu_{\bZ\mid\bx}-\mu^\lambda_{\bZ\mid\bx}
=
\underbrace{\big(\widehat{\mbC}_{\mc{ZX}}-\mbC_{\mc{ZX}}\big)\big(\widehat{\mbC}_{\mc{XX}}+\lambda\mathbb I\big)^{-1}\phi(\bx)}_{=:\,\mathrm{T}_1}
+
\underbrace{\mbC_{\mc{ZX}}\Big[\big(\widehat{\mbC}_{\mc{XX}}+\lambda\mathbb I\big)^{-1}-\big(\mbC_{\mc{XX}}+\lambda\mathbb I\big)^{-1}\Big]\phi(\bx)}_{=:\,\mathrm{T}_2}
\end{equation}
\emph{Term $\mathrm{T}_1$.} By submultiplicativity of the operator norm, Step~0\ref{it:pw_res}, Step~0\ref{it:pw_feat}, and Step~0\ref{it:pw_norms},
\begin{equation}\label{eq:pw_T1}
\|\mathrm{T}_1\|_{\mc H_{\mc Z}}
\le
\big\|\widehat{\mbC}_{\mc{ZX}}-\mbC_{\mc{ZX}}\big\|_{\mathrm{op}}
\,\big\|(\widehat{\mbC}_{\mc{XX}}+\lambda\mathbb I)^{-1}\big\|_{\mathrm{op}}
\,\big\|\phi(\bx)\big\|_{\mc H_{\mc X}}
\le
\frac{\kappa_{\mc X}}{\lambda}\,
\big\|\widehat{\mbC}_{\mc{ZX}}-\mbC_{\mc{ZX}}\big\|_{\mathrm{HS}}
\end{equation}
\emph{Term $\mathrm{T}_2$.} With $A:=\widehat{\mbC}_{\mc{XX}}+\lambda\mathbb I$ and $B:=\mbC_{\mc{XX}}+\lambda\mathbb I$, both boundedly invertible, the resolvent identity
\[
A^{-1}-B^{-1}
=
A^{-1}\big(B-A\big)B^{-1}
\]
(verified by multiplying out: $A^{-1}BB^{-1}-A^{-1}AB^{-1}=A^{-1}-B^{-1}$) gives, since $B-A=\mbC_{\mc{XX}}-\widehat{\mbC}_{\mc{XX}}$,
\[
\big(\widehat{\mbC}_{\mc{XX}}+\lambda\mathbb I\big)^{-1}-\big(\mbC_{\mc{XX}}+\lambda\mathbb I\big)^{-1}
=
\big(\widehat{\mbC}_{\mc{XX}}+\lambda\mathbb I\big)^{-1}
\big(\mbC_{\mc{XX}}-\widehat{\mbC}_{\mc{XX}}\big)
\big(\mbC_{\mc{XX}}+\lambda\mathbb I\big)^{-1}
\]
Therefore, by Step~0\ref{it:pw_cov}, Step~0\ref{it:pw_res} applied twice, and Step~0\ref{it:pw_feat},
\begin{equation}\label{eq:pw_T2}
\|\mathrm{T}_2\|_{\mc H_{\mc Z}}
\le
\|\mbC_{\mc{ZX}}\|_{\mathrm{op}}\,
\frac1\lambda\,
\big\|\widehat{\mbC}_{\mc{XX}}-\mbC_{\mc{XX}}\big\|_{\mathrm{op}}\,
\frac1\lambda\,
\kappa_{\mc X}
\le
\frac{\kappa_{\mc X}^{2}\kappa_{\mc Z}}{\lambda^{2}}\,
\big\|\widehat{\mbC}_{\mc{XX}}-\mbC_{\mc{XX}}\big\|_{\mathrm{op}}
\end{equation}
Let $\Omega_{\bZ}$ denote the event on which both covariance bounds assumed in the theorem hold, namely $\|\widehat{\mbC}_{\mc{ZX}}-\mbC_{\mc{ZX}}\|_{\mathrm{HS}}\le\varepsilon_{\mc{ZX}}(n,\delta)$ and $\|\widehat{\mbC}_{\mc{XX}}-\mbC_{\mc{XX}}\|_{\mathrm{op}}\le\varepsilon_{\mc{XX}}(n,\delta)$. By hypothesis, $\p(\Omega_{\bZ})\ge1-\delta$; Lemma~\ref{lem:mixing_concentration} in Supplementary Material~\ref{app:mixing_concentration} gives radii of this form for a single geometrically mixing sequence. Combining \eqref{eq:pw_T1T2}-\eqref{eq:pw_T2}, on $\Omega_{\bZ}$,
\begin{equation}\label{eq:pw_sampling}
\big\|\widehat\mu_{\bZ\mid\bx}-\mu^\lambda_{\bZ\mid\bx}\big\|_{\mc H_{\mc Z}}
\le
\frac{\kappa_{\mc X}}{\lambda}\,\varepsilon_{\mc{ZX}}(n,\delta)
+
\frac{\kappa_{\mc X}^{2}\kappa_{\mc Z}}{\lambda^{2}}\,\varepsilon_{\mc{XX}}(n,\delta)
\end{equation}

\medskip
\noindent\textit{Step 3: regularization bias under the source condition.}
Let $W$ denote the assumed bounded extension of $\mbC_{\mc{ZX}}\mbC_{\mc{XX}}^{-1/2}$ to all of $\mc H_{\mc X}$. We first record two structural identities.

\smallskip
\noindent\emph{(3a) Factorization $\mbC_{\mc{ZX}}=W\mbC_{\mc{XX}}^{1/2}$ on all of $\mc H_{\mc X}$.}
On $\mathrm{Range}(\mbC_{\mc{XX}}^{1/2})$ this holds by the definition of $W$. For the orthogonal complement, note $\overline{\mathrm{Range}}(\mbC_{\mc{XX}}^{1/2})^{\perp}=\ker(\mbC_{\mc{XX}}^{1/2})=\ker(\mbC_{\mc{XX}})$, and that $\mbC_{\mc{ZX}}$ annihilates $\ker(\mbC_{\mc{XX}})$: if $\mbC_{\mc{XX}}h=0$ then $\E\{h(\bX)^2\}=\langle h,\mbC_{\mc{XX}}h\rangle=0$, so $h(\bX)=0$ almost surely, and hence $\mbC_{\mc{ZX}}h=\E\{\psi(\bZ)\,h(\bX)\}=0$. Since $W\mbC_{\mc{XX}}^{1/2}$ also annihilates $\ker(\mbC_{\mc{XX}})$, both sides agree on $\mathrm{Range}(\mbC_{\mc{XX}}^{1/2})\oplus\ker(\mbC_{\mc{XX}})$, which is dense, and both are bounded; hence $\mbC_{\mc{ZX}}=W\mbC_{\mc{XX}}^{1/2}$ everywhere, and by taking adjoints $\mbC_{\mc{ZX}}^{*}=\mbC_{\mc{XX}}^{1/2}W^{*}$.

\smallskip
\noindent\emph{(3b) Tower identity $\mbC_{\mc{XX}}m_f=\mbC_{\mc{ZX}}^{*}f$ for every $f\in\mc H_{\mc Z}$.}
For arbitrary $h\in\mc H_{\mc X}$, the reproducing property, the tower property of conditional expectation, and the definition of the two covariance operators give
\[
\begin{aligned}
\big\langle h,\ \mbC_{\mc{XX}}m_f\big\rangle_{\mc H_{\mc X}}
&=
\E\big\{h(\bX)\,m_f(\bX)\big\}
=
\E\big\{h(\bX)\,\E[f(\bZ)\mid\bX]\big\}
\\
&=
\E\big\{h(\bX)f(\bZ)\big\}
=
\big\langle \mbC_{\mc{ZX}}h,\ f\big\rangle_{\mc H_{\mc Z}}
\\
&=
\big\langle h,\ \mbC_{\mc{ZX}}^{*}f\big\rangle_{\mc H_{\mc X}}
\end{aligned}
\]
Since $h$ is arbitrary, $\mbC_{\mc{XX}}m_f=\mbC_{\mc{ZX}}^{*}f$.

\smallskip
\noindent\emph{(3c) Representation of the population embedding.}
Combining (3a) and (3b), $\mbC_{\mc{XX}}^{1/2}\big(\mbC_{\mc{XX}}^{1/2}m_f-W^{*}f\big)=\mbC_{\mc{XX}}m_f-\mbC_{\mc{ZX}}^{*}f=0$, so
\[
v_f:=\mbC_{\mc{XX}}^{1/2}m_f-W^{*}f\in\ker\big(\mbC_{\mc{XX}}^{1/2}\big)=\ker\big(\mbC_{\mc{XX}}\big)
\]
For any exponent $a>0$, spectral calculus for the positive self-adjoint operator $\mbC_{\mc{XX}}$ gives $\ker(\mbC_{\mc{XX}}^{a})=\ker(\mbC_{\mc{XX}})$, so in particular $\mbC_{\mc{XX}}^{\tau}v_f=0$. Now fix $f\in\mc H_{\mc Z}$ and use the characterization \eqref{eq:pw_cme_char}, the source condition $\phi(\bx)=\mbC_{\mc{XX}}^{1/2+\tau}g_{\bx}$, and self-adjointness of the fractional powers:
\[
\begin{aligned}
\langle f,\mu_{\bZ\mid\bx}\rangle_{\mc H_{\mc Z}}
&=
\big\langle m_f,\ \mbC_{\mc{XX}}^{1/2+\tau}g_{\bx}\big\rangle_{\mc H_{\mc X}}
\\
&=
\big\langle \mbC_{\mc{XX}}^{\tau}\mbC_{\mc{XX}}^{1/2}m_f,\ g_{\bx}\big\rangle_{\mc H_{\mc X}}
\\
&=
\big\langle \mbC_{\mc{XX}}^{\tau}\big(W^{*}f+v_f\big),\ g_{\bx}\big\rangle_{\mc H_{\mc X}}
\\
&=
\big\langle f,\ W\mbC_{\mc{XX}}^{\tau}g_{\bx}\big\rangle_{\mc H_{\mc Z}}
\end{aligned}
\]
where the last equality uses $\mbC_{\mc{XX}}^{\tau}v_f=0$ and $\langle \mbC_{\mc{XX}}^{\tau}W^{*}f,g_{\bx}\rangle=\langle f,W\mbC_{\mc{XX}}^{\tau}g_{\bx}\rangle$. Since $f$ is arbitrary,
\begin{equation}\label{eq:pw_mu_rep}
\mu_{\bZ\mid\bx}
=
W\,\mbC_{\mc{XX}}^{\tau}\,g_{\bx}
\end{equation}

\smallskip
\noindent\emph{(3d) Representation of the regularized embedding.}
Substituting the source condition into the definition of $\mu^\lambda_{\bZ\mid\bx}$ and using (3a) together with the fact that all functions of the single self-adjoint operator $\mbC_{\mc{XX}}$ commute with one another (functional calculus),
\begin{equation}\label{eq:pw_mulambda_rep}
\mu^\lambda_{\bZ\mid\bx}
=
\mbC_{\mc{ZX}}\big(\mbC_{\mc{XX}}+\lambda\mathbb I\big)^{-1}\mbC_{\mc{XX}}^{1/2+\tau}g_{\bx}
=
W\,\mbC_{\mc{XX}}^{1/2}\big(\mbC_{\mc{XX}}+\lambda\mathbb I\big)^{-1}\mbC_{\mc{XX}}^{1/2+\tau}g_{\bx}
=
W\big(\mbC_{\mc{XX}}+\lambda\mathbb I\big)^{-1}\mbC_{\mc{XX}}^{1+\tau}g_{\bx}
\end{equation}

\smallskip
\noindent\emph{(3e) Bias bound.}
Subtracting \eqref{eq:pw_mu_rep} from \eqref{eq:pw_mulambda_rep} and simplifying inside the functional calculus,
\[
\big(\mbC_{\mc{XX}}+\lambda\mathbb I\big)^{-1}\mbC_{\mc{XX}}^{1+\tau}-\mbC_{\mc{XX}}^{\tau}
=
\big(\mbC_{\mc{XX}}+\lambda\mathbb I\big)^{-1}
\Big[\mbC_{\mc{XX}}^{1+\tau}-\big(\mbC_{\mc{XX}}+\lambda\mathbb I\big)\mbC_{\mc{XX}}^{\tau}\Big]
=
-\lambda\big(\mbC_{\mc{XX}}+\lambda\mathbb I\big)^{-1}\mbC_{\mc{XX}}^{\tau}
\]
so that
\begin{equation}\label{eq:pw_bias_exact}
\mu^\lambda_{\bZ\mid\bx}-\mu_{\bZ\mid\bx}
=
-\lambda\,W\big(\mbC_{\mc{XX}}+\lambda\mathbb I\big)^{-1}\mbC_{\mc{XX}}^{\tau}g_{\bx}
\end{equation}
By the functional calculus, $\big\|\lambda(\mbC_{\mc{XX}}+\lambda\mathbb I)^{-1}\mbC_{\mc{XX}}^{\tau}\big\|_{\mathrm{op}}\le\sup_{t\ge0}\frac{\lambda t^{\tau}}{t+\lambda}$. At $t=0$ the ratio equals zero. For $t>0$, the weighted arithmetic-geometric mean inequality with exponent $\tau\in(0,1]$ gives $t^{\tau}\lambda^{1-\tau}\le\tau t+(1-\tau)\lambda\le t+\lambda$, whence
\[
\sup_{t>0}\frac{\lambda\,t^{\tau}}{t+\lambda}
\le
\sup_{t>0}\frac{\lambda\,t^{\tau}}{t^{\tau}\lambda^{1-\tau}}
=
\lambda^{\tau},
\qquad\text{and therefore}\qquad
\sup_{t\ge0}\frac{\lambda\,t^{\tau}}{t+\lambda}
\le
\lambda^{\tau}
\]
Taking norms in \eqref{eq:pw_bias_exact} therefore yields the deterministic bound
\begin{equation}\label{eq:pw_bias}
\big\|\mu^\lambda_{\bZ\mid\bx}-\mu_{\bZ\mid\bx}\big\|_{\mc H_{\mc Z}}
\le
\big\|W\big\|_{\mathrm{op}}\,\lambda^{\tau}\,\|g_{\bx}\|_{\mc H_{\mc X}}
=
c_{\bx}\,\lambda^{\tau},
\qquad
c_{\bx}=\big\|\mbC_{\mc{ZX}}\mbC_{\mc{XX}}^{-1/2}\big\|_{\mathrm{op}}\,\|g_{\bx}\|_{\mc H_{\mc X}}
\end{equation}
No probabilistic argument is involved in \eqref{eq:pw_bias}.

\medskip
\noindent\textit{Step 4: the pointwise bound.}
Substituting \eqref{eq:pw_sampling} and \eqref{eq:pw_bias} into \eqref{eq:pw_decomp} shows that, on the event $\Omega_{\bZ}$ of probability at least $1-\delta$,
\begin{equation}\label{eq:pw_final}
\big\|\widehat\mu_{\bZ\mid\bx}-\mu_{\bZ\mid\bx}\big\|_{\mc H_{\mc Z}}
\le
\frac{\kappa_{\mc X}}{\lambda}\,\varepsilon_{\mc{ZX}}(n,\delta)
+
\frac{\kappa_{\mc X}^{2}\kappa_{\mc Z}}{\lambda^{2}}\,\varepsilon_{\mc{XX}}(n,\delta)
+
c_{\bx}\,\lambda^{\tau}
=:
\mathfrak B_{\bZ}
\end{equation}
which is the assertion of the theorem.
\end{proof}

\medskip
\begin{proof}[Proof of Corollary~\ref{cor:bellman_residual_error}]
\noindent\textit{Squared Bellman-discrepancy bound.}
The argument in Steps 0-4 of the proof of Theorem~\ref{thm:pointwise_error_bound_intro} uses only \ref{it:pw_feat} boundedness of the two kernels, \ref{it:pw_cov} the covariance-concentration event, the well-specifiedness condition, and the source condition at $\bx$; it remains valid when $\bZ$ is replaced by any response variable satisfying the same conditions. By hypothesis, the Bellman-target pair $(\bX,(\T^\pi\bZ)(\bX))$ satisfies these conditions, with cross-covariance radius $\varepsilon_{\T\mc Z,\mc X}(n,\delta)$ and the same $\mbC_{\mc{XX}}$-concentration radius $\varepsilon_{\mc{XX}}(n,\delta)$. Writing $\Omega_{\T^\pi\bZ}$ for the corresponding covariance event, the same argument applied to this pair gives, on $\Omega_{\T^\pi\bZ}$,
\[
\big\|\widehat\mu_{\T^\pi\bZ\mid\bx}-\mu_{\T^\pi\bZ\mid\bx}\big\|_{\mc H_{\mc Z}}
\le
\frac{\kappa_{\mc X}}{\lambda}\,\varepsilon_{\T\mc Z,\mc X}(n,\delta)
+
\frac{\kappa_{\mc X}^{2}\kappa_{\mc Z}}{\lambda^{2}}\,\varepsilon_{\mc{XX}}(n,\delta)
+
c_{\bx,\T^\pi\bZ}\,\lambda^{\tau}
=:
\mathfrak B_{\T^\pi\bZ}
\]
where $c_{\bx,\T^\pi\bZ}$ is the source constant of the Bellman target. By hypothesis, the two events hold jointly with probability at least $1-\delta$. For example, if each event is constructed with failure probability at most $\delta/2$, the union bound gives $\p(\Omega_{\bZ}\cap\Omega_{\T^\pi\bZ})\ge1-\p(\Omega_{\bZ}^{c})-\p(\Omega_{\T^\pi\bZ}^{c})\ge1-\delta$. Work on $\Omega_{\bZ}\cap\Omega_{\T^\pi\bZ}$ for the remainder of the proof and set
\[
\Delta:=\mu_{\bZ\mid\bx}-\mu_{\T^\pi\bZ\mid\bx},
\qquad
\widehat\Delta:=\widehat\mu_{\bZ\mid\bx}-\widehat\mu_{\T^\pi\bZ\mid\bx}
\]
First, each population embedding is a Bochner conditional expectation of kernel sections, so by the conditional Jensen inequality for Bochner integrals and Step~0\ref{it:pw_feat},
\[
\|\mu_{\bZ\mid\bx}\|_{\mc H_{\mc Z}}
\le
\E\big\{\|\psi(\bZ)\|_{\mc H_{\mc Z}}\mid\bX=\bx\big\}
\le\kappa_{\mc Z},
\qquad
\|\mu_{\T^\pi\bZ\mid\bx}\|_{\mc H_{\mc Z}}\le\kappa_{\mc Z}
\]
(the second bound holds because the Mat\'ern kernel satisfies $k_{\mc Z}(\by,\by)\le\kappa_{\mc Z}^2$ everywhere on $\R^d$, hence also at the Bellman target values), and therefore
\begin{equation}\label{eq:pw_Delta_bound}
\|\Delta\|_{\mc H_{\mc Z}}
\le
\|\mu_{\bZ\mid\bx}\|_{\mc H_{\mc Z}}+\|\mu_{\T^\pi\bZ\mid\bx}\|_{\mc H_{\mc Z}}
\le
2\kappa_{\mc Z}
\end{equation}
Second, by the triangle inequality and the two pointwise bounds established above,
\begin{equation}\label{eq:pw_Delta_err}
\|\widehat\Delta-\Delta\|_{\mc H_{\mc Z}}
\le
\big\|\widehat\mu_{\bZ\mid\bx}-\mu_{\bZ\mid\bx}\big\|_{\mc H_{\mc Z}}
+
\big\|\widehat\mu_{\T^\pi\bZ\mid\bx}-\mu_{\T^\pi\bZ\mid\bx}\big\|_{\mc H_{\mc Z}}
\le
\mathfrak B_{\bZ}+\mathfrak B_{\T^\pi\bZ}
\end{equation}
Third, for any $u,v$ in a real Hilbert space, $\|u\|^2-\|v\|^2=\langle u-v,\,u+v\rangle$; applying this with $u=\widehat\Delta$, $v=\Delta$ and using the Cauchy-Schwarz inequality,
\[
\Big|\|\widehat\Delta\|_{\mc H_{\mc Z}}^{2}-\|\Delta\|_{\mc H_{\mc Z}}^{2}\Big|
=
\Big|\big\langle \widehat\Delta-\Delta,\ \widehat\Delta+\Delta\big\rangle_{\mc H_{\mc Z}}\Big|
\le
\|\widehat\Delta-\Delta\|_{\mc H_{\mc Z}}
\big(\|\widehat\Delta\|_{\mc H_{\mc Z}}+\|\Delta\|_{\mc H_{\mc Z}}\big)
\]
Bounding $\|\widehat\Delta\|\le\|\Delta\|+\|\widehat\Delta-\Delta\|$ by the triangle inequality and then inserting \eqref{eq:pw_Delta_bound}-\eqref{eq:pw_Delta_err},
\begin{align*}
\Big|\|\widehat\Delta\|_{\mc H_{\mc Z}}^{2}-\|\Delta\|_{\mc H_{\mc Z}}^{2}\Big|
&\le
\|\widehat\Delta-\Delta\|_{\mc H_{\mc Z}}
\big(2\|\Delta\|_{\mc H_{\mc Z}}+\|\widehat\Delta-\Delta\|_{\mc H_{\mc Z}}\big)
=
2\|\Delta\|_{\mc H_{\mc Z}}\|\widehat\Delta-\Delta\|_{\mc H_{\mc Z}}
+
\|\widehat\Delta-\Delta\|_{\mc H_{\mc Z}}^{2}
\\
&\le
4\kappa_{\mc Z}\big\{\mathfrak B_{\bZ}+\mathfrak B_{\T^\pi\bZ}\big\}
+
\big\{\mathfrak B_{\bZ}+\mathfrak B_{\T^\pi\bZ}\big\}^{2}
\end{align*}
Both component bounds hold on $\Omega_{\bZ}\cap\Omega_{\T^\pi\bZ}$, an event of probability at least $1-\delta$, which proves the corollary.
\end{proof}

\subsection{Proof of Corollary~\ref{corr:pointwise_consistency_intro} {\small (Pointwise Consistency)}}
\begin{proof}
Fix $\bx\in\mc X$ and let $(\lambda_n)_{n\ge1}$ and $(\delta_n)_{n\ge1}$ satisfy the conditions of the corollary: $\delta_n\to0$, $\lambda_n\to0$, $\varepsilon_{\mc{ZX}}(n,\delta_n)/\lambda_n\to0$, and $\varepsilon_{\mc{XX}}(n,\delta_n)/\lambda_n^{2}\to0$. Write $\widehat\mu^{(n)}_{\bZ\mid\bx}$ for the estimator computed from the first $n$ observations with regularization $\lambda_n$, and define the deterministic sequence
\[
\mathfrak B_n
:=
\frac{\kappa_{\mc X}}{\lambda_n}\,\varepsilon_{\mc{ZX}}(n,\delta_n)
+
\frac{\kappa_{\mc X}^{2}\kappa_{\mc Z}}{\lambda_n^{2}}\,\varepsilon_{\mc{XX}}(n,\delta_n)
+
c_{\bx}\,\lambda_n^{\tau}
\]
For each $n$, Theorem~\ref{thm:pointwise_error_bound_intro} applied with confidence parameter $\delta_n$ and regularization $\lambda_n$ gives
\[
\p\Big(\big\|\widehat\mu^{(n)}_{\bZ\mid\bx}-\mu_{\bZ\mid\bx}\big\|_{\mc H_{\mc Z}}>\mathfrak B_n\Big)
\le
\delta_n
\]
All three summands of $\mathfrak B_n$ vanish: the first two converge to zero by assumption, and the third does because $\lambda_n\to0$ and $\tau>0$; hence $\mathfrak B_n\to0$. Now fix an arbitrary $t>0$ and choose $n_0=n_0(t)$ such that $\mathfrak B_n\le t$ for all $n\ge n_0$. For every $n\ge n_0$, the inclusion of events $\{\|\widehat\mu^{(n)}_{\bZ\mid\bx}-\mu_{\bZ\mid\bx}\|>t\}\subseteq\{\|\widehat\mu^{(n)}_{\bZ\mid\bx}-\mu_{\bZ\mid\bx}\|>\mathfrak B_n\}$ yields
\[
\p\Big(\big\|\widehat\mu^{(n)}_{\bZ\mid\bx}-\mu_{\bZ\mid\bx}\big\|_{\mc H_{\mc Z}}>t\Big)
\le
\p\Big(\big\|\widehat\mu^{(n)}_{\bZ\mid\bx}-\mu_{\bZ\mid\bx}\big\|_{\mc H_{\mc Z}}>\mathfrak B_n\Big)
\le
\delta_n
\longrightarrow0
\]
Since $t>0$ was arbitrary, $\|\widehat\mu^{(n)}_{\bZ\mid\bx}-\mu_{\bZ\mid\bx}\|_{\mc H_{\mc Z}}\overset{p}{\to}0$.
\end{proof}

\subsection{Proof of Theorem \ref{thm:uniform_error_bound_intro} {\small (Uniform Error Bound)}}
\begin{proof}
Fix $\delta\in(0,1)$ and let
\[
\Omega
:=
\left\{
\big\|\widehat{\mbC}_{\mc{ZX}}-\mbC_{\mc{ZX}}\big\|_{\mathrm{HS}}\le\varepsilon_{\mc{ZX}}(n,\delta)
\right\}
\cap
\left\{
\big\|\widehat{\mbC}_{\mc{XX}}-\mbC_{\mc{XX}}\big\|_{\mathrm{op}}\le\varepsilon_{\mc{XX}}(n,\delta)
\right\}
\]
be the covariance-concentration event assumed in Theorem~\ref{thm:pointwise_error_bound_intro}, with $\p(\Omega)\ge1-\delta$. The key observation is that $\Omega$ constrains the empirical covariance \emph{operators} and therefore does not depend on any evaluation point.

\medskip
\noindent\textit{Step 1: the sampling bound is uniform on $\Omega$.}
Inspecting Steps 0-2 of the proof of Theorem~\ref{thm:pointwise_error_bound_intro}, the decomposition \eqref{eq:pw_T1T2} and the bounds \eqref{eq:pw_T1}-\eqref{eq:pw_T2} are deterministic consequences of the operator inequalities defining $\Omega$. The evaluation point enters only through $\|\phi(\bx)\|_{\mc H_{\mc X}}\le\kappa_{\mc X}$, which is uniform over $\mc X$ by Step~0\ref{it:pw_feat}. Hence, on $\Omega$, simultaneously for every $\bx\in\mc X$,
\[
\big\|\widehat\mu_{\bZ\mid\bx}-\mu^\lambda_{\bZ\mid\bx}\big\|_{\mc H_{\mc Z}}
\le
\frac{\kappa_{\mc X}}{\lambda}\,\varepsilon_{\mc{ZX}}(n,\delta)
+
\frac{\kappa_{\mc X}^{2}\kappa_{\mc Z}}{\lambda^{2}}\,\varepsilon_{\mc{XX}}(n,\delta)
\]

\medskip
\noindent\textit{Step 2: the bias bound is deterministic and uniform.}
Step 3 of the proof of Theorem~\ref{thm:pointwise_error_bound_intro} involves no probabilistic argument and yields, for every $\bx\in\mc X$,
\[
\big\|\mu^\lambda_{\bZ\mid\bx}-\mu_{\bZ\mid\bx}\big\|_{\mc H_{\mc Z}}
\le
\big\|\mbC_{\mc{ZX}}\mbC_{\mc{XX}}^{-1/2}\big\|_{\mathrm{op}}\,\|g_{\bx}\|_{\mc H_{\mc X}}\,\lambda^{\tau}
\le
c_\star\lambda^{\tau}
\]
where the final inequality uses the uniform source condition $\sup_{\bx\in\mc X}\|g_{\bx}\|_{\mc H_{\mc X}}<\infty$ and the definition of $c_\star$.

\medskip
\noindent\textit{Step 3: conclusion.}
Combining the two displays through the triangle inequality \eqref{eq:pw_decomp}, on the event $\Omega$ the pointwise bound
\[
\big\|\widehat\mu_{\bZ\mid\bx}-\mu_{\bZ\mid\bx}\big\|_{\mc H_{\mc Z}}
\le
\frac{\kappa_{\mc X}}{\lambda}\,\varepsilon_{\mc{ZX}}(n,\delta)
+
\frac{\kappa_{\mc X}^{2}\kappa_{\mc Z}}{\lambda^{2}}\,\varepsilon_{\mc{XX}}(n,\delta)
+
c_\star\lambda^{\tau}
\]
holds for every $\bx\in\mc X$. Since the right-hand side does not depend on $\bx$, taking the supremum over $\bx\in\mc X$ proves the theorem. A single operator-level event controls all evaluation points, so no covering-number argument or separate Lipschitz bound is required.

\medskip
\noindent The following deterministic continuity fact is used in the strong-consistency argument. If the state-action feature map satisfies $\|\phi(\bx)-\phi(\by)\|_{\mc H_{\mc X}}\le L_{ k_{\mc X}}\,d(\bx,\by)$, then, by linearity of $\widehat{\mbC}_{\mc{ZX}}(\widehat{\mbC}_{\mc{XX}}+\lambda\mathbb I)^{-1}$ and Step~0 of the proof of Theorem~\ref{thm:pointwise_error_bound_intro},
\[
\big\|\widehat\mu_{\bZ\mid\bx}-\widehat\mu_{\bZ\mid\by}\big\|_{\mc H_{\mc Z}}
\le
\big\|\widehat{\mbC}_{\mc{ZX}}\big\|_{\mathrm{op}}
\big\|\big(\widehat{\mbC}_{\mc{XX}}+\lambda\mathbb I\big)^{-1}\big\|_{\mathrm{op}}
\big\|\phi(\bx)-\phi(\by)\big\|_{\mc H_{\mc X}}
\le
\frac{\kappa_{\mc X}\kappa_{\mc Z}L_{ k_{\mc X}}}{\lambda}\,d(\bx,\by)
\]
for every sample realization, so the empirical map is Lipschitz continuous with a deterministic constant whenever the feature map is; this is used only for measurability considerations and plays no role in the error bound itself.
\end{proof}

\subsection{Proof of Corollary~\ref{cor:uniform_strong_consistency_intro} {\small (Strong Consistency)}}
\begin{proof}
Write $\widehat\mu^{(n)}_{\bZ\mid\bx}$ for the estimator based on the first $n$ observations with regularization $\lambda_n$. For each $n\ge1$, define the deterministic sequence
\[
R_n
:=
\frac{\kappa_{\mc X}}{\lambda_n}\,
\varepsilon_{\mc{ZX}}(n,\delta_n)
+
\frac{\kappa_{\mc X}^{2}\kappa_{\mc Z}}{\lambda_n^{2}}\,
\varepsilon_{\mc{XX}}(n,\delta_n)
+
c_\star\lambda_n^{\tau}
\]
which is the right-hand side of Theorem~\ref{thm:unif_error_bound} evaluated at $(\lambda,\delta)=(\lambda_n,\delta_n)$.

\medskip
\noindent\textit{Step 1: high-probability envelope.}
For each $n$, let $\Omega_n$ denote the covariance-concentration event of Theorem~\ref{thm:unif_error_bound} at level $\delta_n$, so that $\p(\Omega_n^{c})\le\delta_n$ and, on $\Omega_n$, the bound $\|\widehat\mu^{(n)}_{\bZ\mid\bx}-\mu_{\bZ\mid\bx}\|_{\mc H_{\mc Z}}\le R_n$ holds for every $\bx\in\mc X$ simultaneously. Note that $\Omega_n$ is measurable by construction (it is defined through norms of the empirical covariance operators), so no measurability of the supremum itself is needed for the probability statements below.

\medskip
\noindent\textit{Step 2: Borel-Cantelli argument.}
By assumption $\sum_{n=1}^\infty\p(\Omega_n^{c})\le\sum_{n=1}^\infty\delta_n<\infty$, so the first Borel-Cantelli lemma yields $\p(\limsup_{n\to\infty}\Omega_n^{c})=0$. Equivalently, on a measurable event of probability one there exists a finite random index $N_0=N_0(\omega)$ such that $\omega\in\Omega_n$ for every $n\ge N_0$, and hence
\[
\sup_{\bx\in\mc X}
\big\|\widehat\mu^{(n)}_{\bZ\mid\bx}-\mu_{\bZ\mid\bx}\big\|_{\mc H_{\mc Z}}
\le R_n
\qquad\text{for all }n\ge N_0
\]

\medskip
\noindent\textit{Step 3: the envelope vanishes.}
We verify $R_n\to0$ term by term. The first two terms converge to zero by the assumed rate conditions
\[
\frac{\varepsilon_{\mc{ZX}}(n,\delta_n)}{\lambda_n}\to0,
\qquad
\frac{\varepsilon_{\mc{XX}}(n,\delta_n)}{\lambda_n^{2}}\to0
\]
and the third term satisfies $c_\star\lambda_n^{\tau}\to0$ because $\lambda_n\to0$ and $\tau>0$. Hence $R_n\to0$.

\medskip
\noindent\textit{Step 4: almost sure uniform convergence.}
Combining Steps 2 and 3: on an event of probability one, for all $n\ge N_0(\omega)$,
\[
0\le
\sup_{\bx\in\mc X}
\big\|\widehat\mu^{(n)}_{\bZ\mid\bx}-\mu_{\bZ\mid\bx}\big\|_{\mc H_{\mc Z}}
\le R_n\longrightarrow0
\]
so $\sup_{\bx\in\mc X}\|\widehat\mu^{(n)}_{\bZ\mid\bx}-\mu_{\bZ\mid\bx}\|_{\mc H_{\mc Z}}\to0$ almost surely.

\medskip
\noindent\textit{Step 5: strong $L^2(\p_{\bX};\mc H_{\mc Z})$ consistency.}
The integrand $\bx\mapsto\|\widehat\mu^{(n)}_{\bZ\mid\bx}-\mu_{\bZ\mid\bx}\|_{\mc H_{\mc Z}}^2$ is measurable. Indeed, as in the proof of Theorem~\ref{thm:pointwise_error_bound_intro}, the kernels are bounded and measurable with separable RKHSs, so the feature map $\bx\mapsto\phi(\bx)$ is strongly measurable and hence so is $\bx\mapsto\widehat\mu^{(n)}_{\bZ\mid\bx}=\widehat{\mbC}_{\mc{ZX}}(\widehat{\mbC}_{\mc{XX}}+\lambda_n\mathbb I)^{-1}\phi(\bx)$ for each realization of the data; the population map $\bx\mapsto\mu_{\bZ\mid\bx}$ is weakly measurable because $\langle f,\mu_{\bZ\mid\bx}\rangle_{\mc H_{\mc Z}}=m_f(\bx)$ with $m_f\in\mc H_{\mc X}$ for every $f$ in a countable dense subset of $\mc H_{\mc Z}$, and therefore strongly measurable by the Pettis measurability theorem. Since $\p_{\bX}$ is a probability measure supported on $\mc X$, the pointwise-in-$\bx$ bound of Step 1 gives, on $\Omega_n$,
\[
\int_{\mc X}
\big\|\widehat\mu^{(n)}_{\bZ\mid\bx}-\mu_{\bZ\mid\bx}\big\|_{\mc H_{\mc Z}}^{2}
\,d\p_{\bX}(\bx)
\le
R_n^{2}
\]
By Steps 2 and 3, almost surely the left-hand side is bounded by $R_n^2\to0$ for all $n\ge N_0(\omega)$, which is strong consistency in $L^2(\p_{\bX};\mc H_{\mc Z})$ in the sense of the definition preceding Theorem~\ref{thm:pointwise_error_bound_intro}.
\end{proof}

\section{Auxiliary Algorithms}
\subsection{Numerical Optimization and Implementation Details}\label{app:optimizer_details}
The reported implementation minimizes \eqref{eq:global_B_objective} using AdamW \citep{loshchilov2019decoupled} with automatic differentiation. The empirical average is computed either over the full conditioning set $\mc X_\star$ of size $M=N$ or over mini-batches of conditioning points. Optimization terminates at the prescribed iteration budget or when the current-batch gradient norm is below $10^{-8}$ at an iteration scheduled for full-objective evaluation. At these evaluation iterations, the objective is computed over the full conditioning set, and $\widehat{\mathbb B}^{\pi}$ denotes the iterate with the smallest such value. AdamW weight decay, when nonzero, is an algorithmic regularizer separate from $\lambda_B$ and is reported with the numerical settings.

\noindent The simulation and real-data analyses use the same estimator with different conditioning designs. In the trajectory-based simulation, $\mc X_\star$ is selected from observed state-action inputs and the optimizer uses mini-batches of conditioning points. State-action basis points are selected as $k$-means landmarks, and the basis size $L$ is distinct from the number of pooled transitions. A transition subsample may be used to construct the Bellman operators at lower computational cost; this choice does not determine the row dimension of $\mathbb B$. The simulation uses a Mat\'ern random-feature approximation, the coefficient-mass penalty, and no simplex or nonnegativity constraint on the fitted coefficients.

\noindent The Expedia analysis uses the same estimator. Its conditioning set contains multiple training inputs, including the evaluation input used for the displayed conditional summaries; the conditional embedding at that input is evaluated from the common fitted coefficient matrix rather than from a separate local fit. Target-to-behavior density ratios are estimated by uLSIF after replacing negative coefficient estimates by zero and calibrating their empirical denominator-sample mean before they enter $\boldsymbol\Phi_L$; no upper clipping is applied. The fitted ratios are finite on the observed sample, but this does not verify population boundedness. The probability proxies in the Expedia figures are recovered after fitting by matching the induced embedding over simplex weights on the return-grid atoms and projecting the discounted click-return coordinate onto the observed one-step count support. This recovery step is separate from the Bellman criterion in \eqref{eq:global_B_objective}.

\subsection{Held-out-Input Projected Bellman Residual}\label{app:heldout_residual_details}
Let $\mc X_{\mathrm{test}}=\{\bx_j^{\mathrm{test}}\}_{j=1}^{N_{\mathrm{test}}}$ be held-out state-action inputs. The fitted coefficient matrix, Gram matrices, reward summaries, and density-ratio estimate remain those obtained from the training transition sample; only the conditioning input is replaced by a test input. For each held-out input, compute
\begin{align*}
\bk_{L,j}^{\mathrm{test}}
&:=
\bk_L(\bx_j^{\mathrm{test}}),
&
\widetilde{\bk}_{N,j}^{\mathrm{test}}
&:=
( k_{\mc X}(\bx_1,\bx_j^{\mathrm{test}}),\ldots, k_{\mc X}(\bx_N,\bx_j^{\mathrm{test}}))^\top,
\\
\boldsymbol{\Gamma}_{j}^{\mathrm{test}}
&:=
(\mathbb K_{\mc X}+N\lambda_\Gamma\mathbb I_N)^{-1}
\widetilde{\bk}_{N,j}^{\mathrm{test}},
&
\boldsymbol{\Phi}_{L,j}^{\mathrm{test}}
&:=
\mathbb K_{L^+}\mathbb D_{\eta}\boldsymbol{\Gamma}_{j}^{\mathrm{test}}
\end{align*}
Let $\mathbb H_{j}^{\mathrm{test}}:=\widehat{\mathbb H}(\bx_j^{\mathrm{test}})$ and $\mathbb G_{j}^{\mathrm{test}}:=\widehat{\mathbb G}(\bx_j^{\mathrm{test}})$ be the reward-shifted matrices defined by \eqref{eq:h_ij} and \eqref{eq:g_ij}. For a candidate matrix $\mathbb B$, define $u_j^{\mathrm{test}}(\mathbb B):=\mathbb B^\top\bk_{L,j}^{\mathrm{test}}$ and $v_j^{\mathrm{test}}(\mathbb B):=\mathbb B^\top\boldsymbol{\Phi}_{L,j}^{\mathrm{test}}$. Substitution of $\widehat{\mathbb B}^{\pi}$ gives the quantities in \eqref{eq:generalization_metric}.

\noindent This diagnostic differs from $(u_j-v_j)^\top\mathbb K_{\mc Z}(u_j-v_j)$, which omits the reward shift and therefore is not the projected Bellman residual in \eqref{eq:gamma_with_W}. When $\mathbb H_{j}^{\mathrm{test}}$ and $\mathbb G_{j}^{\mathrm{test}}$ use the same plug-in reward embedding and the same random-feature map, each summand in \eqref{eq:generalization_metric} is a squared RKHS norm and is nonnegative in exact arithmetic. Small negative values can result from floating-point error or inconsistent matrix approximations and are reported without truncation.

\subsection[uLSIF density-ratio coefficients]{uLSIF Estimation of the Density-Ratio Coefficients}\label{app:uLSIF}
Let the random variable $\bX=(\bS^\top,\bA^\top)^\top$ denote the state-action input with realization $\bx=(\bs^\top,\ba^\top)^\top$. Let $P_{\mathrm{ref}}$ denote the common reference law of the states used to construct the uLSIF samples, and define the state-matched denominator and numerator laws
\[
Q_{\mathrm{den}}(d\bs,d\ba)=P_{\mathrm{ref}}(d\bs)\beta(d\ba\mid\bs),
\qquad
Q_{\mathrm{num}}(d\bs,d\ba)=P_{\mathrm{ref}}(d\bs)\pi(d\ba\mid\bs)
\]
Under overlap, their Radon-Nikodym derivative is
\[
\eta(\bs,\ba)
:=
\frac{dQ_{\mathrm{num}}}{dQ_{\mathrm{den}}}(\bs,\ba)
=
\frac{d\pi(\cdot\mid\bs)}{d\beta(\cdot\mid\bs)}(\ba)
=
\frac{\pi(\ba\mid\bs)}{\beta(\ba\mid\bs)}
\]
where the last equality uses dominated action laws. In KE-DRL, $P_{\mathrm{ref}}$ is the empirical law of the logged successor states: the denominator inputs are the observed successor pairs $(\bs_j',\ba_j')$, and the numerator inputs pair the same successor states with actions drawn from $\pi(\cdot\mid\bs_j')$. Thus uLSIF targets the successor-action ratio used in $\boldsymbol\Phi_L$, not a ratio of policy-specific state occupancy distributions. The ratio is estimated without separately estimating either input law. For the basic denominator-centered form, represent the ratio estimator in an RKHS with kernel $k$ as
\[
\widehat{\eta}(\bx)=\sum_{j=1}^{n_\beta}\alpha_j k(\bx_{\beta,j},\bx)
\]
where $\balpha=(\alpha_1,\ldots,\alpha_{n_\beta})^{\top}$ is estimated using denominator inputs $\{\bx_{\beta,j}\}_{j=1}^{n_\beta}$ and numerator inputs $\{\bx_{\pi,j}\}_{j=1}^{n_\pi}$. Define the empirical kernel matrices
\[
\mathbb K_{bb}=\big[k(\bx_{\beta,i},\bx_{\beta,j})\big]_{i,j=1}^{n_\beta}
\qquad
\mathbb K_{bp}=\big[k(\bx_{\beta,i},\bx_{\pi,j})\big]_{i,j}
\]
and minimize
\[
J(\balpha)=\tfrac12\balpha^{\top}\mathbb V\balpha-\mathbf v^{\top}\balpha
\]
with
\[
\mathbb V=\tfrac{1}{n_\beta}\mathbb K_{bb}\mathbb K_{bb}^{\top}
\qquad
\mathbf v=\tfrac{1}{n_\pi}\mathbb K_{bp}\mathbf 1
\]
Adding a Tikhonov regularization term $\tfrac{\lambda}{2}\|\balpha\|^2$ yields the closed-form normal equation
\[
(\mathbb V+\lambda\mathbb I)\balpha=\mathbf v
\]
which is solved numerically using a Cholesky decomposition for stability. Once $\balpha$ is obtained, the estimated ratio for any realization $\bx$ is
\[
\widehat{\eta}(\bx)=\bk_{b,\bx}^{\top}\balpha
\qquad
\bk_{b,\bx}=\big(k(\bx_{\beta,1},\bx),\ldots,k(\bx_{\beta,n_\beta},\bx)\big)^\top
\]
The kernel $k(\cdot,\cdot)$ employed here is a $\mbox{Mat\'{e}rn}$ kernel of smoothness parameter $\nu$ and length-scale $\ell$.

\begin{algorithm}[H]
\caption{uLSIF with $\mbox{Mat\'{e}rn}$ Kernel for Estimating $\boldsymbol{\alpha}$}
\label{alg:ulsif}
\begin{algorithmic}[1]
\State \textbf{Input:} Denominator inputs $\{\bx_{\beta,j}\}_{j=1}^{n_\beta}$, numerator inputs $\{\bx_{\pi,j}\}_{j=1}^{n_\pi}$ constructed from the same state reference law, kernel $k_\nu$, regularization $\lambda$
\State \textbf{Output:} Estimated coefficient vector $\widehat{\balpha}$ and ratio estimator $\widehat{\eta}(\bx)$
\State Compute $\mathbb K_{bb}=\big[k_\nu(\bx_{\beta,i},\bx_{\beta,j})\big]_{i,j=1}^{n_\beta}$ and $\mathbb K_{bp}=\big[k_\nu(\bx_{\beta,i},\bx_{\pi,j})\big]_{i,j}$
\State Form $\mathbb V=\tfrac{1}{n_\beta}\mathbb K_{bb}\mathbb K_{bb}^{\top}$ and $\mathbf v=\tfrac{1}{n_\pi}\mathbb K_{bp}\mathbf 1$
\State Solve $(\mathbb V+\lambda\mathbb I)\balpha=\mathbf v$ using Cholesky factorization
\State Evaluate $\widehat{\eta}(\bx)=\bk_{b,\bx}^{\top}\widehat{\balpha}$ with $\bk_{b,\bx}=\big(k_\nu(\bx_{\beta,1},\bx),\ldots,k_\nu(\bx_{\beta,n_\beta},\bx)\big)^\top$
\State \textbf{return} $\widehat{\balpha}$ and $\widehat{\eta}(\bx)$
\end{algorithmic}
\end{algorithm}

\noindent Algorithm~\ref{alg:ulsif} states the full denominator-centered, unconstrained form. More generally, for $b_\eta$ centers $\{\bx_{\eta,\ell}\}_{\ell=1}^{b_\eta}$ selected from either input sample, the same objective uses $\mathbb K_{\eta b}=[k(\bx_{\eta,\ell},\bx_{\beta,j})]_{\ell,j}$ and $\mathbb K_{\eta\pi}=[k(\bx_{\eta,\ell},\bx_{\pi,j})]_{\ell,j}$ in place of $\mathbb K_{bb}$ and $\mathbb K_{bp}$. The Expedia analysis uses $1500$ centers sampled from the numerator inputs and three target-action draws at each logged successor state. After solving the unconstrained uLSIF system, negative coefficient estimates are replaced by zero, $\widehat\alpha_\ell\leftarrow\max(\widehat\alpha_\ell,0)$. The resulting coefficient vector is multiplied by a common positive constant so that the empirical denominator-sample mean $n_\beta^{-1}\sum_{j=1}^{n_\beta}\widehat\eta(\bx_{\beta,j})$ equals one. Evaluated ratios are additionally truncated below at zero before entering $\mathbb D_\eta$, and no upper clipping is applied. These are finite-sample modifications of the estimator; the estimand remains $dQ_{\mathrm{num}}/dQ_{\mathrm{den}}$.

\subsection[Grid construction algorithm]{Return-grid construction for $\bZ^\pi$}\label{app:Grid-Z}
Given $N_{\bz}$ observed discounted return samples $\{\bar{\bz}_i\}_{i=1}^{N_{\bz}}\subset\R^d$, $k$-means partitions the samples into $K$ clusters and computes centroids $\{\bc_k\}_{k=1}^K$ by minimizing the within-cluster sum of squares. We use the centroid average $\bar{\bc}=\frac{1}{K}\sum_{k=1}^K\bc_k$ as a reference point for radial expansion. From the centroid set, we select $H$ centroids $\{\mathbf v_j\}_{j=1}^H$ with extreme projections along a fixed collection of directions consisting of the coordinate directions, their negatives, and randomly sampled unit directions. This directional construction avoids an exact convex-hull computation. Each selected centroid $\mathbf v_j$ is expanded radially from $\bar{\bc}$ by a factor $c_{\mathrm{exp}}>1$, yielding
\[
\mathbf v_j'=\bar{\bc}+c_{\mathrm{exp}}(\mathbf v_j-\bar{\bc})
\]
so that the grid extends beyond the observed extremes in the selected directions. The expanded points \emph{replace} the corresponding centroids in the atom set, so the grid is
\[
\mc Z_{\mathrm{grid}}=\left(\{\bc_1,\dots,\bc_K\}\setminus\{\mathbf v_1,\dots,\mathbf v_H\}\right)\cup\{\mathbf v'_1,\dots,\mathbf v'_H\}
\]
containing exactly $K$ atoms: the $K-H$ unselected centroids are retained, and the $H$ selected centroids are moved outward. Thus the return-grid size is $m=K$, the value reported in Sections~\ref{sec:sim} and~\ref{sec:realdata}. The construction retains atoms in regions represented by the observed returns while extending the grid in selected boundary directions. The grid is constructed once for each fitted analysis from the discounted-return samples supplied to the estimator, and the same $m$ atoms parameterize the estimated embedding \eqref{eq:mu_Z} at every state-action input. In the simulation of Section~\ref{sec:sim}, each behavior-target scenario is fitted to a separate offline sample, so the resulting grids may differ across scenarios. In the Expedia analysis of Section~\ref{sec:realdata}, both policies are evaluated using a single data-adaptive grid constructed from the logged data, as recorded in Supplementary Material~\ref{app:Exp_data}.

\noindent The grid atoms and the locations used to evaluate the Bellman target are distinct. The fitted class \eqref{eq:mu_Z} places coefficients on the kernel sections $k_{\mc Z}(\bz_i,\cdot)$ at the $m$ atoms, whereas the Bellman operators evaluate the return kernel at the shifted locations $\br+\gamma\bz_i$ induced by the affine Bellman map \eqref{eq:pushforward_map}, as in the entries of $\widehat{\mathbb H}(\bx)$ and $\widehat{\mathbb G}(\bx)$ in \eqref{eq:h_ij}-\eqref{eq:g_ij}. The shifted points need not be grid atoms because the Mat\'ern kernel is defined on all of $\R^d$. The atomic pushforward transports each atom $\bz_i$ to $\br+\gamma\bz_i$ without projecting the transported section back onto the grid. When random Fourier features are used, all evaluations at these shifted locations use the common feature approximation described below. Accordingly, grid adequacy concerns approximation quality rather than whether shifted points are themselves grid atoms. The attainable projected Bellman residual \eqref{eq:gamma_def} depends on how well the span of $k_{\mc Z}(\bz_1,\cdot),\ldots,k_{\mc Z}(\bz_m,\cdot)$ represents the embeddings on both sides of the Bellman recursion, including the reward-shifted sections evaluated on the empirical Bellman image $\{\br_j+\gamma\bz_i\}$.

\noindent When every reward coordinate has bounded support, an exactly Bellman-invariant choice of the return region is available: taking $\mc Z$ to be the coordinatewise box obtained by scaling each reward-support interval by $(1-\gamma)^{-1}$ gives the inclusion
\[
\mc R+\gamma\mc Z\subseteq\mc Z
\]
so every shifted point $\br+\gamma\bz$ with $\br\in\mc R$ and $\bz\in\mc Z$ remains in $\mc Z$, and a grid filling this box covers the Bellman image up to its mesh. In the simulation of Section~\ref{sec:sim}, the Gaussian rewards have unbounded support, so no compact region satisfies this inclusion exactly. Finite-grid coverage therefore requires empirical assessment. The radial expansion by $c_{\mathrm{exp}}$ moves selected atoms beyond the observed sampled-return extremes so that the shifted locations $\br_j+\gamma\bz_i$ encountered during estimation are more likely to lie within or near the region represented by the grid.

\noindent A direct empirical assessment can be based on the normalized residual obtained by projecting each shifted section $k_{\mc Z}(\br_j+\gamma\bz_i,\cdot)$ onto the span of the grid sections. This quantity is computed from the return-grid Gram matrix $\mathbb K_{\mc Z}$ and the kernel evaluations between the shifted point and the grid atoms, and is normalized by the RKHS norm of the shifted section. Values near zero indicate that the section is well represented on the grid. Its median and upper percentiles over sampled pairs $(\br_j,\bz_i)$, together with the fraction of shifted points in the convex hull of the expanded grid, would summarize coverage at the reported resolution. Such diagnostics can be evaluated at the locations used by the empirical operators \eqref{eq:h_ij}-\eqref{eq:g_ij}. They are not reported here, so the numerical conclusions remain conditional on the selected grid.
\section{Simulation Details}

\JASASimulationRepositoryText\ We use a trajectory-based offline design with $n=300$ trajectories of length $T=50$ and discard a burn-in of $100$ behavior-policy steps. Each offline replicate therefore contains $N=n(T-1)=300\times 49=14{,}700$ pooled transitions of the form $(\bs_j,\ba_j,\br_j,\bs_j',\ba_j')$. The state, action, and reward dimensions are $p=5$, $q=1$, and $d=3$; the discount factor is $\gamma=0.7$; and the return grid contains $m=100$ points.

The procedure separates the pooled transition sample from the parameterization of the estimated mean embedding. The Bellman equations are evaluated on the pooled transitions and conditioning set, whereas the embedding map is represented on a smaller state-action basis
\[
\mc C_L=\{\underline \bx_\ell\}_{\ell=1}^L
\qquad
\underline \bx_\ell=(\bar\bs_\ell,\bar\ba_\ell)
\]
selected by $k$-means from the state-action inputs. For a query $\bx=(\bs,\ba)$ write
\[
\bk_L(\bx)=\left( k_{\mc X}(\underline \bx_1,\bx),\ldots, k_{\mc X}(\underline \bx_L,\bx)\right)^\top
\]
The numerically fitted policy-specific coefficient matrix $\widehat{\mathbb B}^{\pi}\in\mathbb R^{L\times m}$ yields
\[
\widehat{\bso}(\bs,\ba)=(\widehat{\mathbb B}^{\pi})^{\top}\bk_L(\bs,\ba)
\]
so that the estimated conditional return embedding is represented by coefficients on the $m$ return-grid points. Relative to the full $N\times m$ representation, $L$ controls the number of estimated state-action basis coefficients, whereas $m$ controls the return-grid resolution.

For each evaluation point $\bx=(\bs,\ba)$ the conditional embedding weights use the kernel-ridge vector
\[
\boldsymbol\Gamma(\bx)=\left(\mathbb K_{\mc X}+N\lambda_\Gamma\mathbb I_N\right)^{-1}\widetilde\bk_N(\bx)
\]
where $\mathbb K_{\mc X}$ is the state-action Gram matrix over the pooled transitions and $\widetilde\bk_N(\bx)$ evaluates the state-action kernel between all observed current inputs and $\bx$. For uLSIF, the logged successor inputs form the denominator sample, and target-policy actions drawn at the same successor states form the numerator sample. The resulting density-ratio weights $\widehat\eta_j$ are evaluated at the next state-action inputs, giving the projected successor feature
\[
\boldsymbol\Phi_L(\bx)=\mathbb K_{L^+}\mathbb D_\eta\boldsymbol\Gamma(\bx)
\]
with
\[
[\mathbb K_{L^+}]_{\ell j}= k_{\mc X}(\underline \bx_\ell,\bx_j')
\]
and $\mathbb D_\eta$ the diagonal matrix of estimated density-ratio weights.

The return-space Bellman operators are built on the grid $\{\bz_i\}_{i=1}^m$. For fixed indices $(i,j)$
\[
\big[\bk_{\bR}^{(ij)}\big]_r=k_{\mc Z}(\br_r,\bz_i-\gamma\bz_j)
\]
\[
\big[\mathbb K_{\gamma\bz_i+\bR,\gamma\bz_j+\bR}\big]_{r\ell}=k_{\mc Z}(\gamma\bz_i+\br_r,\gamma\bz_j+\br_\ell)
\]
For the reported simulation settings, these operators are not formed by the exact $O(m^2N^2)$ construction; instead, the Mat\'ern return kernel is approximated by random Fourier features
\[
k_{\mc Z}(\bz,\bz')\approx\varphi_Q(\bz)^\top\varphi_Q(\bz')
\]
with the pair-specific value of $Q$ reported in Table~\ref{tab:sim_pair_params}. The same finite feature map is used for $\mathbb K_{\mc Z}$, $\mathbb H_{\bx}$, and $\mathbb G_{\bx}$, so all three matrices correspond to a common approximate return kernel. Table~\ref{tab:sum_eq_updated} summarizes the empirical quantities in this finite-dimensional approximation.

\begin{table}[H]
\centering
\caption{Empirical quantities and dimensions for the finite-dimensional $L\times m$ KE-DRL approximation.}
\label{tab:sum_eq_updated}
\small
\begin{tabular}{@{}lll@{}}
\toprule\toprule
\textbf{Object} & \textbf{Size} & \textbf{Role} \\
\midrule
$\widetilde\bk_N(\bs,\ba)$ & $N\times 1$ & Kernel vector from all pooled current inputs to $(\bs,\ba)$ \\
$\mathbb K_{\mc X}$ & $N\times N$ & State-action Gram matrix for $\boldsymbol\Gamma(\bx)$ \\
$\bk_L(\bs,\ba)$ & $L\times 1$ & Conditioning-basis feature vector for embedding evaluation \\
$\mathbb K_L$ & $L\times L$ & Gram matrix over selected basis points $\mc C_L$ \\
$\mathbb K_{L^+}$ & $L\times N$ & Kernel matrix from basis points to successor inputs \\
$\boldsymbol\Gamma(\bx)$ & $N\times 1$ & Kernel-ridge conditional embedding weights \\
$\boldsymbol\Phi_L(\bs,\ba)$ & $L\times 1$ & Target-policy successor feature after density-ratio correction \\
$\mathbb K_{\mc Z}$ & $m\times m$ & Return-grid Gram matrix, approximated by RFF in the reported simulations \\
$\widehat{\mathbb H}(\bx)$ & $m\times m$ & Linear Bellman return operator for one conditioning point \\
$\widehat{\mathbb G}(\bx)$ & $m\times m$ & Quadratic Bellman return operator for one conditioning point \\
$\boldsymbol\alpha$ & $b_\eta\times 1$ & uLSIF coefficient vector for $b_\eta$ kernel centers \\
$\widehat{\mathbb B}^{\pi}$ & $L\times m$ & Numerically fitted mean-embedding coefficient matrix \\
\bottomrule\bottomrule
\end{tabular}
\end{table}

The state-action basis dimension $L$, random-feature dimension $Q$, and tuning parameters vary by policy pair, as reported in Table~\ref{tab:sim_pair_params}. These numerical settings are provided for reproducibility, but the study does not establish a common selection rule or assess sensitivity to them. The numerical comparisons are therefore descriptive and conditional on the listed settings.

\begin{table}[H]
\centering
\caption{Pair-specific estimation parameters. Here $L$ is the number of selected state-action basis points in the fitted coefficient matrix $\widehat{\mathbb B}^{\pi}\in\mathbb R^{L\times m}$, and $Q$ is the number of random Fourier features used to approximate the return-space Mat\'ern operators.}
\label{tab:sim_pair_params}
\small
\begin{tabular}{@{}llcccccccc@{}}
\toprule\toprule
\textbf{Behavior} & \textbf{Target} & $L$ & $Q$ & $\lambda_\Gamma$ & $\lambda_B$ & $\ell_X$ & $\sigma_X$ & \textbf{Iterations} & \textbf{Batch size} \\ \midrule
Gaussian & Uniform  & 320 & 1024 & 0.0020 & 0.004 & 1.80 & 1.00 & 7000 & 256 \\
Gaussian & Logistic & 320 & 1024 & 0.0025 & 0.006 & 1.65 & 0.95 & 7000 & 256 \\
Logistic & Uniform  & 260 & 768  & 0.0040 & 0.014 & 1.35 & 0.85 & 6000 & 200 \\
Logistic & Gaussian & 280 & 768  & 0.0040 & 0.016 & 1.50 & 0.90 & 6500 & 200 \\
Uniform  & Gaussian & 250 & 768  & 0.0030 & 0.012 & 1.25 & 0.85 & 5500 & 150 \\
Uniform  & Logistic & 220 & 640  & 0.0040 & 0.018 & 1.20 & 0.80 & 5000 & 150 \\
\bottomrule\bottomrule
\end{tabular}
\end{table}

For the six rows of Table~\ref{tab:sim_pair_params}, the coefficient-mass penalties $\lambda_{\mathrm{mass}}$ are $4.0$, $3.5$, $2.0$, $2.5$, $1.5$, and $1.4$, respectively. The initial learning rate is $2\times10^{-4}$ for the two Gaussian-behavior rows, $2.5\times10^{-4}$ for the two Logistic-behavior rows, and $3\times10^{-4}$ for the two Uniform-behavior rows. AdamW weight decay is $5\times10^{-5}$ except for the Uniform-to-Logistic pair, where it is $8\times10^{-5}$. The target mass is $c_{\mathrm{mass}}=1$ for every pair, and weight decay is separate from $\lambda_B$.

\subsection{Simulation Setting}\label{app:sim_setting}

The state-transition and reward models are linear Gaussian
\[
\bs'=\mathbf b_s+\mathbb W_s^\top[\bs,\ba]+\boldsymbol{\varepsilon}_s
\qquad
\boldsymbol{\varepsilon}_s\sim\mathcal N(0,\Sigma_s)
\]
\[
\br=\mathbf b_r+\mathbb W_r^\top[\bs,\ba]+\boldsymbol{\varepsilon}_r
\qquad
\boldsymbol{\varepsilon}_r\sim\mathcal N(0,\Sigma_r)
\]
The transition and reward innovations are generated independently across the two equations, trajectories, and time points.
The transition matrix is
\[
\mathbb{W}_s =
\begin{bmatrix}
0.4 & -0.2 & 0.1 & 0.5 & 0.3 & -0.1 \\
0.03 & 0.3 & -0.2 & 0.15 & 0.5 & 0.1 \\
0.15 & -0.05 & 0.2 & 0.6 & 0.5 & -0.2 \\
0.2 & 0.05 & -0.1 & 0.3 & -0.15 & 0.2 \\
0.1 & -0.3 & 0.25 & -0.2 & 0.4 & 0.15
\end{bmatrix}
\qquad
\mathbf b_s =
\begin{bmatrix}
0 \\ 0 \\ 0 \\ 0 \\ 0
\end{bmatrix}
\]
and the reward matrix is
\[
\mathbb{W}_r =
\begin{bmatrix}
0.02 & 0.1 & -0.5 & 0.3 & -0.1 & 0.2 \\
0.1 & -0.3 & 0.2 & 0.25 & -0.2 & 0.4 \\
0.15 & 0.1 & -0.1 & 0.35 & 0.1 & -0.05
\end{bmatrix}
\qquad
\mathbf b_r =
\begin{bmatrix}
0 \\ 0 \\ 0
\end{bmatrix}
\]
The covariance matrices are
\[
\Sigma_s=
\begin{bmatrix}
0.1 & 0.05 & 0.02 & 0.01 & 0.03 \\
0.05 & 0.2 & 0.03 & 0.02 & 0.04 \\
0.02 & 0.03 & 0.3 & 0.05 & 0.01 \\
0.01 & 0.02 & 0.05 & 0.25 & 0.02 \\
0.03 & 0.04 & 0.01 & 0.02 & 0.35
\end{bmatrix}
\qquad
\Sigma_r=
\begin{bmatrix}
0.2 & 0.01 & 0.03 \\
0.01 & 0.25 & 0.02 \\
0.03 & 0.02 & 0.3
\end{bmatrix}
\]

\subsection*{Policy specifications}
Actions are generated from state-dependent stochastic policies on their native real-valued scale; no sigmoid transformation is applied. For a state $\bs$, the Gaussian policy uses
\[
\mu(\bs)=\theta_\mu^\top \bs+\epsilon_\mu
\qquad
\sigma(\bs)=\exp(\theta_\sigma^\top \bs+\epsilon_\sigma)
\qquad
a\sim \mathcal N(\mu(\bs),\sigma^2(\bs))
\]
the Uniform policy uses
\[
\ell(\bs)=\theta_L^\top\bs+\epsilon_L
\qquad
u(\bs)=\theta_U^\top\bs+\epsilon_U
\qquad
a\sim \mathrm{Unif}(\ell(\bs),u(\bs))
\]
with the upper endpoint shifted upward when needed to preserve $u(\bs)>\ell(\bs)$, and the Logistic policy uses
\[
loc(\bs)=\theta_{\ell}^\top\bs+\epsilon_{\ell}
\qquad
scale(\bs)=\exp(\theta_{sc}^\top\bs+\epsilon_{sc})
\]
\[
a=loc(\bs)+scale(\bs)\log\!\left(\frac{u}{1-u}\right)
\qquad u\sim\mathrm{Unif}(0,1)
\]
\noindent The pair-specific policy parameters are chosen to maintain adequate overlap between behavior and target policies while still inducing nontrivial off-policy shifts, and are listed in Table~\ref{tab:policy_params}.

\begin{table}[H]
\centering
\caption{Policy specification parameters used in the six off-policy simulation scenarios. The scenario column is ordered as behavior policy $\rightarrow$ target policy, and the role column identifies whether the row gives the behavior or target policy used in that scenario. Gaussian policies use $a\mid s\sim\mathcal N(\theta_\mu^\top s+\epsilon_\mu,\exp\{2(\theta_\sigma^\top s+\epsilon_\sigma)\})$. Logistic policies use $a=loc(s)+scale(s)\log\{u/(1-u)\}$, where $u\sim\mathrm{Unif}(0,1)$, $loc(s)=\theta_\ell^\top s+\epsilon_\ell$, and $scale(s)=\exp(\theta_{sc}^\top s+\epsilon_{sc})$. Uniform policies use $a\mid s\sim\mathrm{Unif}(\theta_L^\top s+\epsilon_L,\theta_U^\top s+\epsilon_U)$. In all reported scenarios, $\theta_\sigma=\theta_{sc}=\mathbf 0$, so Gaussian and Logistic scale parameters are state independent.}
\label{tab:policy_params}
\footnotesize
\setlength{\tabcolsep}{2pt}
\renewcommand{\arraystretch}{1}
\begin{tabular}{@{}p{0.17\textwidth}p{0.08\textwidth}p{0.09\textwidth}p{0.32\textwidth}p{0.12\textwidth}p{0.12\textwidth}@{}}
\toprule\toprule
\textbf{Scenario} & \textbf{Role} & \textbf{Policy} & \textbf{Linear coefficient vector} & \textbf{\shortstack{Location\\lower offset}} & \textbf{\shortstack{Scale\\upper offset}} \\
\midrule
Gaussian $\rightarrow$ Uniform
& Behavior
& Gaussian
& $\theta_\mu=(0.12,-0.08,0.12,-0.35,0.04)$, $\theta_\sigma=\mathbf 0$
& $\epsilon_\mu=0.02$
& $\epsilon_\sigma=-1.60$ \\

Gaussian $\rightarrow$ Uniform
& Target
& Uniform
& $\theta_L=\theta_U=(0.12,-0.08,0.12,-0.35,0.04)$
& $\epsilon_L=-0.16$
& $\epsilon_U=0.20$ \\
\midrule

Gaussian $\rightarrow$ Logistic
& Behavior
& Gaussian
& $\theta_\mu=(0.12,-0.08,0.12,-0.35,0.04)$, $\theta_\sigma=\mathbf 0$
& $\epsilon_\mu=0.02$
& $\epsilon_\sigma=-1.60$ \\

Gaussian $\rightarrow$ Logistic
& Target
& Logistic
& $\theta_\ell=(0.12,-0.08,0.12,-0.35,0.04)$, $\theta_{sc}=\mathbf 0$
& $\epsilon_\ell=0.02$
& $\epsilon_{sc}=-2.80$ \\
\midrule

Logistic $\rightarrow$ Uniform
& Behavior
& Logistic
& $\theta_\ell=(0.05,-0.12,0.18,-0.40,-0.03)$, $\theta_{sc}=\mathbf 0$
& $\epsilon_\ell=0.03$
& $\epsilon_{sc}=-1.80$ \\

Logistic $\rightarrow$ Uniform
& Target
& Uniform
& $\theta_L=\theta_U=(0.05,-0.12,0.18,-0.40,-0.03)$
& $\epsilon_L=-0.15$
& $\epsilon_U=0.21$ \\
\midrule

Logistic $\rightarrow$ Gaussian
& Behavior
& Logistic
& $\theta_\ell=(0.05,-0.12,0.18,-0.40,-0.03)$, $\theta_{sc}=\mathbf 0$
& $\epsilon_\ell=0.03$
& $\epsilon_{sc}=-1.80$ \\

Logistic $\rightarrow$ Gaussian
& Target
& Gaussian
& $\theta_\mu=(0.05,-0.12,0.18,-0.40,-0.03)$, $\theta_\sigma=\mathbf 0$
& $\epsilon_\mu=0.03$
& $\epsilon_\sigma=-2.70$ \\
\midrule

Uniform $\rightarrow$ Gaussian
& Behavior
& Uniform
& $\theta_L=\theta_U=(0.10,-0.10,0.15,-0.45,0.00)$
& $\epsilon_L=-0.725$
& $\epsilon_U=0.775$ \\

Uniform $\rightarrow$ Gaussian
& Target
& Gaussian
& $\theta_\mu=(0.10,-0.10,0.15,-0.45,0.00)$, $\theta_\sigma=\mathbf 0$
& $\epsilon_\mu=0.025$
& $\epsilon_\sigma=-2.60$ \\
\midrule

Uniform $\rightarrow$ Logistic
& Behavior
& Uniform
& $\theta_L=\theta_U=(0.08,-0.12,0.16,-0.42,-0.02)$
& $\epsilon_L=-0.72$
& $\epsilon_U=0.78$ \\

Uniform $\rightarrow$ Logistic
& Target
& Logistic
& $\theta_\ell=(0.08,-0.12,0.16,-0.42,-0.02)$, $\theta_{sc}=\mathbf 0$
& $\epsilon_\ell=0.03$
& $\epsilon_{sc}=-2.80$ \\
\bottomrule\bottomrule
\end{tabular}
\end{table}

\begin{table}[H]
\centering
\caption{Shared parameters for the reported simulation study. These settings apply to all six behavior-target policy pairs unless a pair-specific value is listed separately.}
\label{tab:sim_shared_params}
\small
\begin{tabular}{@{}ll@{}}
\toprule\toprule
\textbf{Parameter} & \textbf{Value} \\ \midrule
Offline trajectories & $n=300$ \\
Recorded time points per trajectory & $T=50$ \\
Pre-recording burn-in & $100$ behavior-policy steps, discarded before analysis \\
Pooled transitions & $N=n(T-1)=14{,}700$ \\
State dimension & $p=5$ \\
Action dimension & $q=1$ \\
Reward dimension & $d=3$ \\
Discount factor & $\gamma=0.7$ \\
Return-grid size & $m=100$ \\
Independent offline-data replicates & $100$ \\
Benchmark target points & $10$ \\
Monte Carlo benchmark trajectories & $10{,}000$ \\
Monte Carlo benchmark horizon & $T_{\mathrm{MC}}=500$ \\
Training conditioning set & all $14{,}700$ pooled current state-action inputs \\
Return kernel & Mat\'ern, $\nu=5.5$ \\
Operator construction & Mat\'ern random Fourier features \\
State-action basis size & pair-specific $L$; see Table \ref{tab:sim_pair_params} \\
\bottomrule\bottomrule
\end{tabular}
\end{table}

\begin{table}[H]
\centering
\caption{Benchmark evaluation points for the Uniform behavior policy and Gaussian target policy. The first four points are fixed by the simulation specification; the remaining six are target-policy draws satisfying the prespecified support criterion. Values are rounded to three decimals.}
\label{tab:test_target_points_UG}
\small
\begin{tabular}{@{}c|rrrrr|r@{}}
\toprule\toprule
\textbf{Point} & $s_1$ & $s_2$ & $s_3$ & $s_4$ & $s_5$ & $a$ \\ \midrule
1  &  0.600 & -0.800 &  0.700 & -0.500 &  0.900 &  0.500 \\
2  &  0.000 &  0.000 &  0.000 &  0.000 &  0.000 &  0.025 \\
3  &  0.200 & -0.200 &  0.100 & -0.100 &  0.300 &  0.125 \\
4  & -0.200 & -0.100 &  0.100 & -0.200 &  0.200 &  0.120 \\
5  & -0.721 & -0.913 & -0.086 &  0.336 & -0.259 & -0.081 \\
6  & -1.163 & -0.894 & -1.316 &  0.837 & -0.713 & -0.676 \\
7  &  0.882 &  1.198 &  1.529 & -0.260 & -0.374 &  0.373 \\
8  & -0.586 & -0.498 &  0.889 & -0.430 &  0.146 &  0.354 \\
9  & -0.499 &  0.292 & -0.105 &  0.418 & -0.157 & -0.292 \\
10 & -0.364 & -0.034 & -0.313 & -0.376 &  1.669 &  0.153 \\
\bottomrule\bottomrule
\end{tabular}
\end{table}

\subsection{Additional Simulation Results}\label{app:sim_results}
Figures~\ref{fig:LG_sum}-\ref{fig:LU_sum} extend the pointwise benchmark comparison in the main text to the remaining behavior-target pairings. In these scenarios, the plotted pointwise discrepancies between the KE-DRL estimates and Monte Carlo reference embeddings are comparable to those reported in the main text, conditional on the pair-specific numerical settings. This descriptive comparison does not assess tuning robustness or comparative performance against other estimators.

\begin{figure}
    \centering
    \includegraphics[width=0.96\linewidth]{plots/simulation_res/mu_summary_benchmarks_LG_3x2.png}
\caption{\setstretch{0.8}Logistic behavior policy and Gaussian target policy. Pointwise evaluations of the estimated mean-embedding functions are compared with the corresponding Monte Carlo benchmark evaluations over the plotted evaluation grids at the benchmark target points. Shared simulation settings are given in Table \ref{tab:sim_shared_params}; pair-specific estimation settings are given in Table \ref{tab:sim_pair_params}.}    \label{fig:LG_sum}
\end{figure}

\begin{figure}
    \centering
    \includegraphics[width=0.96\linewidth]{plots/simulation_res/mu_summary_benchmarks_GU_3x2.png}
\caption{\setstretch{0.8}Gaussian behavior policy and Uniform target policy. Pointwise evaluations of the estimated mean-embedding functions are compared with the corresponding Monte Carlo benchmark evaluations over the plotted evaluation grids at the benchmark target points. Shared simulation settings are given in Table \ref{tab:sim_shared_params}; pair-specific estimation settings are given in Table \ref{tab:sim_pair_params}.}    \label{fig:GU_sum}
\end{figure}

\begin{figure}
    \centering
    \includegraphics[width=0.96\linewidth]{plots/simulation_res/mu_summary_benchmarks_GL_3x2.png}
\caption{\setstretch{0.8}Gaussian behavior policy and Logistic target policy. Pointwise evaluations of the estimated mean-embedding functions are compared with the corresponding Monte Carlo benchmark evaluations over the plotted evaluation grids at the benchmark target points. Shared simulation settings are given in Table \ref{tab:sim_shared_params}; pair-specific estimation settings are given in Table \ref{tab:sim_pair_params}.}    \label{fig:GL_sum}
\end{figure}

\begin{figure}
    \centering
    \includegraphics[width=0.96\linewidth]{plots/simulation_res/mu_summary_benchmarks_UL_3x2.png}
\caption{\setstretch{0.8}Uniform behavior policy and Logistic target policy. Pointwise evaluations of the estimated mean-embedding functions are compared with the corresponding Monte Carlo benchmark evaluations over the plotted evaluation grids at the benchmark target points. Shared simulation settings are given in Table \ref{tab:sim_shared_params}; pair-specific estimation settings are given in Table \ref{tab:sim_pair_params}.}    \label{fig:UL_sum}
\end{figure}

\begin{figure}
    \centering
    \includegraphics[width=0.96\linewidth]{plots/simulation_res/mu_summary_benchmarks_LU_3x2.png}
\caption{\setstretch{0.8}Logistic behavior policy and Uniform target policy. Pointwise evaluations of the estimated mean-embedding functions are compared with the corresponding Monte Carlo benchmark evaluations over the plotted evaluation grids at the benchmark target points. Shared simulation settings are given in Table \ref{tab:sim_shared_params}; pair-specific estimation settings are given in Table \ref{tab:sim_pair_params}.}    \label{fig:LU_sum}
\end{figure}

\newpage

\subsection{Pointwise Error Aggregation and Repeated Monte Carlo Evaluation}\label{app:mc-repeat-target}

\noindent We first record the aggregation used for Table~\ref{tab:metric_summary_across_points}. For evaluation point $j=1,\ldots,10$ and offline-data replicate $r=1,\ldots,100$, let $\mathcal G_j=\{\mathbf g_{j1},\ldots,\mathbf g_{jG_j}\}$ denote the common grid of size $G_j$ used to evaluate the fitted embedding and its Monte Carlo reference, and define
\[
e_{jru}
:=
\widehat\mu_{jr}(\mathbf g_{ju})-\mu_j^{\mathrm{MC}}(\mathbf g_{ju}),
\qquad u=1,\ldots,G_j
\]
The replicate-specific summaries are
\[
\operatorname{Bias}_{jr}=\frac{1}{G_j}\sum_{u=1}^{G_j}e_{jru},
\qquad
\operatorname{RMSE}_{jr}=\left\{\frac{1}{G_j}\sum_{u=1}^{G_j}e_{jru}^2\right\}^{1/2},
\qquad
\operatorname{MAE}_{jr}=\frac{1}{G_j}\sum_{u=1}^{G_j}|e_{jru}|
\]
For each evaluation point, these summaries are averaged over the $100$ offline-data replicates. The entries of Table~\ref{tab:metric_summary_across_points} in the main text then average the point-specific summaries over the $10$ evaluation points, with parentheses giving their standard deviations across points.

\noindent Figure~\ref{fig:MC_box_plots} reports a repeated Monte Carlo reference diagnostic for the evaluation-point-specific mean embeddings used in the simulation study. Rather than comparing the estimate with a single Monte Carlo reference, we hold the estimated KE-DRL embedding fixed and repeatedly generate independent target-policy return samples from the same benchmark state-action input. Each sample defines a new Monte Carlo reference embedding.

\noindent For benchmark evaluation point $j$ with initial state-action pair $(\bs_j^\star,\ba_j^\star)$, the fitted procedure produces
\[
\widehat \mu_j(\bz)=\sum_{\ell=1}^{m_j}\widehat \omega_{j\ell}\,k_{\mc Z}(\bz,\bz_{j\ell})
\]
\noindent where $\{\bz_{j\ell}\}_{\ell=1}^{m_j}$ are the return-grid atoms and $k_{\mc Z}$ is the return-space kernel. For Monte Carlo repeat $b=1,\ldots,B$, let $\bZ_{j1}^{(b)},\ldots,\bZ_{j,n_{\mathrm{MC}}}^{(b)}$ denote independent simulated target-policy returns from $(\bs_j^\star,\ba_j^\star)$. The Monte Carlo reference embedding for repeat $b$ is
\[
\mu_j^{(b)}(\bz)=\frac{1}{n_{\mathrm{MC}}}\sum_{v=1}^{n_{\mathrm{MC}}}k_{\mc Z}(\bz,\bZ_{jv}^{(b)})
\]
The estimated and reference embeddings are compared on a common grid $\mathcal G_j=\{\mathbf g_{j1},\ldots,\mathbf g_{jG_j}\}$ with pointwise error
\[
e_{ju}^{(b)}=\widehat \mu_j(\mathbf g_{ju})-\mu_j^{(b)}(\mathbf g_{ju})
\]
and the plotted summaries are
\[
\mathrm{RMSE}_j^{(b)}=\left\{\frac{1}{G_j}\sum_{u=1}^{G_j}\left(e_{ju}^{(b)}\right)^2\right\}^{1/2}
\qquad
\mathrm{MAE}_j^{(b)}=\frac{1}{G_j}\sum_{u=1}^{G_j}\left|e_{ju}^{(b)}\right|
\qquad
\mathrm{Bias}_j^{(b)}=\frac{1}{G_j}\sum_{u=1}^{G_j}e_{ju}^{(b)}
\]
\noindent The bias panel reports the signed average discrepancy between the estimated and repeated Monte Carlo reference embeddings over the common grid. These summaries are computed from pointwise function evaluations and are not RKHS-norm or MMD errors. Positive values indicate that the estimated embedding is larger on average over this grid, whereas negative values indicate that it is smaller on average. Because the estimated embedding is held fixed across repetitions, variation within each boxplot reflects Monte Carlo variability in the reference embedding. Differences in boxplot centers across evaluation points reflect the combined effect of point-specific estimation error and the finite Monte Carlo reference. Wider boxplots indicate greater Monte Carlo variability at an evaluation point, but do not by themselves identify limited overlap or weak information in the offline data. In the displayed experiments, agreement with the Monte Carlo reference is stronger at some initial conditions than at others.

\begin{figure}[p]
\centering
\captionsetup{font=footnotesize}
\captionsetup[subfigure]{font=footnotesize,skip=3pt}
\setlength{\tabcolsep}{2pt}
\renewcommand{\arraystretch}{1.15}
\begin{tabular}{@{}cc@{}}
\subcaptionbox{Behavior: Gaussian; Target: Logistic\label{fig:mc_GL}}[0.49\linewidth]{%
\includegraphics[width=0.96\linewidth]{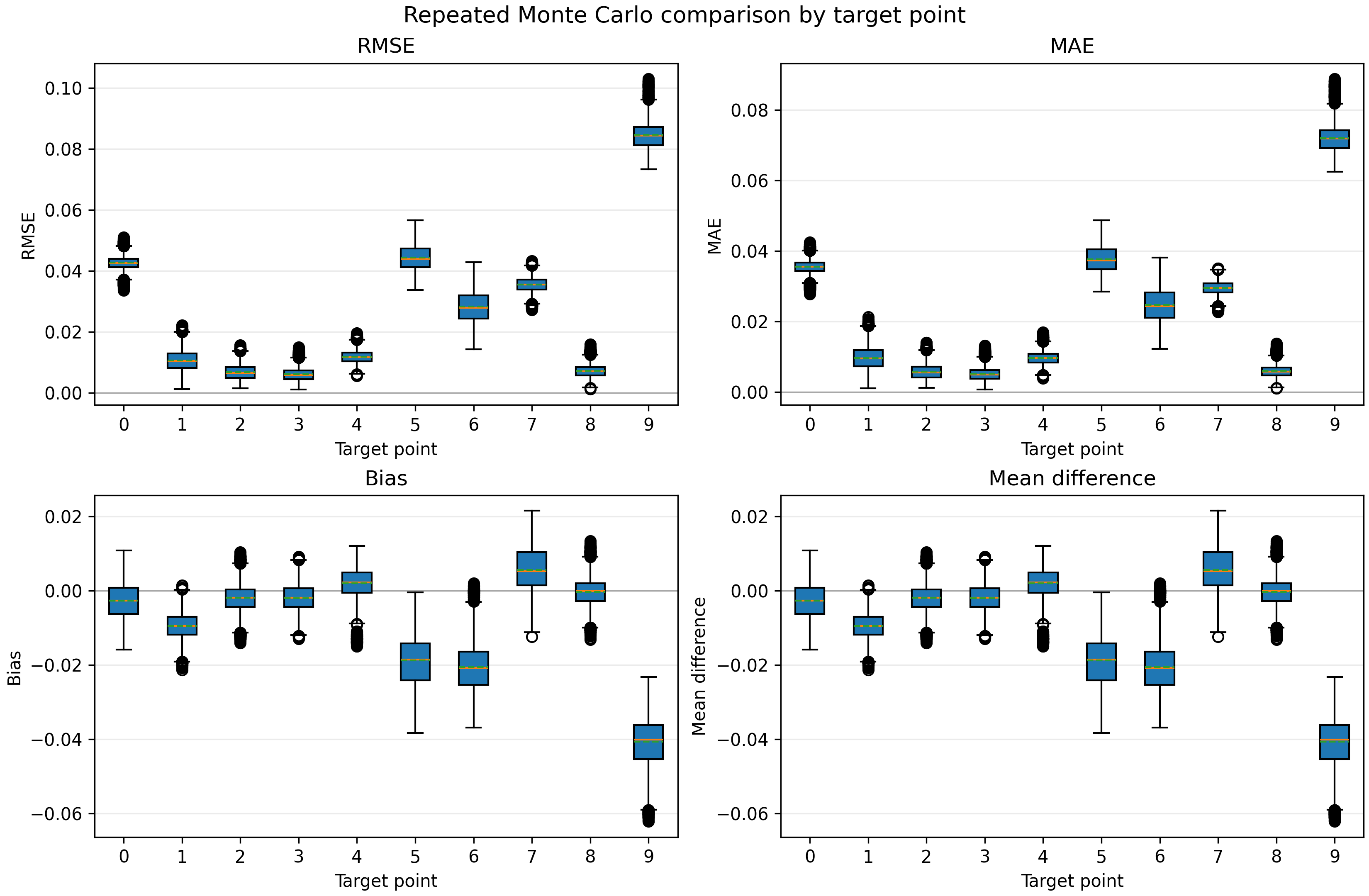}}
&
\subcaptionbox{Behavior: Gaussian; Target: Uniform\label{fig:mc_GU}}[0.49\linewidth]{%
\includegraphics[width=0.96\linewidth]{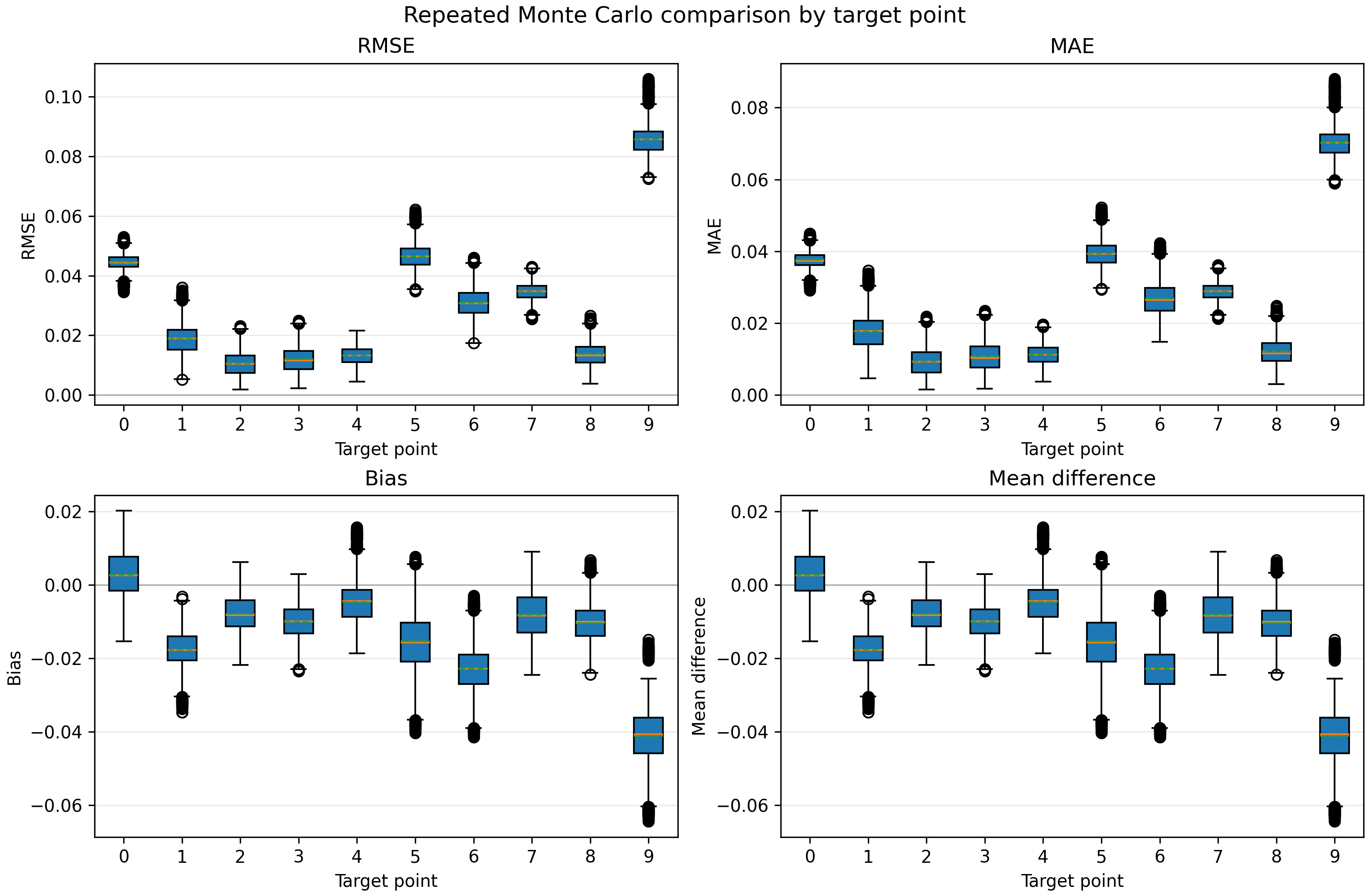}}
\\[0.25em]
\subcaptionbox{Behavior: Logistic; Target: Gaussian\label{fig:mc_LG}}[0.49\linewidth]{%
\includegraphics[width=0.96\linewidth]{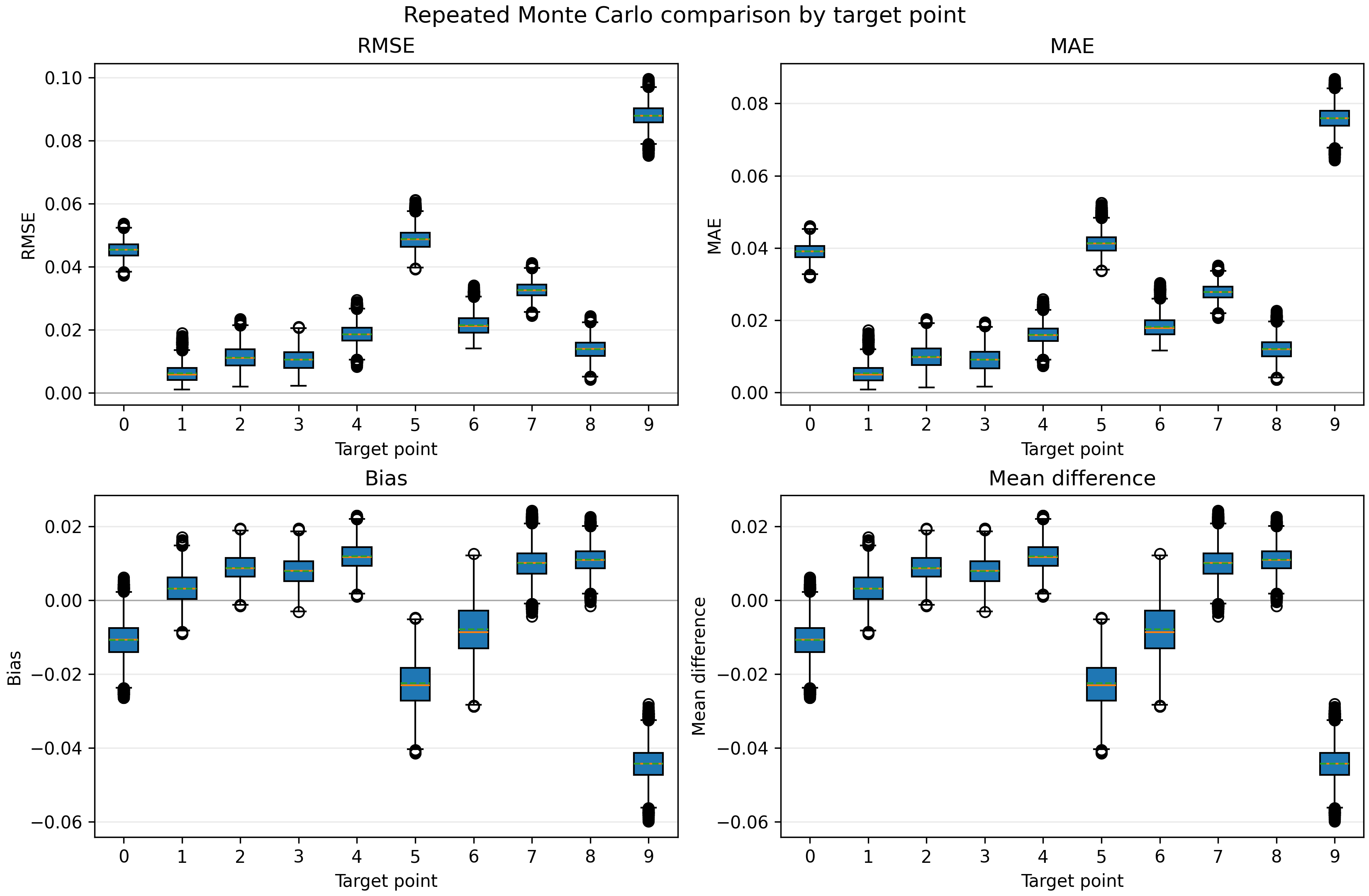}}
&
\subcaptionbox{Behavior: Logistic; Target: Uniform\label{fig:mc_LU}}[0.49\linewidth]{%
\includegraphics[width=0.96\linewidth]{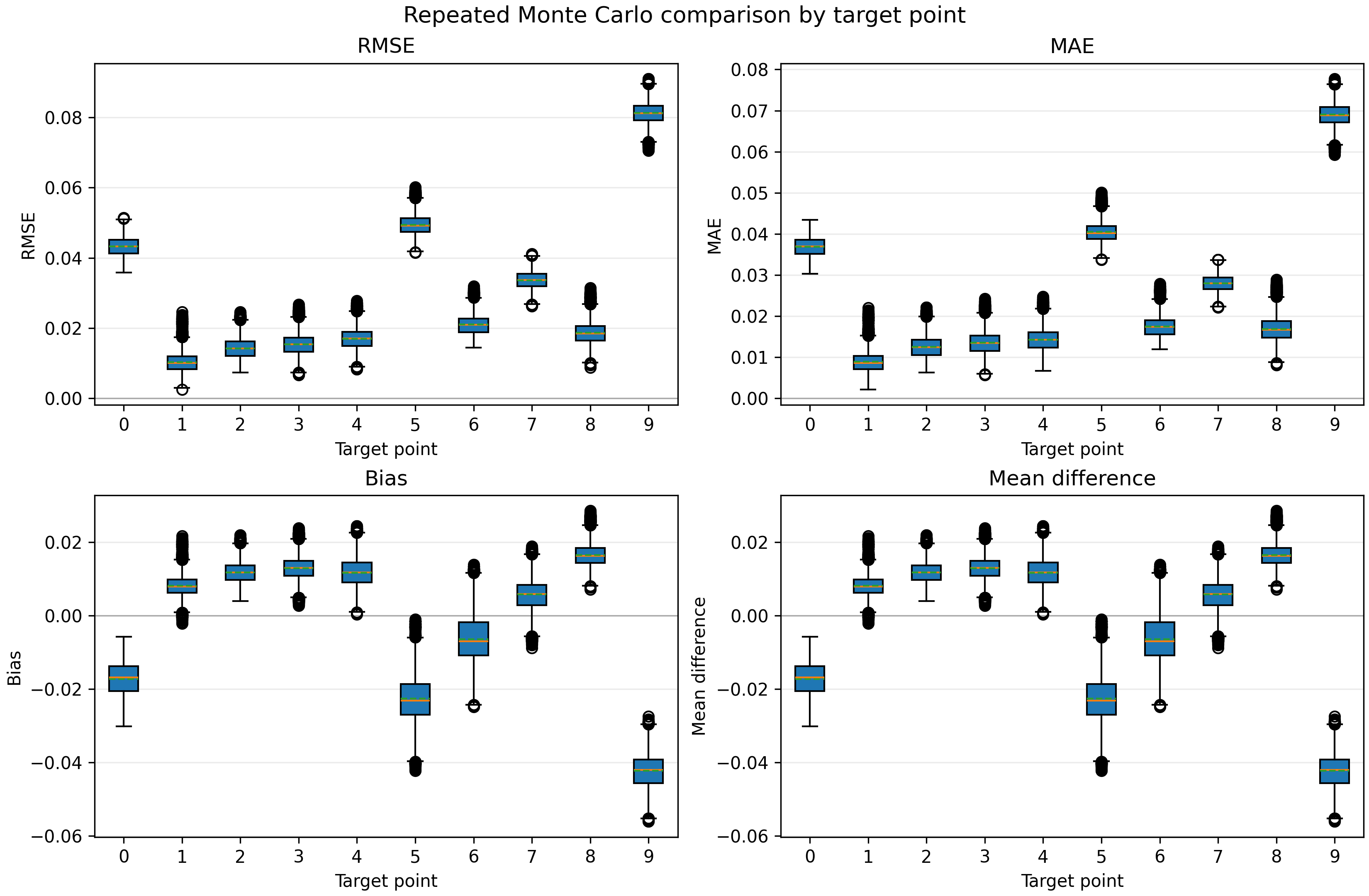}}
\\[0.25em]
\subcaptionbox{Behavior: Uniform; Target: Gaussian\label{fig:mc_UG}}[0.49\linewidth]{%
\includegraphics[width=0.96\linewidth]{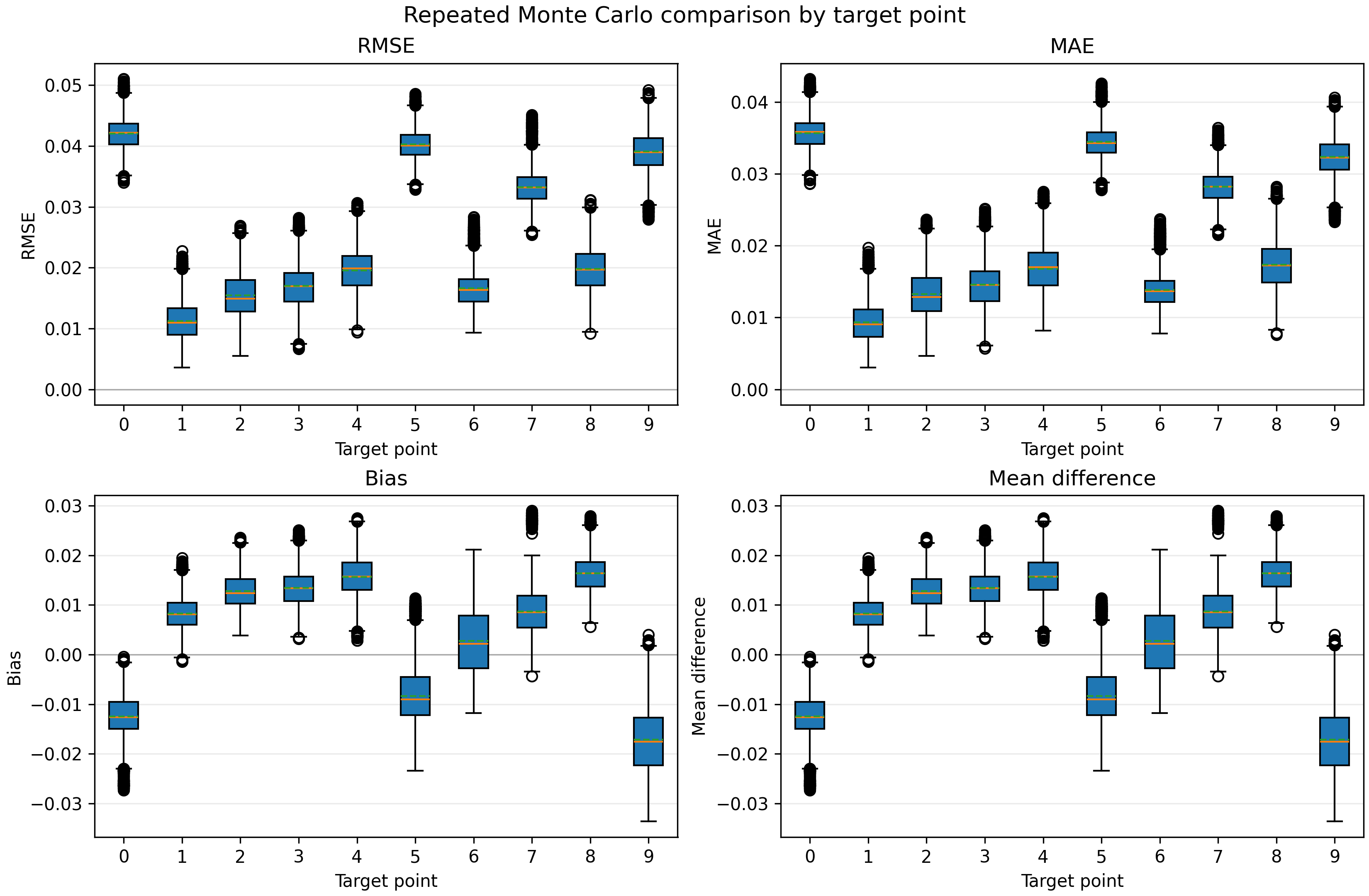}}
&
\subcaptionbox{Behavior: Uniform; Target: Logistic\label{fig:mc_UL}}[0.49\linewidth]{%
\includegraphics[width=0.96\linewidth]{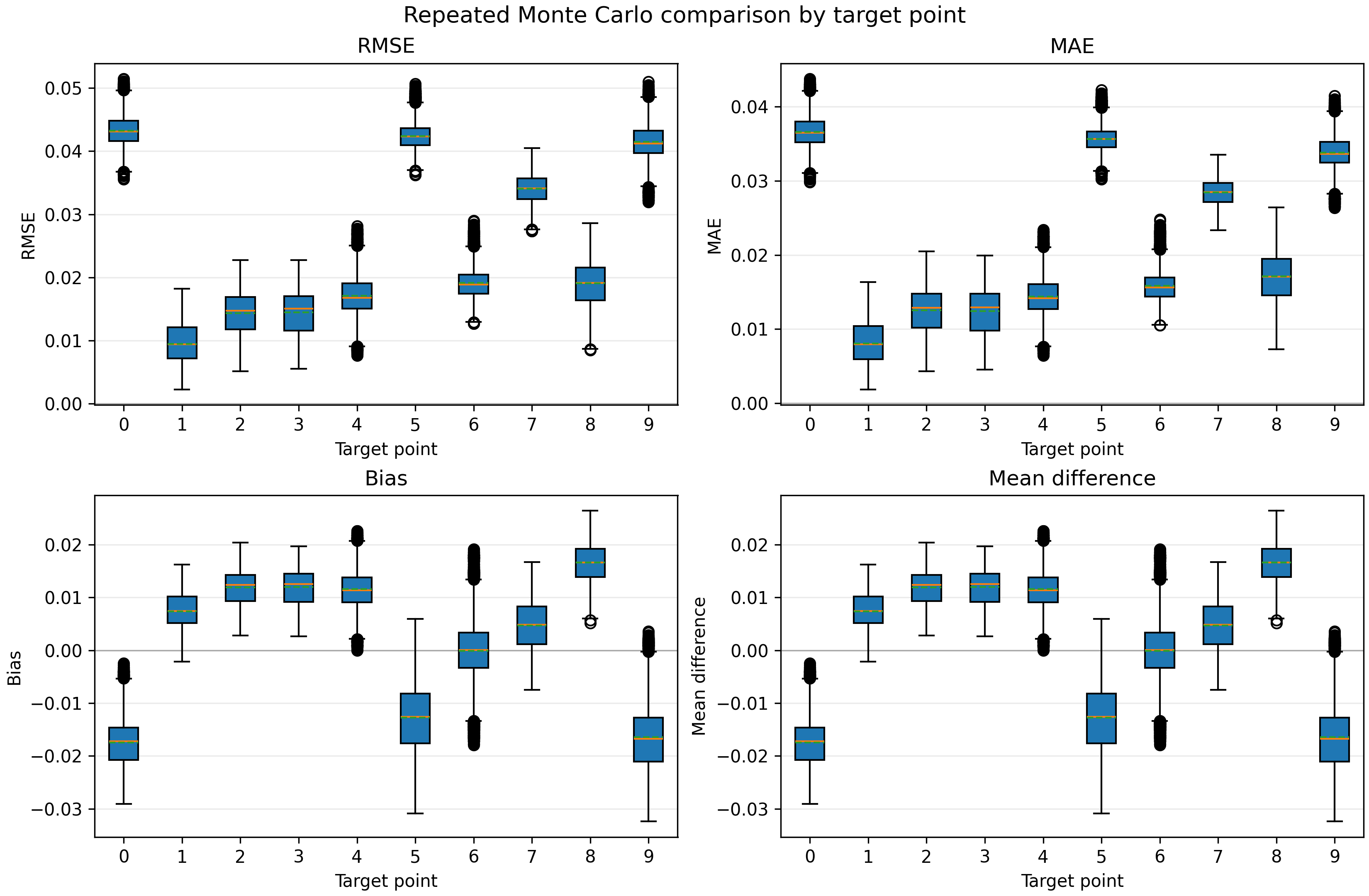}}
\end{tabular}

\caption{\setstretch{0.85}
Pointwise embedding-evaluation discrepancies across repeated Monte Carlo reference samples for six off-policy simulation settings. The KE-DRL estimate is held fixed within each panel, so boxplot variation reflects Monte Carlo reference variability rather than sampling variability of the estimator. These summaries are not RKHS-norm or MMD discrepancies.}
\label{fig:MC_box_plots}
\end{figure}

\section{Expedia Hotel Search Data Analysis}\label{app:Exp_data}

This appendix documents the empirical analysis underlying the Expedia illustration in Section~\ref{sec:realdata}. The application is a methodological case study in offline multivariate distributional policy evaluation. Its purpose is not to estimate a causal business effect of a hotel-ranking intervention, but to illustrate the KE-DRL estimation procedure on a large observational platform dataset with a multivariate one-step reward, continuous action summaries, and policy-indexed conditional return summaries. The reported fit uses the separated Bellman construction under Assumption~\ref{ass:reward_continuation_separation}, and its summaries are interpreted under this working moment restriction.

\subsection{Source Data and Trajectory Construction}

We use a cleaned subset of the Expedia Hotel Search data from the Kaggle \emph{Expedia Personalized Sort} competition \citep{expedia_data}. In the raw data each row corresponds to a hotel property returned for a user search; the original variables include search-level descriptors, hotel characteristics, price information, promotion indicators, competitor comparisons, clicks, bookings, and transaction value.

The raw data are property-level, so to construct a sequential panel we first aggregate property-level rows to the search-list level and treat each search list as one decision point. Trajectories are indexed by destination: destination \texttt{srch\_destination\_id}$=i$ defines a trajectory, and searches within the same destination are ordered by their time index $t=1,\ldots,T_i$, so destination $i$ yields
\[
\left\{(\bs_{i1},\ba_{i1},\br_{i1}),\ldots,(\bs_{iT_i},\ba_{iT_i},\br_{iT_i})\right\}
\]
We construct the transition sample by pairing successive searches within each destination trajectory:
\[
\left\{(\bs_{it},\ba_{it},\br_{it},\bs_{i,t+1},\ba_{i,t+1})\right\}_{t=1}^{T_i-1}
\]
dropping rows without a valid successor. This yields a destination-indexed offline RL dataset and should be read as an approximate sequential abstraction of an observational platform dataset rather than as evidence that the action summaries were randomized.

The list-level state vector contains seven raw variables:
\begin{align*}
\bs_{it}= \Big(
&\texttt{mean\_hist\_price}_{it},
\texttt{std\_hist\_price}_{it},
\texttt{prop\_location\_score1}_{it},\\
& \texttt{srch\_room\_count}_{it}, \texttt{srch\_length\_of\_stay}_{it},
\texttt{prop\_review\_score}_{it},
\texttt{comp\_inv}_{it}
\Big)^\top
\end{align*}
The categorical state variables \texttt{srch\_room\_count}, \texttt{srch\_length\_of\_stay}, and \texttt{comp\_inv} are one-hot encoded before policy fitting and KE-DRL evaluation. The raw state vector has dimension $7$, and the resulting encoded state vector has dimension $28$.

The two-dimensional reward is
\[
\br_{it}
=
\begin{bmatrix}
r^{(1)}_{it}\\[0.1em]
r^{(2)}_{it}
\end{bmatrix}
=
\begin{bmatrix}
\texttt{gross\_revenue\_per\_night}_{it}\\[0.1em]
\texttt{total\_clicks}_{it}
\end{bmatrix}
\]
The first coordinate is continuous and nonnegative, and the second is a nonnegative count. This mixed structure provides an example in which policies can be compared through a joint return distribution rather than a scalarized outcome.

The platform's underlying ranking, pricing, and property-level intervention variables are not observed. We therefore define a low-dimensional continuous action descriptor for each displayed list
\[
\ba_{it}
=
\begin{bmatrix}
\texttt{avg\_price\_per\_night}_{it}\\
\texttt{total\_promotions}_{it}\\
\texttt{std\_price\_usd}_{it}
\end{bmatrix}
\]
These coordinates summarize the average displayed nightly price, the number of promoted properties, and the within-list price dispersion, and should be interpreted as descriptors of the realized display rather than randomized interventions. The empirical action space is treated as a subset of $\mathbb R^3$, with \texttt{total\_promotions} rounded when deterministic policy summaries are displayed.

\begin{table}[H]
\centering
\small
\caption{Variables used in the Expedia analysis.}
\label{tab:expedia_data_vars}
\resizebox{\textwidth}{!}{
\begin{tabular}{@{}llp{10cm}@{}}
\toprule\toprule
\textbf{Variable} & \textbf{Role or scale} & \textbf{Definition} \\
\midrule
\texttt{srch\_destination\_id} & Trajectory identifier & Destination identifier used to define trajectories \\
\texttt{time\_idx} & Ordering index & Within-destination ordering index for searches \\
\midrule
\underline{\textbf{State Variables}}\\
\texttt{mean\_hist\_price} & Continuous & Mean historical displayed price across properties in the list \\
\texttt{std\_hist\_price} & Continuous & Standard deviation of historical displayed prices across properties in the list \\
\texttt{prop\_location\_score1} & Continuous score & List-level location desirability score \\
\texttt{srch\_room\_count} & Count & Number of hotel rooms specified in the search \\
\texttt{srch\_length\_of\_stay} & Count & Number of nights searched \\
\texttt{prop\_review\_score} & Continuous score & List-level customer review score \\
\texttt{comp\_inv} & Binary indicator & Competitor inventory indicator aggregated at the list level \\
\midrule
\underline{\textbf{Action Variables}}\\
\texttt{avg\_price\_per\_night} & Continuous & Average displayed nightly price across properties in the list \\
\texttt{total\_promotions} & Count & Total number of promoted properties in the list \\
\texttt{std\_price\_usd} & Continuous & Standard deviation of displayed prices across properties in the list \\
\midrule
\underline{\textbf{Reward Variables}}\\
\texttt{gross\_revenue\_per\_night} & Continuous & Aggregate revenue outcome at the list level \\
\texttt{total\_clicks} & Count & Total number of clicks across properties in the list \\
\bottomrule\bottomrule
\end{tabular}}
\end{table}

\begin{table}[H]
\centering
\caption{Summary statistics for the Expedia training sample. All variables are reported on the raw scale.}
\label{tab:expedia_data_desc}
\begin{tabular}{lrrrrrrr}
\toprule\toprule
\textbf{Variable} & \textbf{Mean} & \textbf{SD} & \textbf{Min.} & \textbf{Q1} & \textbf{Median} & \textbf{Q3} & \textbf{Max.} \\
\midrule
\underline{\textbf{State Variables}}\\
\texttt{mean\_hist\_price}      & 5.006 & 0.299 & 1.856 & 4.794 & 4.981 & 5.203 & 6.140 \\
\texttt{std\_hist\_price}       & 0.256 & 0.109 & 0.000 & 0.199 & 0.250 & 0.303 & 2.542 \\
\texttt{prop\_location\_score1} & 2.791 & 1.450 & 0.000 & 1.790 & 2.710 & 3.830 & 6.950 \\
\texttt{srch\_room\_count}      & 1.132 & 0.459 & 1.000 & 1.000 & 1.000 & 1.000 & 8.000 \\
\texttt{srch\_length\_of\_stay} & 2.333 & 1.818 & 1.000 & 1.000 & 2.000 & 3.000 & 14.000 \\
\texttt{prop\_review\_score}    & 3.937 & 0.620 & 1.000 & 3.500 & 4.000 & 4.500 & 5.000 \\
\texttt{comp\_inv}              & 0.453 & 0.498 & 0.000 & 0.000 & 0.000 & 1.000 & 1.000 \\
\midrule
\underline{\textbf{Action Variables}}\\
\texttt{avg\_price\_per\_night} & 126.769 & 27.919 & 64.727 & 105.247 & 124.002 & 146.644 & 210.000 \\
\texttt{total\_promotions}      & 3.579 & 4.316 & 0.000 & 1.000 & 2.000 & 5.000 & 29.000 \\
\texttt{std\_price\_usd}        & 29.531 & 10.669 & 0.566 & 22.485 & 30.393 & 36.848 & 79.977 \\
\midrule
\underline{\textbf{Reward Variables}}\\
\texttt{gross\_revenue\_per\_night} & 4.755 & 24.307 & 0.000 & 0.000 & 0.000 & 0.000 & 213.250 \\
\texttt{total\_clicks}              & 0.321 & 0.610 & 0.000 & 0.000 & 0.000 & 1.000 & 11.000 \\
\midrule
Total Observations & 10000 & & & & & & \\
\bottomrule\bottomrule
\end{tabular}
\end{table}

\subsection{Construction of Example Target Policies}\label{subsec:expedia_policy_construction}

We construct two fixed reward-specific target policies from the same offline sample. The revenue-focused policy $\pi^{(\mathrm{rev})}$ uses the reward coordinate \texttt{gross\_revenue\_per\_night}, and the click-focused policy $\pi^{(\mathrm{clk})}$ uses \texttt{total\_clicks}. Both policies are fitted directly as reward-weighted linear-Gaussian models using the logged state-action pairs.

Let $S^{\mathrm{enc}}$ denote the encoded state vector. For reward coordinate $c\in\{\mathrm{rev},\mathrm{clk}\}$, the policy is fitted on the standardized action scale as
\[
A^{\mathrm{std}}
\mid
S^{\mathrm{enc}}=s
\sim
\mathcal N_3\!
\left(
s^\top\Theta_{\mu,c}+\epsilon_{\mu,c}
\quad
\operatorname{diag}\!
\left\{
\exp\!
\left(
s^\top\Theta_{\sigma,c}+\epsilon_{\sigma,c}
\right)^2
\right\}
\right)
\]
The mean and log-standard-deviation parameters are estimated by reward-weighted Gaussian likelihood, which gives greater weight to logged actions with larger values of reward coordinate $c$. The coordinatewise clipping described below limits marginal extrapolation but does not establish joint overlap.

The fitted standardized policy is mapped back to the raw action scale before KE-DRL evaluation. For deterministic summaries and figures we use the clipped greedy action
\[
a_c^{\mathrm{gr}}(s)=\Pi_{\mathcal A}\!\left\{\mu_c(s)\right\}
\]
where $\Pi_{\mathcal A}$ clips each coordinate to its observed training range and \texttt{total\_promotions} is rounded to the nearest integer. This produces an interpretable action profile for each policy while preventing extrapolation beyond the observed marginal action ranges.

Table~\ref{tab:expedia_policy_action_summary} reports the greedy action means on the test split. Relative to the logged behavior, the revenue-focused policy slightly increases average displayed price and price dispersion while keeping promotions close to the logged level, whereas the click-focused policy lowers average displayed price, increases promotions, and lowers price dispersion. The two policies are thus behaviorally distinct, but the separation is moderate rather than extreme.

\begin{table}[tbh]
\centering
\caption{Greedy action summaries for the logged behavior and the two constructed target policies on the test split, together with held-out-input projected Bellman risk. Policy actions are evaluated after clipping to the observed coordinatewise training ranges and rounding of \texttt{total\_promotions}. The risk measures held-out agreement with the projected Bellman equation and is not a business-performance measure.}
\label{tab:expedia_policy_action_summary}
\begin{tabular}{lccc}
\toprule\toprule
\textbf{Action Component}
& \textbf{Logged Behavior}
& \textbf{$\pi^{(\mathrm{rev})}$}
& \textbf{$\pi^{(\mathrm{clk})}$} \\
\midrule
\texttt{avg\_price\_per\_night} & 128.6127 & 130.9611 & 123.2302 \\
\texttt{total\_promotions}      &   3.6080 &   3.4828 &   5.1903 \\
\texttt{std\_price\_usd}        &  29.5638 &  31.2111 &  26.9577 \\
\midrule
Held-out-input projected Bellman risk & NA & 0.0513 & 0.0478 \\
\bottomrule\bottomrule
\end{tabular}
\end{table}

The mean normalized Euclidean distance between the two greedy action rules on the test split is $0.6326$, with median $0.6711$, and the mean absolute raw-scale differences are $7.7308$ for \texttt{avg\_price\_per\_night}, $1.7075$ for \texttt{total\_promotions}, and $4.2534$ for \texttt{std\_price\_usd}. These diagnostics indicate that the two target policies encode different reward priorities while remaining comparable on the same empirical action scale. Despite these level shifts, their conditional mean-action rules are highly correlated across test states: the coordinatewise correlations between $a^{\mathrm{gr}}_{\mathrm{rev}}(s)$ and $a^{\mathrm{gr}}_{\mathrm{clk}}(s)$ are $0.9999$, $0.9739$, and $0.9990$ for price, promotions, and price dispersion. These correlations indicate similar state dependence, with the policy differences occurring primarily in their levels. Both Gaussian policies also have substantial conditional dispersion. The average conditional standard deviations on the test split are approximately $22.7$ for \texttt{avg\_price\_per\_night}, $3.7$ for \texttt{total\_promotions}, and $9.5$ for \texttt{std\_price\_usd}. Relative to this conditional dispersion, the between-policy mean differences correspond to approximately $0.34$, $0.46$, and $0.45$ conditional standard deviations in the three coordinates. These summaries suggest substantial coordinatewise overlap between the two action distributions, although they do not establish joint overlap. This pattern is consistent with the similarity of the estimated embeddings.

\begin{figure}[H]
\centering
\subfloat[Raw action distributions]{%
\centering
\includegraphics[width=0.98\linewidth]{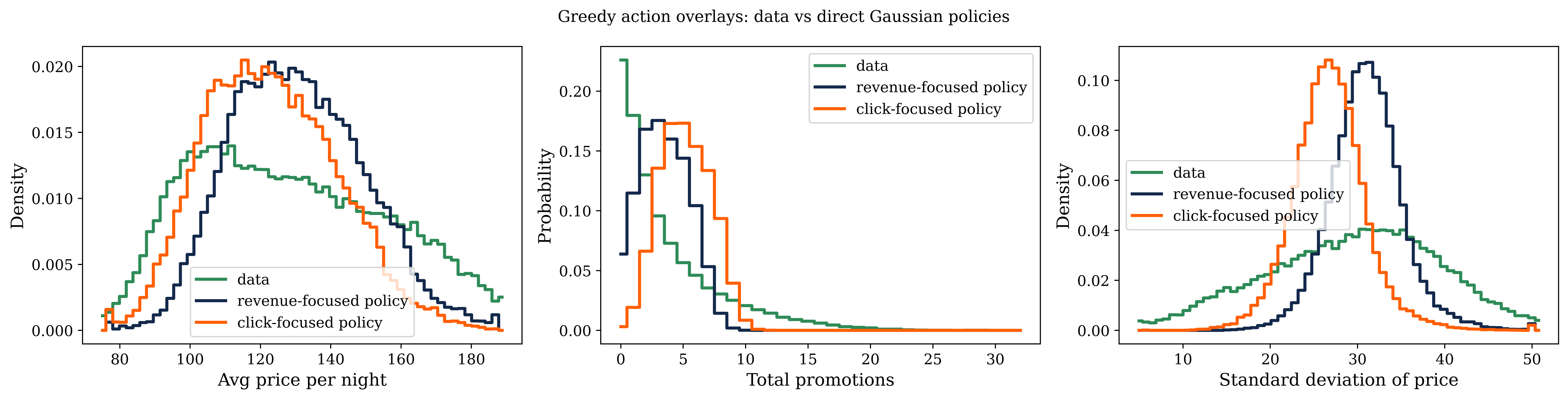}
\label{fig:overlay_action_distributions_greedy}}\\
\subfloat[Standardized action profiles]{%
\centering
\includegraphics[width=0.7\linewidth]{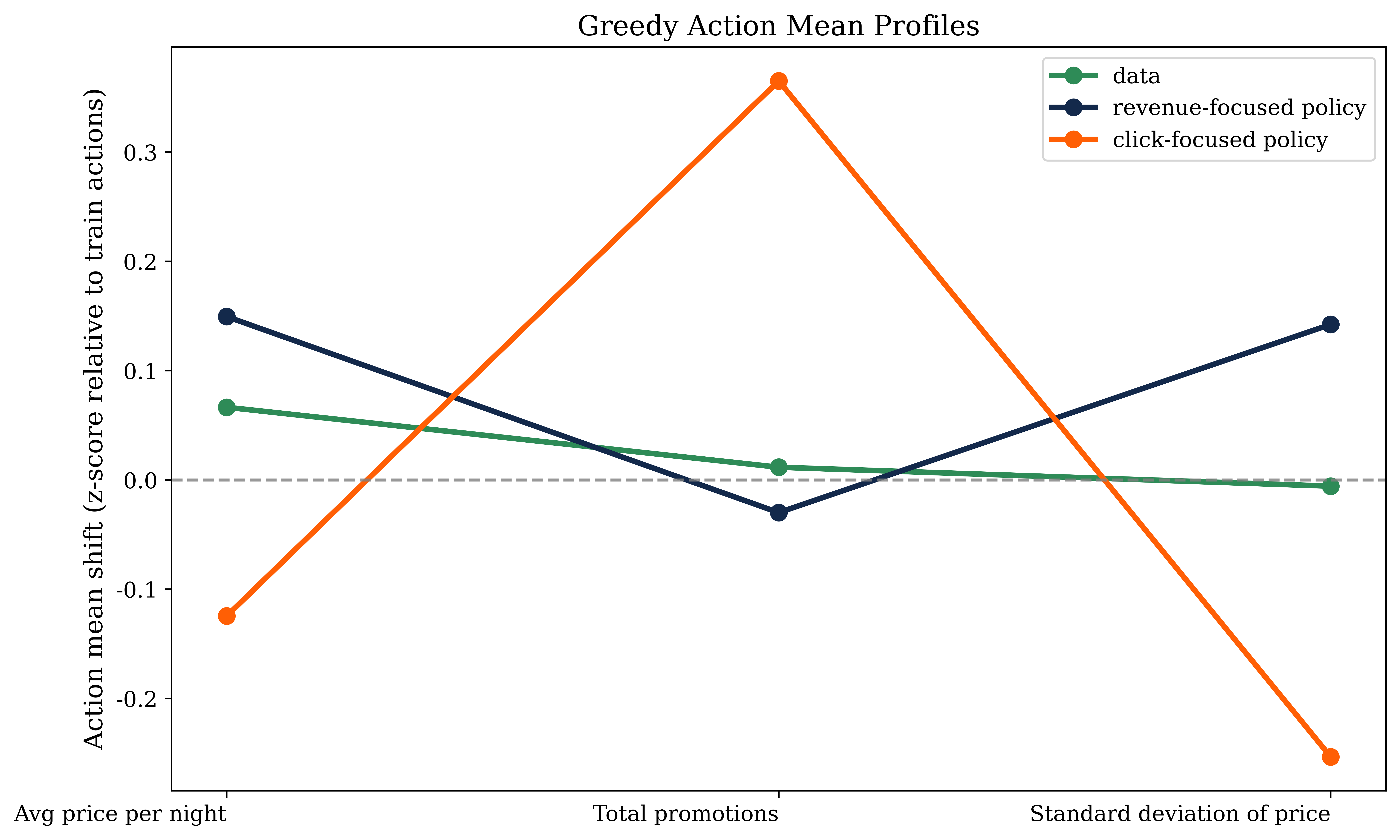}
\label{fig:overlay_action_mean_profile_zscore}}\\
\vspace{1em}
\caption{\setstretch{0.85} Action-level summaries for the two constructed target policies. The policies have distinct but overlapping action distributions on the same logged-data scale.}
\label{fig:expedia_policies}
\end{figure}

\subsection{KE-DRL Evaluation Design}\label{subsec:expedia_policy_eval_details}

Given the two fitted target policies, the subsequent KE-DRL analysis conditions on their realized fits, and the policies are not re-estimated during this step. The reported comparison therefore does not propagate policy-construction uncertainty. The goal is to estimate and compare their conditional bivariate return embeddings under a common offline design. The analysis uses the encoded state vector, the three-dimensional action descriptor, the two-dimensional reward, $10{,}000$ training transitions, and $4{,}000$ test transitions. Both policies are evaluated with the same kernel, regularization parameters, return grid, state-action basis, post-estimation recovery procedure, and target-state selection rule. Thus the displayed differences are not induced by changes in the estimation or tuning specification. Table~\ref{tab:expedia_appendix_setup} gives the complete numerical specification.

\begin{table}[t]
\centering
\small
\caption{Common numerical specification for the two Expedia target-policy evaluations.}
\label{tab:expedia_appendix_setup}
\begin{tabular}{@{}p{0.36\textwidth}p{0.58\textwidth}@{}}
\toprule\toprule
Quantity & Value \\
\midrule
Raw state variables / encoded state dimension & \(7\) / \(28\) \\
Action dimension / reward dimension & \(3\) / \(2\) \\
Training / test size & \(10000 / 4000\) \\
Discount factor & \(\gamma=0.8\) \\
Return-grid size & \(m=300\) \\
State-action basis & \(L=6000\), \(k\)-means centers, basis ridge \(10^{-6}\) \\
Embedding optimization & AdamW, \(40000\) iterations, initial learning rate \(5\times10^{-5}\), weight decay \(0\) \\
Coefficient-mass penalty / target mass & \(\lambda_{\mathrm{mass}}=10\), \(c_{\mathrm{mass}}=1\) \\
Coefficient regularization / constraints & \(\lambda_B=10^{-10}\), \(\lambda_{\mathrm{neg}}=0\); no simplex calibration or nonnegativity constraint \\
Return-space kernel & Mat\'ern, \((\nu_Z,\ell_Z,\sigma_Z)=(3.5,3.0,1.0)\) \\
State-action kernel & Mat\'ern, \((\nu_{SA},\ell_{SA},\sigma_{SA})=(3.5,1.0,1.0)\) \\
Conditional-weight ridge & \(\lambda_{\Gamma}=10^{-4}\) \\
Density-ratio (uLSIF) & \(1500\) numerator-drawn basis points, \(3\times\) target samples, nonnegative coefficients, ridge \(10^{-2}\), mean calibration to \(1\) \\
Post-estimation recovery optimization & \(20000\) iterations, learning rate \(10^{-2}\), reference vector and initialization from the positive part of \(\widehat{\boldsymbol\beta}^{\pi}\), \(\lambda_2=10^{-2}\), \(\lambda_{\mathrm{KL}}=0\) \\
Recovery ridge & \(\lambda_{\mathrm{den}}=\max(10^{-5},\,10^{-6}\lambda_{\max}(\mathbb K_{\mc Z}))\) (spectrum-scaled) \\
Revenue smoothing bandwidth & \(h_R=70\) (raw revenue scale); click-return proxy discrete \\
Click recovery convention & Discounted click-return coordinate projected to observed one-step count support \\
Target-state selection & median first principal-component score over the training states \\
\bottomrule\bottomrule
\end{tabular}
\end{table}

For both policies, the clipped conditional mean actions lie within the observed training range of each action coordinate, and the same is true of sampled actions after clipping. This check rules out extrapolation beyond the observed marginal ranges by construction. Beyond this coordinatewise support check, the off-policy Bellman update uses an estimated target-to-behavior density ratio fitted by uLSIF. The $10000$ logged successor inputs $(\bs_j',\ba_j')$ form the denominator sample. At each logged successor state $\bs_j'$, three target actions are drawn from $\pi(\cdot\mid\bs_j')$, giving $30000$ state-matched numerator inputs. The fit uses $1500$ numerator-drawn basis points, ridge $10^{-2}$, and nonnegative coefficients. The fitted ratio is rescaled so that its empirical mean over the logged denominator sample equals one, with calibration factors $1.4760$ for $\pi^{(\mathrm{rev})}$ and $1.3393$ for $\pi^{(\mathrm{clk})}$. No upper clipping is applied, and the calibrated ratios range over $[0, 4.39]$ and $[0, 6.24]$, respectively. The corresponding effective sample sizes are $5767.3$ and $5202.0$, or $57.7\%$ and $52.0\%$ of the logged sample. Thus the estimated weights are not concentrated on a small fraction of the observations in the reported analysis. The marginal-range and fitted-weight diagnostics are empirical checks only; they establish neither joint population overlap nor causal identification.

For a fixed target policy $\pi$, define the bivariate discounted return
\[
\mathbf Z^\pi=\sum_{t\geq 0}\gamma^t\mathbf R_t^\pi\in\mathbb R^2
\]
KE-DRL estimates the conditional kernel mean embedding
\[
\mu_{\mathbf Z^\pi\mid \bx}
=
\E\!\left\{k_{\mathcal Z}\!\left(\mathbf Z^\pi,\cdot\right)\mid \bX=\bx\right\}
\]
where $\bX=(\bS^\top,\bA^\top)^\top$. With return-grid atoms $\mathbf Z_m=\{\bz_1,\ldots,\bz_m\}$, the fitted finite-dictionary RKHS element is
\[
\widehat\mu_{\mathbf Z^\pi\mid \bx}
=
\sum_{j=1}^{m}\widehat\beta_{\pi,j}(\bx)\,k_{\mathcal Z}\!\left(\bz_j,\cdot\right)
\]
with coefficients obtained from the fitted state-action basis dictionary
\[
\widehat{\boldsymbol\beta}_{\pi}(\bx)
=
(\widehat{\mathbb B}^{\pi})^{\top}\bk_L(\bx)
\]
where $\mc C_L$ is the selected $L=6000$ state-action basis and $\widehat{\mathbb B}^{\pi}\in\R^{L\times m}$. With $m=300$, the coefficient matrix has dimension $6000\times 300$.

Among the evaluated AdamW iterates, we select the one with the smallest projected Bellman criterion over the full training conditioning set and evaluate the corresponding embedding at the $4000$ held-out state-action inputs. The resulting risk measures held-out agreement with the projected Bellman equation through the squared RKHS residual; it is not an estimate of revenue, clicks, welfare, or business value.

\begin{table}[tbh!]
\centering
\caption{Held-out-input projected Bellman risks for the two Expedia target-policy evaluations. Smaller values indicate closer held-out agreement with the plug-in projected Bellman equation under the common estimation and tuning specification, but the risks are not business-performance measures.}
\label{tab:expedia_policy_eval_risks}
\begin{tabular}{lc}
\toprule\toprule
\textbf{Policy} & \textbf{Held-out-Input Projected Bellman Risk} \\
\midrule
\(\pi^{(\mathrm{rev})}\) & 0.0513 \\
\(\pi^{(\mathrm{clk})}\) & 0.0478 \\
\bottomrule\bottomrule
\end{tabular}
\end{table}

The two risks are of similar magnitude under the common estimation and tuning specification. The slightly smaller risk for $\pi^{(\mathrm{clk})}$ should not be interpreted as evidence that the click-focused policy is better; it indicates only a smaller held-out-input projected Bellman risk in this specification.

\enlargethispage{\baselineskip}
The reported fit used a soft coefficient-mass penalty evaluated at the training conditioning points, with $\lambda_{\mathrm{neg}}=0$ and no simplex calibration. It therefore imposed neither exact unit coefficient mass nor nonnegativity. Over the $4000$ test inputs, the coefficient row sums average $1.126$ (standard deviation $1.056$) for $\pi^{(\mathrm{rev})}$ and $1.105$ (standard deviation $0.868$) for $\pi^{(\mathrm{clk})}$, and the average fractions of negative coefficient mass are $0.187$ and $0.193$, respectively. These diagnostics quantify how the raw embedding coefficients depart from probability weights out of sample and motivate the simplex-constrained recovery step in Section~\ref{subsec:expedia_density_recovery}. Probability summaries are computed from the recovered weights $\widehat w\in\Delta_m$, not directly from $\widehat{\boldsymbol\beta}_{\pi}(\bx)$.

\subsection{Recovery of Probability Summaries}\label{subsec:expedia_density_recovery}

\begin{figure}[H]
\centering
\includegraphics[width=0.88\textwidth]{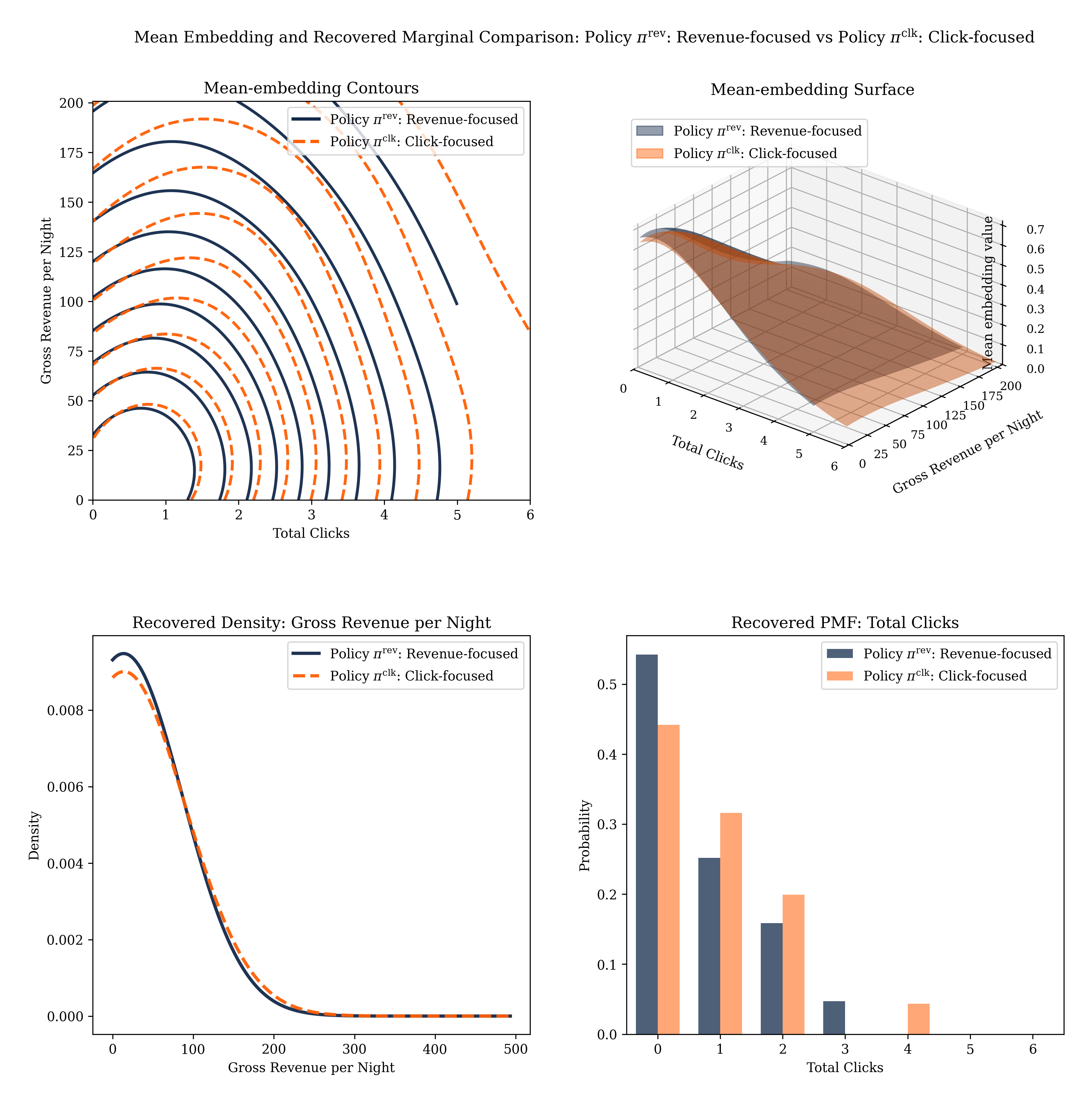}
\caption{Estimated conditional return embeddings and post-estimation marginal proxies for the two Expedia target policies. The top panels show the estimated RKHS mean-embedding functions as contours and surfaces. The bottom panels show a smoothed density for gross revenue per night and a discrete proxy for the discounted click-return coordinate.}
\label{fig:expedia_res_full}
\end{figure}

\noindent The finite-dictionary KE-DRL estimate is an RKHS element represented by embedding coefficients. These coefficients are neither probabilities nor constrained to be nonnegative, so the bottom panels of Figure~\ref{fig:expedia_res_full} do not interpret $\widehat{\boldsymbol\beta}_{\pi}(\bx)$ as a probability vector. Instead, for visualization we construct a mixed continuous-discrete proxy law whose induced embedding approximates the estimated KE-DRL element. This proxy is not a unique inversion, and its discrete click support is an imposed recovery convention rather than an asserted support property of the discounted click return.

\noindent Let $\bx_\star^\pi$ denote the policy-specific state-action evaluation point used in Figure~\ref{fig:expedia_res_full}. Its state component is the training state whose first principal-component score is closest to the median score over the training sample. For the reported analysis, the selected list has mean historical log price $5.19$, price dispersion $0.23$, location score $4.23$, one room, a two-night stay, review score $4.0$, and no competitor inventory. Its action component is the greedy action of the evaluated policy at that state: $(143.0, 4, 30.6)$ for $\pi^{(\mathrm{rev})}$ and $(135.3, 6, 26.3)$ for $\pi^{(\mathrm{clk})}$ on the raw (price, promotions, price-dispersion) scale, compared with a logged action of $(142.7, 9, 28.9)$ at the same state. The estimated embedding at this input is
\[
\widehat\mu_{\mathbf Z^\pi\mid \bx_\star^\pi}
=
\sum_{i=1}^{m}\widehat\beta^\pi_i\,k_{\mathcal Z}\!\left(\bz_i,\cdot\right)
\]
with $\widehat\beta^\pi_i=\widehat\beta_{\pi,i}(\bx_\star^\pi)$. Each normalized grid atom $\bz_i=(z_{i,R},z_{i,C})^\top$ is mapped to the raw reward scale by
\[
\by_i
=
\widehat{\bm m}_{\br}+\widehat{\bs}_{\br}\odot\bz_i
=
(r_i,c_i)^\top
\]
where $r_i$ and $c_i$ are the raw-scale coordinates of a discounted-return grid atom. Although the one-step variable \texttt{total\_clicks} is count-valued, the discounted click return $\sum_{t\ge0}\gamma^t C_t$ need not be integer-valued. For the displayed recovery only, we impose a discrete proxy by projecting the click-return coordinate onto the observed one-step count support:
\[
\bar c_i=\Pi_{\mathcal C}\!\left(c_i\right)
\]
with $\mathcal C$ the observed support of \texttt{total\_clicks}.

We estimate the atom weights by induced-embedding matching under the simplex constraint:
\[
\Delta_m=\left\{w\in\mathbb R^m\mid w_i\ge 0\quad\sum_{i=1}^{m}w_i=1\right\}
\]
For a candidate $w\in\Delta_m$, define the mixed continuous-discrete proxy law
\[
Q_w(dr,c)
=
\sum_{i=1}^{m}w_i\,\phi_{h_R}\!\left(r-r_i\right)dr\,\mathbb 1\!\left\{c=\bar c_i\right\}
\]
where $\phi_{h_R}$ is a Gaussian smoothing kernel on the raw revenue scale and smoothing is applied only to gross revenue per night, while the click coordinate remains discrete in the proxy. The embedding induced by $Q_w$ is
\[
\mu_w(\cdot)
=
\sum_{i=1}^{m}w_i\int k_{\mathcal Z}\!\left(\frac{(r,\bar c_i)^\top-\widehat{\bm m}_{\br}}{\widehat{\bs}_{\br}},\cdot\right)\phi_{h_R}\!\left(r-r_i\right)dr
\]
Writing this in the return-grid dictionary gives
\[
\widetilde{\bm\beta}(w)
=
\left(\mathbb K_{\mc Z}+\lambda_{\mathrm{den}}\mathbb I_m\right)^{-1}Aw
\]
with
\[
\left[\mathbb K_{\mc Z}\right]_{\ell q}=k_{\mathcal Z}\!\left(\bz_\ell,\bz_q\right)
\qquad
A_{\ell i}=\int k_{\mathcal Z}\!\left(\bz_\ell,\frac{(r,\bar c_i)^\top-\widehat{\bm m}_{\br}}{\widehat{\bs}_{\br}}\right)\phi_{h_R}\!\left(r-r_i\right)dr
\]
The ridge parameter $\lambda_{\mathrm{den}}$ is used only in this post-estimation recovery step and is distinct from the KE-DRL Bellman regularization parameter; it is scaled to the spectrum of the grid Gram matrix, $\lambda_{\mathrm{den}}=\max\{10^{-5},10^{-6}\lambda_{\max}(\mathbb K_{\mc Z})\}$, so that the conditioning of the inverse is comparable across kernel length scales. The recovered weights are chosen by induced-embedding matching
\[
\widehat w
=
\arg\min_{w\in\Delta_m}
\left[
\left\{\widetilde{\bm\beta}(w)-\widehat{\bm\beta}^{\pi}\right\}^{\top}
\mathbb K_{\mc Z}
\left\{\widetilde{\bm\beta}(w)-\widehat{\bm\beta}^{\pi}\right\}
+
\lambda_2\left\|w-w_0\right\|_2^2
+
\lambda_{\mathrm{KL}}\operatorname{KL}\!\left(w\|w_0\right)
\right]
\]
For the reported recovery, both the initial value and the reference vector $w_0$ are obtained from the normalized positive part of the fitted embedding coefficients, with $\lambda_2=10^{-2}$ and $\lambda_{\mathrm{KL}}=0$. The simplex constraint is enforced through the softmax parameterization
\[
w_i(\theta)=\frac{\exp(\theta_i)}{\sum_{j=1}^{m}\exp(\theta_j)}
\]
The criterion is minimized by Adam for $20000$ iterations with learning rate $10^{-2}$.
Using $\widehat w$, the recovered revenue marginal is
\[
\widehat f_R(r)=\sum_{i=1}^{m}\widehat w_i\,\phi_{h_R}\!\left(r-r_i\right)
\]
the recovered click proxy is the probability mass function
\[
\widehat p_C(c)=\sum_{i=1}^{m}\widehat w_i\,\mathbb 1\!\left\{\bar c_i=c\right\}
\]
and the mixed joint summary is
\[
\widehat p(r,c)=\sum_{i=1}^{m}\widehat w_i\,\phi_{h_R}\!\left(r-r_i\right)\mathbb 1\!\left\{\bar c_i=c\right\}
\]
These recovered summaries are valid probability representations of the imposed proxy law. They should not be interpreted as a unique inversion of the RKHS embedding or as an assertion that the discounted click return is intrinsically count-valued. Accordingly, Table~\ref{tab:expedia_recovered_marginals}, which reports the recovered marginal proxies for both policies at their respective targets $\bx_\star^\pi$, should be read as a sensitivity analysis under the chosen recovery specification rather than as a direct characterization of the return distribution. Consistent with the ill-posedness analysis of Supplementary Material~\ref{app:emb_ill_posed}, the recovered click proxies contain isolated masses and support gaps, which may result from simplex-constrained embedding matching on a finite atom grid. The revenue summaries are additionally sensitive to the smoothing bandwidth. With $h_R=70$ on a reward scale whose logged one-step mean is $4.75$ and standard deviation is $24.3$, one-sided smoothing near the origin shifts the recovered revenue-return means ($63.9$ and $67.7$) upward. These values should not be interpreted as calibrated return levels. Comparisons under the common recovery specification avoid differences induced by separate tuning choices, but remain conditional on that specification.

\begin{table}[tbh!]
\centering
\caption{Recovered marginal return proxies at the policy-specific targets $\bx_\star^\pi$, from the common recovery specification of Table~\ref{tab:expedia_appendix_setup}. Revenue rows refer to the recovered discounted gross-revenue-per-night marginal. Click rows refer to the discrete recovery proxy for the discounted click-return coordinate under the imposed observed-count support projection. Absolute revenue levels are sensitive to the smoothing bandwidth; between-column differences should be interpreted conditional on the common recovery specification.}
\label{tab:expedia_recovered_marginals}
\begin{tabular}{lccc}
\toprule\toprule
\textbf{Recovered summary} & \(\pi^{(\mathrm{rev})}\) & \(\pi^{(\mathrm{clk})}\) & \(\Delta\;(\mathrm{clk}-\mathrm{rev})\) \\
\midrule
Revenue mean & 63.92 & 67.66 & $+3.73$ \\
\(\Pr(\text{revenue}\ge 50)\) & 0.5348 & 0.5566 & $+0.0218$ \\
\(\Pr(\text{revenue}\ge 100)\) & 0.2087 & 0.2360 & $+0.0272$ \\
\midrule
Projected click-proxy mean & 0.7103 & 0.8875 & $+0.1772$ \\
\(\Pr(\text{click proxy}\ge 1)\) & 0.4576 & 0.5583 & $+0.1008$ \\
\(\Pr(\text{click proxy}\ge 2)\) & 0.2058 & 0.2424 & $+0.0366$ \\
\(\Pr(\text{click proxy}\ge 3)\) & 0.0470 & 0.0434 & $-0.0036$ \\
\bottomrule\bottomrule
\end{tabular}
\end{table}

\subsection{Interpretation of the Expedia Results}\label{subsec:expedia_interpretation}

\noindent Figure~\ref{fig:expedia_res_full} compares two fitted policy-indexed conditional return summaries at the same selected state and the respective greedy initial actions, so the contrast combines the policy-specific continuation rules and initial actions. The top-left panel shows contours of the fitted RKHS mean-embedding function over the bivariate return grid, and the top-right panel shows the same fitted embedding as a surface; these top panels are embedding-value plots, not probability densities. Within these top panels, the two estimated embedding surfaces are similar. As documented in Section~\ref{subsec:expedia_policy_construction}, the evaluated objects are stochastic Gaussian policies whose conditional standard deviations exceed the between-policy mean differences by factors of two to three and whose conditional mean-action rules have correlations above $0.97$ in every coordinate. The similarity of the estimated embeddings is consistent with these policy summaries. In portions of the grid with larger click-return coordinates, $\pi^{(\mathrm{clk})}$ has relatively larger embedding values than $\pi^{(\mathrm{rev})}$.

\noindent The bottom panels report post-estimation recovery proxies obtained by induced-embedding matching, with numerical summaries in Table~\ref{tab:expedia_recovered_marginals}. The recovered revenue marginals overlap substantially. The recovery proxy for the click-focused policy has a recovered mean that is $3.73$ larger and a recovered probability above $100$ that is $0.027$ larger, but these differences are sensitive to the bandwidth and are not interpreted substantively. The discrete click proxy shows larger differences at the lower projected thresholds: relative to the revenue-focused policy, the proxy under the click-focused policy is higher by $0.1772$ in projected mean, $0.1008$ in mass at or above one, and $0.0366$ in mass at or above two, whereas the difference at or above three is small.

\noindent The results do not indicate that one policy uniformly dominates the other. The procedure yields distinct but substantially overlapping bivariate conditional summaries. The density-ratio effective sample sizes, held-out-input projected Bellman risks, out-of-sample coefficient-mass diagnostics, and recovery sensitivity limit what can be inferred from these summaries. Because the analysis uses observational data and constructed action descriptors, the results are an illustration of offline distributional policy evaluation rather than causal business recommendations. The example applies the proposed estimation procedure to continuous action summaries, a 28-dimensional encoded state, mixed one-step rewards, and policy-indexed conditional return summaries.
\ifsupplementonly
\bibliography{bib_JASA}
\fi
\fi
\end{document}